\documentclass{article}

\usepackage{arxiv}

\usepackage[utf8]{inputenc} 
\usepackage[T1]{fontenc}    
\usepackage{hyperref}       
\usepackage{url}            
\usepackage{booktabs}       
\usepackage{amsfonts}       
\usepackage{nicefrac}       
\usepackage{microtype}      
\usepackage{lipsum}         
\usepackage{graphicx}
\usepackage{natbib}
\usepackage{doi}
\usepackage{xspace}

\usepackage{amsmath}
\usepackage{multirow}
\usepackage{lscape}
\usepackage{tabularx}
\usepackage{csquotes}
\usepackage{longtable}
\usepackage{cleveref}       

\usepackage{tocloft}

\newcommand{\nateme}{\textit{Hum-EmCom}\xspace}
\newcommand{\symeme}{\textit{Mac-EmCom}\xspace}
\newcommand{\emecom}{\textit{EmCom}\xspace}

\hypersetup{%
    pdfborder = {0 0 0}
}
\title{Towards More Human-like AI Communication}

\date{}

\author{ \href{https://orcid.org/0000-0002-3191-6623}{\includegraphics[scale=0.06]{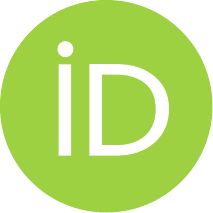}\hspace{1mm}Nicolo' Brandizzi}\\
	Dept. of Computer, Control and Management Engineering\\
	Sapienza University of Rome\\
	via Ariosto 25, Rome 00185, IT \\
	\texttt{brandizzi@diag.uniroma1.it}
}

\renewcommand{\headeright}{Preprint}
\renewcommand{\undertitle}{A Review of Emergent Communication Research}
\renewcommand{\shorttitle}{A Review of Emergent Communication Research}

\hypersetup{
pdftitle={Towards More Human-like AI Communication: A Review of Emergent Communication Research},
pdfsubject={cs.AI,cs.CL, cs.GT, cs.HC, cs.MA},
pdfauthor={Nicolo' Brandizzi},
pdfkeywords={Multi agent systems , Reinforcement learning , Human machine interaction, Emergent communication},
}

\begin{document}
\maketitle

\begin{abstract}
	In the recent shift towards human-centric AI, the need for machines to accurately use natural language has become increasingly important. While a common approach to achieve this is to train large language models, this method presents a form of learning misalignment where the model may not capture the underlying structure and reasoning humans employ in using natural language, potentially leading to unexpected or unreliable behavior. Emergent communication (\emecom) is a field of research that has seen a growing number of publications in recent years, aiming to develop artificial agents capable of using natural language in a way that goes beyond simple discriminative tasks and can effectively communicate and learn new concepts. In this review, we present \emecom under two aspects. Firstly, we delineate all the common proprieties we find across the literature and how they relate to human interactions. Secondly, we identify two subcategories and highlight their characteristics and open challenges. We encourage researchers to work together by demonstrating that different methods can be viewed as diverse solutions to a common problem and emphasize the importance of including diverse perspectives and expertise in the field. We believe a deeper understanding of human communication is crucial to developing machines that can accurately use natural language in human-machine interactions.
\end{abstract}


\clearpage
\tableofcontents
\clearpage

\section{Introduction}
\label{sec:intro}

In the initial phase of AI research following the second AI winter, the focus was on identifying new areas where AI could outperform humans, with famous examples including chess~\citep{silver2017mastering}, Go~\citep{silver2016mastering}, and Starcraft~\citep{vinyals2019grandmaster}. While this was a limited application to games, it set the tone for research to prioritize building AI agents with superhuman capabilities. However, over the last decade, the research community has witnessed a shift towards a human-centric approach that aims to leverage AI to aid humans in everyday tasks and relieve them of repetitive duties~\citep{3328485,hbe2,shneiderman2022human}.

The interaction between humans and machines is a crucial aspect of human-centric AI~\citep{Mikolov2016}, and it should take place in domains where humans are already familiar and require little to no training. Therefore, applications that involve niche practices, such as coding and mathematics, should be avoided in favor of language-based applications. In particular, human-machine communication should be grounded in natural language, which presents the challenge of teaching artificial agents to communicate in multiple languages. Recent advances in natural language processing (NLP) have led to the emergence of the transformer architecture~\citep{vaswani2017attention}, which has become the preferred approach for language-based applications, as exemplified by Language Models (LMs) such as GPT3~\citep{Brown2020}, LLaMA~\citep{abs-2302-13971}, and Lamda~\citep{thoppilan2022lamda}.

One of the challenges for language model architectures is their focus on predicting the next word in a sentence rather than comprehending the broader context and purpose of language usage. While humans use language as a tool for coordination and communication to thrive in a shared environment, artificial intelligence may struggle to understand the subtleties and complexities of language fully.

\subsection{Emergent Communication}

\citet{Linzen2020} investigated the phenomenon of learning misalignment in addressing challenges associated with natural language processing. This phenomenon has been a driving force behind the development of the Emergent Communication (\emecom) field~\citep{Wagner2003}.
To explain \emecom, we refer to \citet{king2009emergent}, where the author defines an emergent communication strategy as a \textit{communication construct derived from the interaction between reader/hearer response, situated context, and discursive patterns}. As the author states, the definition is derived from a plethora of works spanning from business strategy to emergence in organizational complexity theory and other communication theories. Building on the previous  definition, we define \emecom as the research area involving \textit{learning to communicate by interacting with other agents to solve collaborative tasks in complex and diverse environments}.

While \emecom has various applications~\citep{boldt2022recommendations}, it has found widespread use in artificial settings where researchers are interested in improving team-play~\citep{jaderberg2019human}, building architectures of neural networks~\citep{liu2021discrete}, and generalization~\citep{lake2018generalization,jiang2019language}. Additionally, \emecom's techniques has been applied to study the fields of language evolution~\citep{selten2007emergence,winters2015languages,raviv2019larger}, language development~\citep{Rączaszeklang} and language acquisition \citep{Graesser2019, Li2019,abs-1912-05676}.

\paragraph{General description}
All these fields focus on the interaction between agents, whether artificial or human. These interactions typically occur in a specific context, referred to as an environment, defined by a set of rules that mimic certain aspects of the real world. Agents can perceive the environment through \textit{observation} and alter their states by performing specific \textit{actions}.

Two fields heavily influence this pipeline in artificial intelligence, Multi-Agent Systems (MAS) and Reinforcement Learning (RL), which have shown promise in mimicking aspects of human society and mind in AI systems \citep{YangYBWZW18}. MAS techniques can model social interactions and coordination, while RL captures human learning and decision-making aspects. By merging these two approaches, researchers in \emecom aim to create environments that more effectively replicate the complexity and adaptability of human-like behavior.

These environments often incorporate game mechanics, which have been shown to significantly impact the learning process in the animal kingdom~\citep{Spinka2001}, the development of social skills in children~\citep{Tahmores2011}, and for educational purposes~\citep{Kirriemuir2004}. Furthermore, RL's primary application is in games, making \emecom heavily reliant on game frameworks.

\begin{figure}
    \centering
    \includegraphics[width=\linewidth]{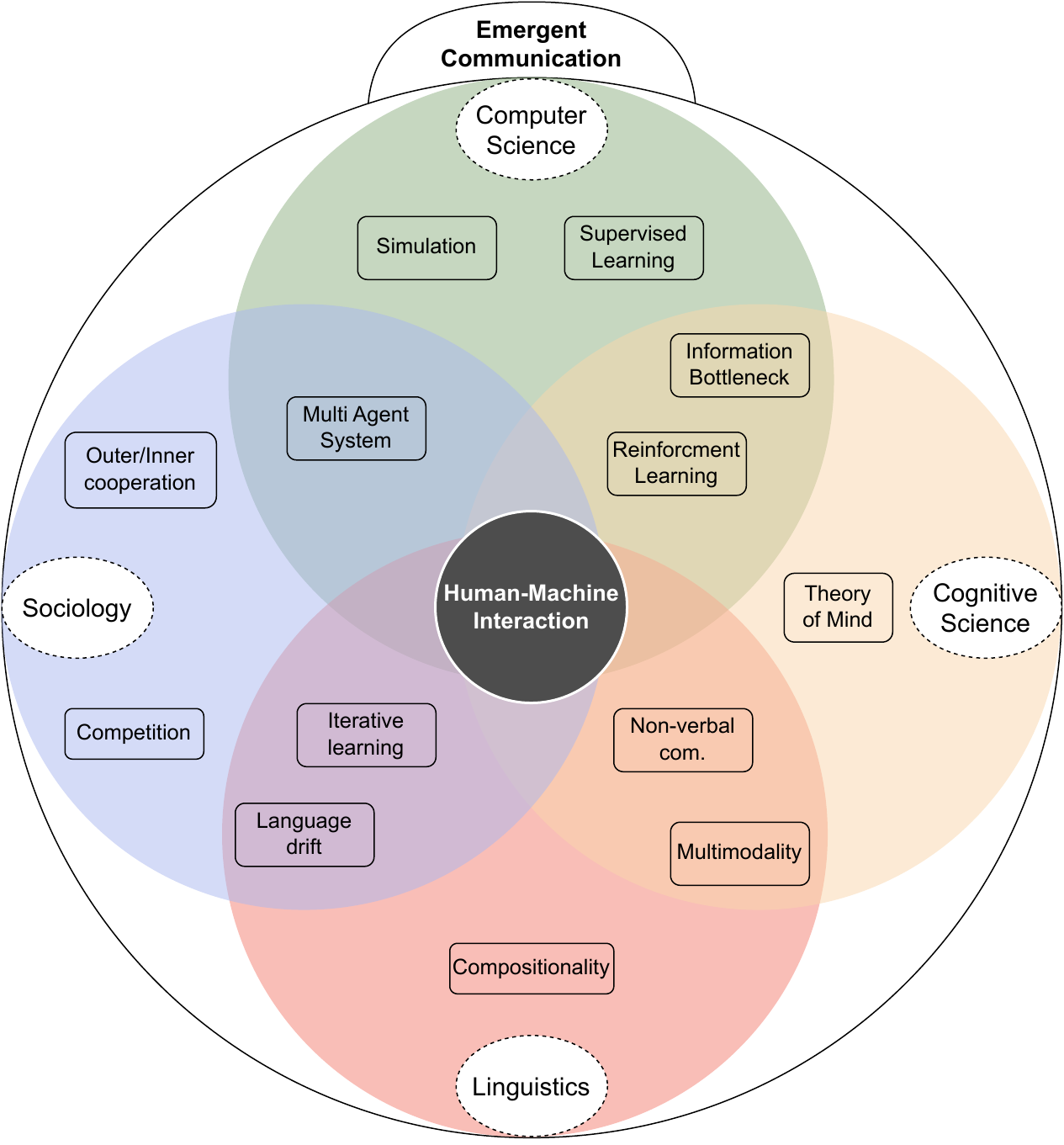}
    \caption{Exploring the multidisciplinary nature of Emergent Communication: A Venn Diagram showcasing the intersections between Linguistics, Cognitive Science, Computer Science, and Sociology. Each field contributes unique characteristics to the study of \emecom (shown in the figure as encompassing the other fields), with some commonalities across multiple fields. At the center of our analysis lies the crucial area of Human-Machine Interaction.}
    \label{fig:prop}
\end{figure}

\paragraph{Objectives}

As mentioned, Emergent Communication (\emecom) intersects with numerous other fields, as illustrated in Figure \ref{fig:prop}. Consequently, various objectives may arise depending on the specific research question being addressed. In this regard, \emecom can be considered more as a framework than a field in itself, and as with many other scientific disciplines, it is challenging to delineate precise boundaries.

Given the involvement of multiple fields and the predominance of Deep Neural Networks (DNN) works in the literature~\citep{abs-2006-02419}, it is the case that there are numerous potential research questions. However, one general overarching objective can be identified as \textit{the development of an artificial language that evolves to resemble human language within an artificial setting} \citep{LazaridouPB16a}. Despite this broad objective, several questions emerge, such as: How should the similarity between artificial and human languages be defined? Which characteristics of human language do we want to develop? And how should the artificial setting be structured?

In this work, we attempt to address these questions and more by concentrating on a sub-objective of \emecom, namely human-machine interaction. Indeed, when the broader goal is to create artificial agents that communicate using a language similar to humans, these agents can be employed for interactions with humans. As a result, this literature review primarily focuses on this perspective, although we occasionally cover other aspects and goals of \emecom as well.

\subsection{Methodology}
In conducting this review, we employed a two-stage methodology for selecting the papers included in our analysis. During the initial phase of our study, we adopted an exploratory approach, navigating through the emergent communication literature by following the references cited in the papers we read and identifying prominent names in the field. While this approach may not adhere to a strict methodology, it facilitated a broad understanding of the state of the art.

To complement and refine our initial selection, we used \textit{Connected Papers} \citep{connected_papers} during the second stage of our methodology. This tool allowed us to visualize the network of references and connections between the works we had already covered, identifying the most interconnected and influential papers. Moreover, it enabled us to spot additional influential publications not encountered in the first phase, ensuring a more comprehensive and robust field review.

\subsubsection{Contribution}

While the following review aims to be accessible to a broad audience, it is worth noting that a background in reinforcement learning and natural language processing  is recommended to comprehend the technical details fully.  With that being said, our goal is twofold:
\begin{itemize}
    \item First, we address new researchers interested in this new and exciting field. Our work can be seen as an introduction to \emecom, where we include relevant literature from the past years and give a general overview of challenges and methodologies.
    \item Second, we address researchers already involved in the field. We connect several pieces of work apparently unrelated to each other under the broader umbrella of human-machine interaction. We advocate that \emecom would benefit from more interconnection with other fields such as linguistics, cognitive science, and sociology, Figure~\ref{fig:prop}. By all means, this benefit would also be reciprocal. 
\end{itemize}

\subsubsection{Paper Structure}

This review is structured as follows. In Section~\ref{sec:proprieties}, we identify four common characteristics in the \emecom literature and point out their parallels to human interactions: the \textit{game environment}, Sec.~\ref{ssec:prop:env}; the 
\textit{learning paradigm}, Sec.~\ref{ssec:prop:learning}, analyzing the different learning methodologies that can be found in \emecom; \textit{interaction types}, Sec.~\ref{ssec:prop:interaction}, defining the possible configuration between agents in a shared environment; and \textit{Theory of Mind}, Sec.~\ref{ssec:prop:tom}, where agents are aware of other intelligent entities in the environment and actively try to model their cognitive states.

Next, we introduce two main categories of \emecom being: Machine-centered \emecom in Section~\ref{ssec:methods:symbolic}, dealing with artificial emergent languages (AELs) through disentangled pre-defined representations, and Human-centered \emecom, in Section~\ref{ssec:natural}, whose characteristic is to use Human Natural Language (HNL), e.g. English, in artificial settings. These sections also provide a large collection of related works from various researchers and an analysis of the state of the art.

Finally, in Section~\ref{sec:conclusion}, we carry out a brief summary and provide a complete table of referenced papers, Table~\ref{tab:list}, which includes the categories each paper falls under, corresponding to the proprieties discussed throughout this work.

\section{Common Proprieties}
\label{sec:proprieties}
In Section~\ref{sec:intro}, we mentioned how \emecom spans across multiple fields with different characteristics, each concerned with a different aspect of human communication. 
In this section, we define a set of intrinsic proprieties of the examined literature and point out the connections to diverse aspects of human interactions.

\subsection{Game environment}
\label{ssec:prop:env}
Game design is an essential aspect of emergent communication research, as it defines the environment in which agents interact and communicate with each other. The literature presents a multitude of game environments, each tailored to address specific research questions. In this review, we identify two fundamental properties that feature prominently in the design of game environments used in emergent communication studies.

Firstly, the role of communication is a crucial aspect of game design and can take on varying degrees of importance depending on the research objectives. Communication can either serve as the game's primary objective or as an auxiliary tool to help agents achieve other goals. For instance, in some studies, the focus is on the emergence of a communication protocol between agents, while in others, agents are tasked with coordinating their actions to accomplish a shared goal, and communication merely aids in their cooperation and coordination.

Secondly, the choice of input representation is another critical aspect of game design that can significantly impact the emergence of language. In particular, how information is represented and presented to agents can influence the type of language that emerges from their communication. For example, representing images as raw pixels or as a bag of attributes can affect the level of ambiguity in the emergent language and the agents' performance in completing the task.

\subsubsection{Role of Communication}
\label{sssec:prop:env:comm}
Communication can be either the primary goal or a supporting feature in \emecom, which we categorize as either \textit{Communication-focused} or \textit{Communication-assisted}. It's worth noting that \textit{Communication-focused} games are a subset of \textit{Communication-assisted} settings, as the latter requires non-communicative actions in addition to communicative ones.

\paragraph{Communication-focused}
In the first category of communication games, communication functions as both the method employed by agents and the ultimate goal of the study. Although this may seem like a simplified version of human language, it is observed in the animal world, as seen in the referential gestures exhibited by ravens~\citep{Allee1949}.

\subparagraph{Referential Games}

\begin{figure}
    \centering
    \includegraphics[width=\linewidth]{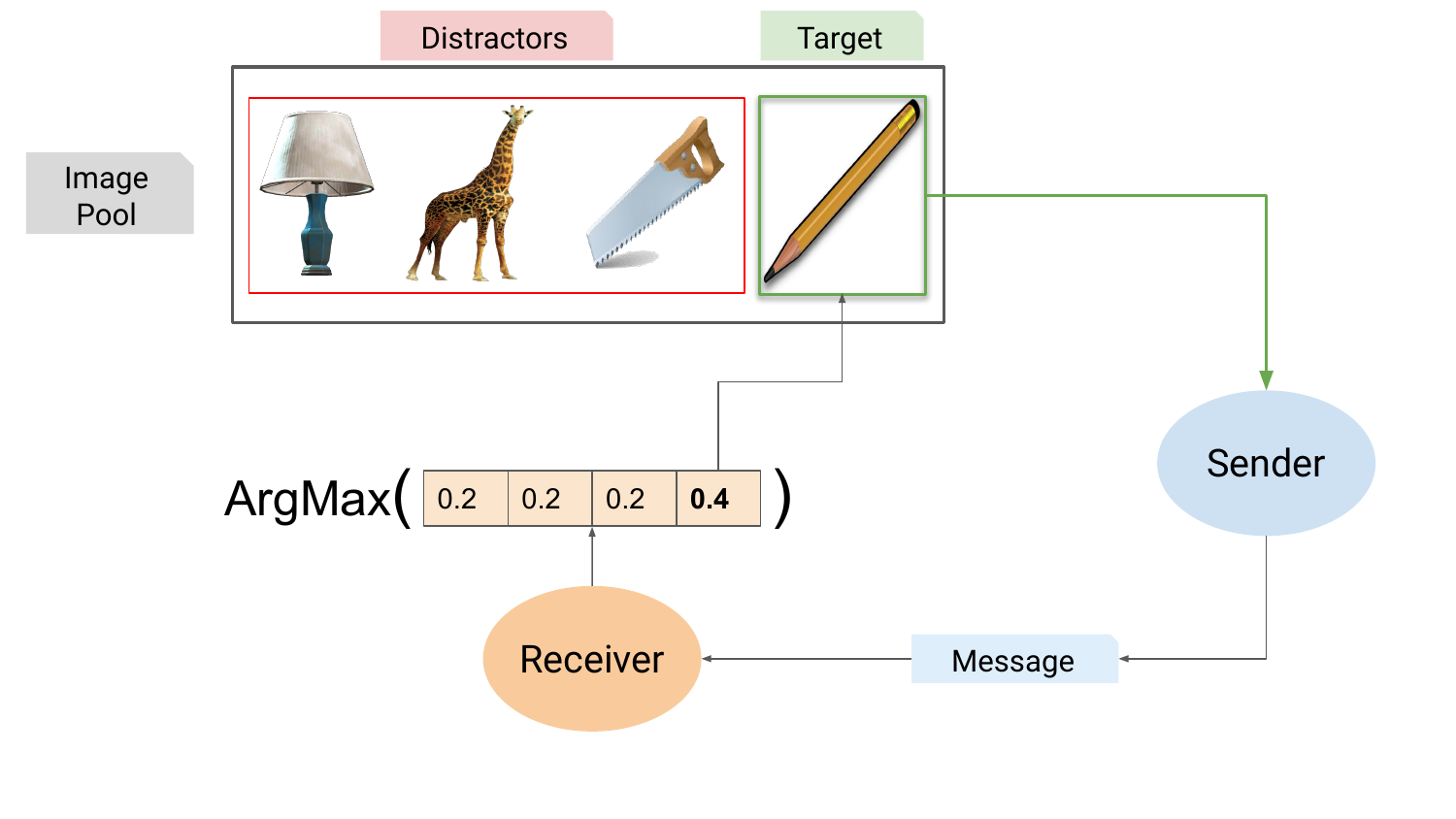}
    \caption{General pipeline for a discriminative referential game. 
    The sender is shown a \textit{target} image (a pencil) and is tasked to generate a \textit{message}. The receiver sees a pool of images (\textit{distractors}) containing the target and must choose the correct one based on the message.}
    \label{fig:ref}
\end{figure}

The most well-known game in this category is the discrimination Referential game (\textit{Ref}), Figure~\ref{fig:ref}, first introduced by \citet{Lewis2008}. This game involves a sender and a receiver with distinct roles, each presented with a set of images. The sender must generate a message to describe a given image, which the receiver must then identify from a pool of images. The first instance in the RL community can be found in~\citet{Das2017}, which compares a synthetic world made up of primitive geometries to a real-world image using a visual dialog system.
Although there are several variations of the discrimination referential game, as illustrated in Table~\ref{tab:ref_types}, the primary focus of research has been analyzing the language that emerges from it. This has been demonstrated in various studies, including \citep{abs-2001-07752,Rodriguez2019,Graesser2019,Li2019,abs-2001-03361,Chaabouni2020,Havrylov2017,Lazaridou2016,abs-2110-05422}.

\begin{table}
\centering
\begin{tabular}{|l|m{9cm}|}
\hline
Type of Ref. game & Description 
\\ \hline
Discrimination   & The receiver has to discriminate between a set of stimuli, comprised of the target stimulus observed by the sender/speaker and some additional distractor stimuli, and find the target.
\\\rule{0pt}{1cm}    

Generative       & The receiver has to generate an output, which can for instance, be the task of reconstructing the target stimulus itself or some of its (symbolic) attributes. 
\\\rule{0pt}{1cm}    

Multi-step       & The game's rules are the same as the discrimination game, but the receiver can choose to ask the sender for more information which will start another step in the game. The game ends when the receiver chooses a target or after a fixed number of steps.
\\\rule{0pt}{1cm}    

Multi-modal      & The sender and receiver have access to different modalities: usually either vision or textual information.  
\\   

\hline
\end{tabular}
\caption{Different types of referential games.}
\label{tab:ref_types}
\end{table}

\subparagraph{Task and Talk}
In contrast, \citet{Kottur2017} developed a question-answering game called Task and Talk (\textit{TnT}) where a sender bot is given an object\footnote{In this case, the object is seen as a set of symbolic representation through disentangled attribute-values.} unseen by the receiver. The receiver must ask the sender questions to determine two of its attributes. Unlike \textit{Ref}, the \textit{TnT} architecture is more dynamic and iterative, with a question-and-answer format that more closely resembles human communication. However, this architecture also introduces complications in the training procedure, such as agents retaining memory of previous conversations~\citep{Sally1995}. Another difference is that Kottur et al. and others~\citep{Liang2020,abs-1904-09067} focus on objects defined by predetermined properties rather than perceptually realistic inputs, distinguishing between symbolic and realistic inputs.

\paragraph{Communication-assisted}
Unlike previous \textit{Communication-focused} games, this second category utilizes communication as a means to achieve a goal different from the communicative act itself. The agents' action space includes both communication signals and other game-dependent actions, which range from physic simulators~\citep{Grover2018,Mordatch2018}, navigation tasks~\citep{Das2019,Lowe2017,abs-1912-05676,Chaabouni2019,abs-2107-05697}, negotiation settings~\citep{Bachrach2020,Cao2018,abs-2005-05441} to social deduction games~\citep{Brandizzi2021,Nakamura2016,abs-1810-08647}. These games aim to recreate the environment in which language emerged, emphasizing the view that human language did not emerge as a goal itself but rather as a means of coordinating actions between humans. While \textit{Communication-assisted} games may be considered closer to real human interaction, recent works typically rely on one-shot communication signals rather than dialogue systems.

\subsubsection{Input representation}
\label{sssec:prop:env:input}

Input representation is a crucial aspect of language emergence in artificial communication systems. How information is encoded and presented to agents can significantly affect the type of language that emerges from their communication. In particular, differences in input representations can profoundly impact the learnability and generalizability of emergent languages. In this section, we explore several studies investigating the effects of different input representations on the emergence of language in artificial communication systems. 

The impact of input type on emergent communication in a referential game is significant, as demonstrated by~\citet{Lazaridou2018}. The authors conducted two referential games with different input types: one using symbolic representation, where objects were represented as a bag of attributes, and the other using raw pixels. They observed a high level of ambiguity in the raw pixel input game due to the difficulty of the exploration task. To address this, they developed an experiment where distractors were selected from a target-specific context distribution reflecting normalized object co-occurrence statistics. The results showed an above-random performance in generalization, tied to language compositionality, with a high topographic measure suggesting that similar objects received similar messages. In the raw pixel input experiment, the authors investigated which image attributes were primarily captured by the agent to complete the task. They found that the encoder-decoder architecture overfits the dataset, resulting in an unstable protocol.

Furthermore, other studies have also examined the impact of input representation on emergent communication. For example, \citet{abs-1910-05291} investigate the effect of different input types, such as image representation, a concatenation of one-hot vectors representing the count of each object type, and a bag of one-hot vectors denoting the quantity of different object types. They demonstrate a significant relationship between input design and language learnability, revealing the emergence of compositionality in the first two cases. Similarly, \citet{abs-2001-07752} applies their setups to two different datasets, a symbolic dataset called a \textit{number set} and a \textit{3D Object} dataset, encoded with one-hot-vector and a convolutional neural network, respectively. The latter showed easier learning\footnote{Learnability measures the easiness in learning a language for a new speaker.} and faster convergence, suggesting a reduced possibility space compared to the symbolic dataset.

Additionally, \citet{abs-2012-10776} examines the correlation between the structure of input and generalization abilities in a referential game. They experiment with the number of attributes in the dSprites dataset \citep{HigginsMPBGBML17}, providing results for 2, 3, 4 attributes and visual representations. Their results support the hypothesis advanced by \citet{Chaabouni2020}, that generalization occurs naturally when the input space is large, although they do not experiment with multiple input structures simultaneously.

While human communication in a referential game appears to rely solely on visual information, the human mind can access additional tools when referring to objects in the real world. Human categorization~\citep{anderson1991adaptive, wierzbicka1984apples} plays a crucial role in the semantic structure of language and has been linked to hierarchical relationship attribution between super-ordinate and lower-level categories~\citep{markman1989categorization}. Although the works mentioned above examine the differences between inputs in the form of visual or categorical representations, no effort is made to utilize more input representations at once.

\paragraph{Information bottleneck}

The input space plays a significant role in the emergence of language, as demonstrated by previous studies. When the input space is large enough, agents tend to develop a more generalized language. This aspect relates to the human mind and language, as discussed by ~\citet{zaslavsky2018efficient}. They argue that language is used to compress ideas into words efficiently, and this compression involves a trade-off between lexical complexity and accuracy. The authors conducted a color-naming game across human participants and demonstrated that languages achieve near-optimal efficiency based on the information bottleneck principle\footnote{The information bottleneck method is a technique aimed at finding the optimal balance between accuracy and compression when clustering or summarizing a random variable \textit{X} based on a joint probability distribution $p(X,Y)$ between \textit{X} and an observed relevant variable \textit{Y}. } (IB)~\citep{physics-0004057}.
The information bottleneck effect has been studied in multi-agent communication with message pruning~\citep{abs-1912-05304}, limited bandwidth~\citep{abs-1911-06992}, and message entropy~\citep{KharitonovCBB20}; however, few studies have taken into account its evolutionary advantage.

A study by \citet{kirby2015compression} explores the same line of work on human participants, where simulated rational learners were tested to validate the trade-off between expressiveness and compressibility under different constraints. Simulating cultural evolution\footnote{A setting where population and iterative learning are merged, see Section~\ref{ssec:prop:interaction}.}, they proved how a hierarchical organization of language emerges where learners experience pressure on both the learning and communication side. 

Similarly, \citet{Kottur2017} carried out a \textit{TnT} reference game involving two agents and investigated the necessary constraints for a generalized language to emerge. They discovered that a limited vocabulary size and memory-less models fostered the development of a language in which individual symbols were grounded in attributes. 

According to \citet{Resnick2019}, there is a connection between learnability, capacity, bandwidth, and the use of structured language for language learning. They hypothesize that learning compositional communication requires less capacity than learning a non-compositional code, shedding new light on the problem  of artificial language learning. Unlike other works that advocate for deeper and larger architectures to emulate the human brain, Resnick et al. identify the problem as the overpowering ability of machines to memorize input spaces. 

In conclusion, the input space and the trade-off between informativeness and complexity are essential for language emergence. A recent study by \citet{NEURIPS2022_8bb5f663} introduces the Vector-Quantized Variational Information Bottleneck (VQ-VIB) method, which combines task-specific utility maximization with general communicative constraints. VQ-VIB agents can adapt to changing communicative needs, develop meaningful embedding spaces, and demonstrate improved utility and faster convergence rates. This framework offers new insights into human language evolution and artificial emergent communication, paving the way for future research in complex domains and human-agent interactions. 

\subsubsection{Open Challenges}

Drawing from the studies discussed thus far, particularly those by \citet{Resnick2019} and \citet{Kottur2017}, we observe that artificial agents possess exceptional memory capabilities.
These capacities enable agents to discover communicative shortcuts, resulting in degenerate languages that are more akin to holistic (one-to-one mapping-like) languages rather than compositional ones. A common solution emerging in the literature is to increase the environmental complexity, i.e., the input space, towards more human-like environments. Indeed, the majority of approaches discussed in this and subsequent sections primarily focus on implementing more challenging environments rather than examining sub-optimal architectures. 

Nevertheless, enhancing the environmental complexity and limiting an agent's communication capacity are complementary solutions to deter machines from excessive memorization. Although the former is a valid approach, it introduces supplementary variables into the system, which must be considered when analyzing the resultant language. We argue that researchers in this field should emphasize preventing memorization from both perspectives, giving increased attention to the utilization of smaller, more manageable neural networks that necessitate generalization in order to solve the task.

Furthermore, as mentioned in the introduction, this review focuses on the improvement of human-machine communication. To achieve this goal, developing machines with similar generalization requirements could be beneficial. An intriguing research question arises: how can we design learning paradigms and neural networks that inherently favor generalization over memorization? 

From this inquiry, we can also consider whether merely expanding the input size is sufficient to achieve generalization. This may be explored by drawing parallels with human learning in relatively simple tasks: do humans resort to memorization when possible, or is generalization an inherent aspect of human learning?

Addressing these questions may help researchers devise strategies to encourage generalization over memorization in artificial agents, leading to the emergence of more structured and organized languages that better reflect human-like learning processes.

\subsection{Learning paradigm}
\label{ssec:prop:learning}

In \emecom research, exploring the vast space of possible communication utterances, or state space, can be too complex for simple networks to handle. Deep neural networks have become a popular choice for training agents on a discrete set of symbols for communication. To train \textit{DNN}, multiple learning frameworks can be deployed, the main ones being \textit{reinforcement} and \textit{supervised} learning.

\subsubsection{Reinforcement Learning}
\label{sssec:prop:learning:rl}

Reinforcement learning is a crucial approach in the field of emergent communication for two main reasons. First, it allows artificial agents to learn from interactions with a game environment, which mirrors the human ability to adapt to changing circumstances. In this context, it is essential to differentiate between single-agent and multi-agent reinforcement learning, as the latter introduces additional complexities arising from the need for coordination and communication among multiple agents.

The second  motivation for utilizing reinforcement learning in \emecom is its ability to facilitate backpropagation through symbols.. Indeed, training agents to communicate through a set of discrete symbols presents a challenge: backpropagation is difficult due to the non-differentiability of the variables involved. This problem arises specifically because researchers aim to model human language on a discrete, predominantly word-based level, which captures the essential structure and characteristics of natural language.

To overcome this, various techniques have been developed, including the reparameterization trick (such as VQ-VIB \citep{NEURIPS2022_8bb5f663}), semantic hashing~\citep{SalakhutdinovH09,abs-1801-09797}, Gumbel Soft-max~\citep{Jang2016,Maddison2016} and the REINFORCE algorithm. These methods enable backpropagation through non-differentiable variables, allowing for effective training of communication networks

\paragraph{REINFORCE}

The REINFORCE algorithm~\citep{williams1992simple} is a well-known method for estimating loss function gradients with respect to stochastic policy parameters, and it has been used in many works in the field of \emecom to backpropagate through symbols, as shown in Table~\ref{tab:list}. Its simplicity and effectiveness have made it a popular choice, but it can suffer from high variance and instability during training. To mitigate these issues, various modifications have been proposed, such as the use of baselines \citep{MnihG14,GuLSM15}.

\begin{figure}[t]
    \centering
    \includegraphics[width=\linewidth]{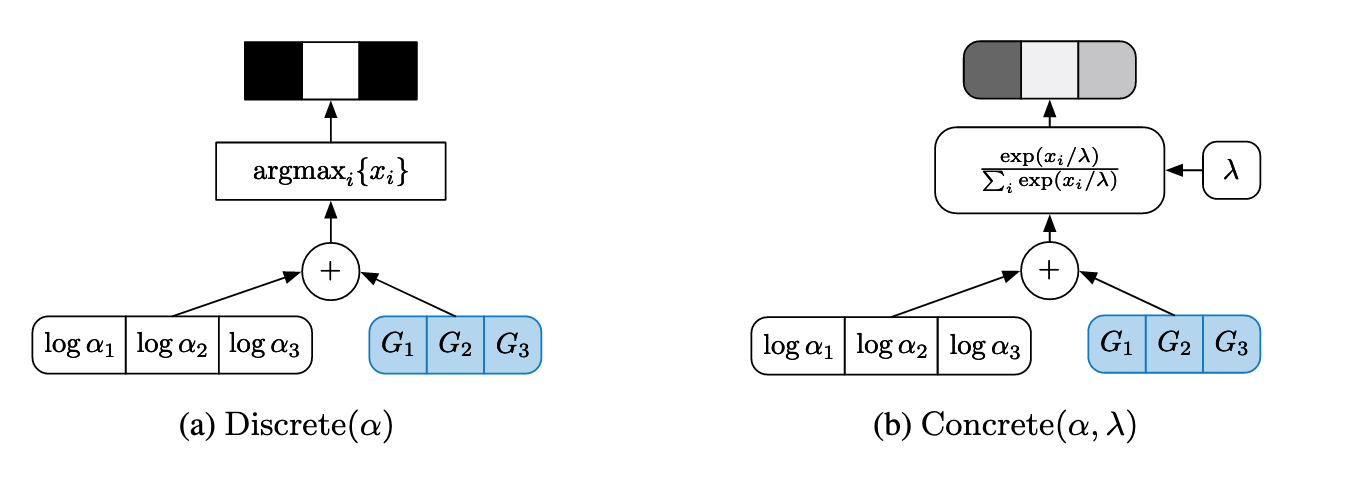}
    \caption{Visualization of sampling graphs for 3-ary discrete $D \sim Discrete(\alpha)$ and 3-ary Concrete $X \sim Concrete(\alpha, \lambda)$. White operations are deterministic, blue are stochastic, rounded are continuous, square discrete. The top node is an example state; brightness indicates a value in [0,1].\\
    Image and caption taken from \citet{Maddison2016}.}
    \label{fig:concrete}
\end{figure}

\paragraph{Gumbel Soft-max \& Concrete distribution}
As noted earlier, previous research on stochastic gradient estimation has primarily focused on addressing the high variance issues of the REINFORCE algorithm by augmenting it with Monte Carlo variance reduction techniques or biased path derivative estimators for Bernoulli variables~\citep{BengioLC13}. Until the recent introduction of the Gumbel Soft-max and Concrete distributions~\citep{Jang2016,Maddison2016}, no gradient estimator had been specifically designed for categorical variables to facilitate backpropagation through symbols.
Unlike one-hot encoding of categories, which does not provide a gradient, the Gumbel Soft-max and Concrete distributions provide a continuous relaxation of the categorical distribution, with a noise component, as shown in Figure~\ref{fig:concrete}. Additionally, the relaxation intensity can be regulated with the parameter $\lambda$, where the Concrete distribution converges to the categorical distribution with $\lambda$ approaching zero.

On one hand, this kind of relaxation has been reported to achieve better performance than REINFORCE in several works~\citep{Havrylov2017,Omidshafiei2019,Chaabouni2020,abs-2107-05697,abs-2204-12982,abs-2005-05441,KharitonovCBB20}. For instance, \citet{Havrylov2017} demonstrated a positive correlation between message length and communication protocol convergence speed, not observed in REINFORCE-like algorithms. They also combined REINFORCE and Gumbel-softmax, and show improved loss function gradient estimation and game setting differentiability, resulting in a structured, hierarchical encoding scheme with faster convergence than standard RL frameworks.

On the other hand, one could question whether allowing the gradient to flow directly through the communication channel bears any resemblance to human cognition. Studies in cognitive science and neuroscience have suggested that human communication and learning processes might not rely on such direct optimization methods, but rather on more intricate processes, such as episodic memory \citep{buzsaki2018space}, analogy \citep{gentner2001analogical}, and social learning \citep{rumjaun2020social}. As a result, the extent to which these relaxation techniques can inform our understanding of human language evolution and cognition remains an open question.

\subsubsection{Supervised Learning}
\label{sssec:prop:learning:sl}

Supervised learning (SL) is another common learning framework used in \emecom. Unlike RL, SL uses labeled data. In communication games with human HNL (usually English), supervised learning can be used for tasks such as language pre-training~\citep{Das2017,Cogswell2020,Li2020}, distribution shift mitigation~\citep{abs-2003-12694,HawkinsKSG20,abs-2010-02975}, and visual classification~\citep{Lazaridou2016,Lazaridou2020}. However, in games with symbolic languages, SL is rarely used, with some exceptions such as the modified reference game presented in~\citet{Graesser2019}, where RL is used to coordinate agents through a communication signal, while SL is used to estimate a probability distribution for captions. 

Supervised learning is also used in works that incorporate Theory of Mind, Section~\ref{ssec:prop:tom}, which equips agents with a prediction module to estimate other agents' beliefs and future actions. In these cases, SL can be used to predict actions given current observations~\citep{abs-1810-08647,Raileanu2018,JaquesLHGOSLF19,Rodriguez2019} or coupled with the obverter technique to influence policy based on an agent's own understanding~\citep{abs-1804-02341,abs-1809-00549}.

\subsubsection{Open challenges}
\label{sssec:prop:learning:op}

The field of \emecom involves a diverse range of learning techniques. Given the interactive and game-like nature of the environment, reinforcement learning plays a central role, as can be seen from Table \ref{tab:list}. RL enables the emergence of novel behaviors by exploring and exploiting the system's intrinsic dynamics. Consequently, \emecom draws heavily on human experience, specifically on our understanding of dynamic real-world systems and our ability to adapt to them. 

Supervised learning is also utilized in certain cases, particularly for auxiliary tasks, as it mirrors the human ability to learn through demonstration. The capacity of SL to learn from demonstration complements RL, which can build upon knowledge and experiences gained by previous generations without always starting from scratch (see Iterative Learning in Section \ref{ssec:prop:interaction:time}). A common challenge in these cases is striking the right balance between supervised language tasks and reinforcement-based referential games, as discussed in Section \ref{sssec:method:natural:balance}.

To improve human-machine communication, bridging the gap between human and machine learning paradigms may be a promising approach. This requires the integration of multiple learning aspects within a single framework as well as exploring new learning paradigms. However, in the literature analyzed for this review, a surprisingly limited number of studies explore alternative learning approaches, such as unsupervised learning~\citep{Grover2018,abs-1912-05304}, evolution strategies~\citep{abs-2001-03361}, stochastic computational graphs~\citep{KharitonovCBB20} and self-supervised learning \citep{abs-2106-04258}.

Moving forward, future research should prioritize investigating these alternative learning methods and their potential synergies with existing paradigms. By expanding the scope of learning techniques employed in the field, researchers may uncover novel strategies that more effectively mimic human learning processes, ultimately leading to enhanced human-machine communication and collaboration. Additionally, interdisciplinary approaches that draw from cognitive science, sociology, and linguistics may provide valuable insights to inform the development of more human-like learning algorithms.

\subsection{Interaction type}
\label{ssec:prop:interaction}
In this section, we categorize the types of interactions among agents and study their contributions.

We differentiate between inner and outer interaction types.
Given a set $A_N$ of $N$ agents, we define a team $T_m \subseteq A_N$ as a subset of agents and we denote with $\{ T_1, \ldots T_k \}$ a partition of all the agents in $k$ disjoint teams with $\sum_{i=1,\ldots,k} T_i  = N$.

Given a partition of agents (i.e., a set of teams), we define a game to be cooperative when any negotiation between teams leads to  a sub-optimal outcome $S_{so}$ for all the parties involved. While each team actively tries to reach an optimal state $S_o$, through negotiation, it can avoid the worst outcome (pessimistic) $S_p$ where $S_o < S_{so} < S_p$.
On the other hand, a zero-sum game competitive setting always implies the worst outcome for all teams but one. In the latter, negotiation becomes ineffective, and each team must prevail over the others.

\begin{figure}
    \centering
    \includegraphics[width=\linewidth]{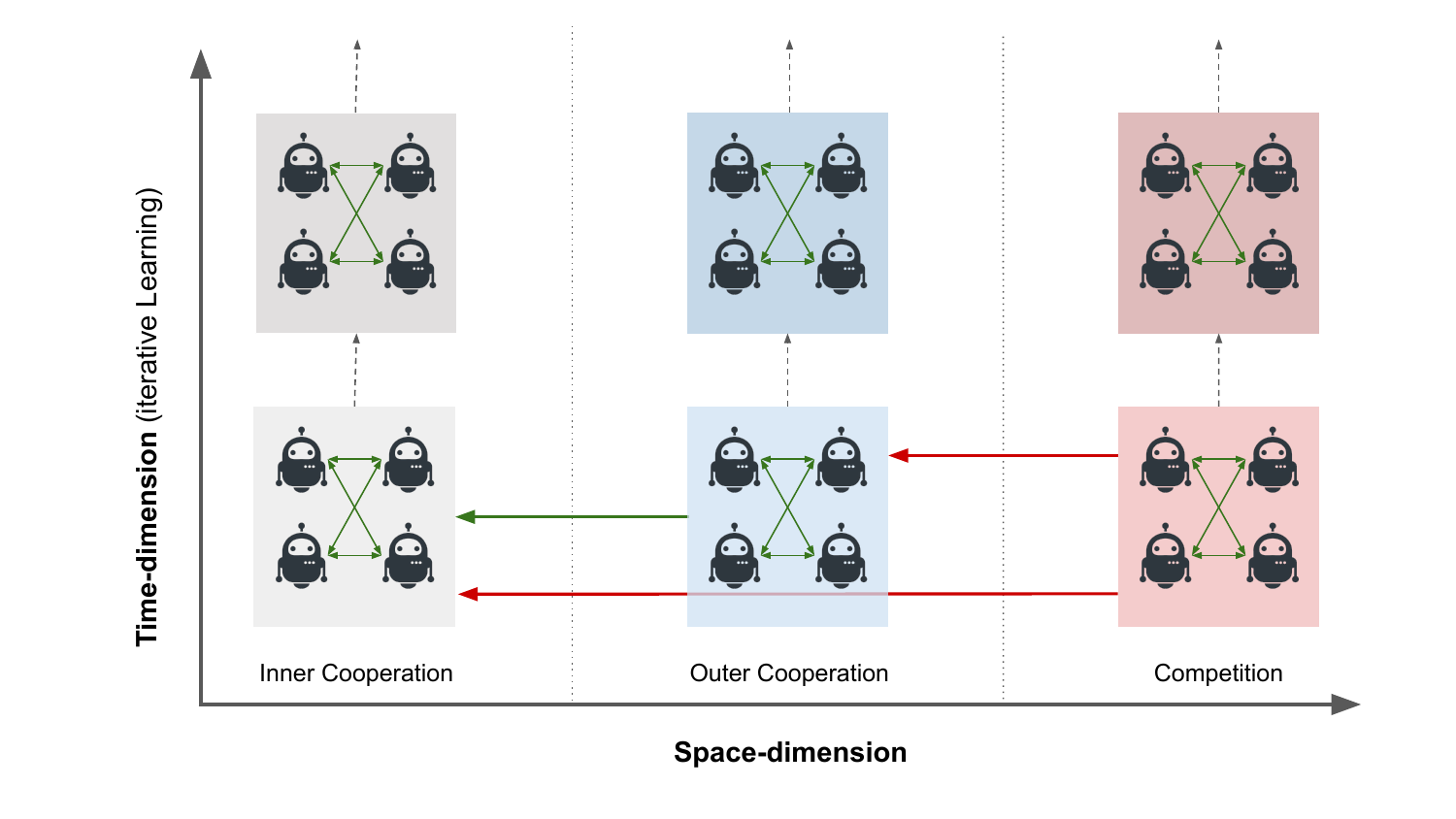}
    \caption{Visualization of interaction types in Emergent Communication, with a horizontal space dimension and a vertical time dimension. The horizontal dimension is split into three parts for inner, outer cooperation, and outer competition. Teams are represented by squares, and their interconnections are indicated by arrows of different colors: green for cooperative, red for competitive, and gray dotted lines for time.}
    \label{fig:interaction_type}
\end{figure}

In the following section, we introduce three categories that focus on the spatial aspect of human interaction, distinguished by the number of agents or teams involved. However, humans also engage in interactions across the temporal dimension, not only through the transfer of information and experiences from generation to generation but also in turn-taking during conversations. This temporal aspect leads to a type of interaction known as iterative learning, which contrasts with the spatially-oriented categories. To visually represent these differences, Figure~\ref{fig:interaction_type} illustrates the various interaction types across both spatial and temporal dimensions, emphasizing the need to consider both aspects in \emecom research.

\subsubsection{Space-oriented}
\label{ssec:prop:interaction:space}

We define \textit{inner cooperation} as the setting where all the agents of one team are cooperative with one another. 
On the other hand, \textit{outer cooperation} is defined as two or more disjoint teams of agents sharing a common goal, thus being incentivized to cooperate. 
Differently, \textit{competition} arises in zero-sum games, where one team's victory implies a pessimal outcome for all other teams.
In these cases, negotiation is not useful.

\paragraph{Inner Cooperation}
Cooperation is crucial for communication to arise~\citep{Smith2010, Nowak1999}. As seen in Table~\ref{tab:list}, most works focus on inner cooperation. For instance, \citet{Lazaridou2016} demonstrate the necessity of cooperation in referential games for successful communication. 
Moreover, \citet{Cao2018} studies how pro-social agents favor cheap talks and achieve better results than selfish ones. On the same line, \citet{Graesser2019} suggests how intricate language evolution can emerge from simple social interactions between agents. 

\paragraph{Outer Cooperation}

Outer cooperation in emergent communication can occur in two different ways. The first is the standard instance of outer cooperation where disjoint teams collaborate by interacting with the environment and each other, as explored in various works~\citep{Evtimova2017,Lowe2017,Bachrach2020,abs-2005-05441}. 
A distinctive characteristic of this method is that every agent is initialized at the same time and has no prior experience of the environment.
In contrast, population learning involves training teams together before introducing them to other teams.

This setting is influenced by linguistics and sociology, specifically by the study of language development in different cultures~\citep{Briscoe2002,Kirby2014}.
Studies have shown that a population of agents generalizes better than a pair~\citep{abs-1912-06208} and how the amount of group connectivity determines the evolution of mutually intelligible languages~\citep{Graesser2019}. Moreover, the interaction between populations of agents can lead to the emergence of a language that is easier to teach and understand~\citep{Li2019, Lowe2019,Fitzgerald2019,abs-2107-05697,abs-2110-05422,abs-1904-09067}.

\paragraph{Competition}

Competition can lead to improved communication protocols and general performance, as shown by \citet{Liang2020}. They suggest that competition among agents can prioritize compositionality, performance, and convergence in communication protocols. Similarly, \citet{Nakamura2016} set up a social deduction game\footnote{The Werewolves of Millers Hollow.} where agents must infer the trustworthiness of others through interactions and hard-coded communication actions. \citet{Brandizzi2021} also study this setup, but in a canonical \emecom environment without communication constraints. Despite highly adverse scenarios, such as incomplete information and strategic deception, the authors show that communicating agents can still defeat opponents through effective communication and collaboration.

\subsubsection{Time-oriented}
\label{ssec:prop:interaction:time}

Notably, the first three categories focus on the spatial component of human connection, using the number of agents/teams to characterize their differences. However, humans exist in a temporal dimension where information and experiences are passed from  generation to generation, leading to a type of interaction called iterative learning, which is diametrically opposed to those previously addressed.

\paragraph{Iterative Learning}
The concept of iterative (or iterated) learning is closely linked to population learning and it is heavily inspired by the field of language development and evolution. Iterative learning occurs when a population of agents passes on their knowledge to a new one, repeating indefinitely. 

In linguistics, iterative learning is compared to a bottleneck in the learning system, which enables generalization~\citep{kirby2001spontaneous,Nowak1999,ScottPhillips2010}. Studies such as~\citet{Kirby2014} and~\citet{Graesser2019} have explored iterative learning's role in the emergence of natural language, demonstrating that it can shift languages towards consistency with prior biases and lead to language conversion to a lower complexity majority protocol with linguistic contact over time.

The transmission of language across generations requires older agents to teach younger ones. To investigate the learnability and generalization in referential games, several works have focused on the aspect of teaching. \citet{Li2019} have implemented referential games in which old agents are periodically swapped with new ones and receiver have their parameter reset periodically. \citet{abs-2002-01365} trains new agents on data generated by older agents. Both of these approaches have shown a strong correlation between ease of teaching and the speed of convergence to a generalized language. 

Interestingly, \citet{ZhouVLC22} draws a connection between parameter reset and the lottery ticket hypothesis\footnote{According to the authors:\textit{ dense, randomly-initialized, feed-forward networks contain subnetworks (winning tickets) that, when trained in isolation, reach test accuracy comparable to the original network in a similar number of iterations}.}~\citep{abs-1803-03635}, where the authors regard forgetting as a valuable part of iterative learning and point out its usefulness in language evolution~\citep{barrett2009role}.
 
Focusing more on the teaching aspect, \citet{Omidshafiei2019} define explicit teacher-student roles and demonstrate how their teaching agents not only learn significantly faster but also learn to coordinate in tasks where existing methods fail. 
Interestingly, there is a parallel between teaching in multi-agent communication and multi-agent transfer learning~\citep{SilvaWCS20}, as suggested by~\citet{Omidshafiei2019}. This connection highlights the potential for cross-fertilization between these fields and the importance of further exploring their interconnections. Indeed, reinforcement learning is a commonly used training paradigm for iterative learning, with Self Play~\citep{Tesauro94} being the preferred approach~\citep{Lowe2020,Graesser2019,GuptaLFKP19}. Other approaches include seeded iterative learning~\citep{abs-2010-02975,abs-2003-12694}, and evolution strategies~\citep{abs-2001-03361}.

In conclusion, these works demonstrate how language structures are transmitted across generations of learning agents and refined with each subsequent iteration, resulting in more efficient, communicable, and learnable languages~\citep{abs-2002-01365,abs-1912-06208,Chaabouni2019,abs-1904-09067}.

\subsubsection{Open challenges}

As evident from the literature, there is a considerable focus on iterative and population learning, as they appear to impose the necessary constraints for the emergence of easily learnable languages. However, there is limited work on investigating the effects of competition and, more specifically, on balancing cooperation and competition. Exploring the optimal balance between cooperation and competition in mixed human-robot teams presents a significant challenge that could improve the overall performance and effectiveness of such teams.

Furthermore, non-verbal communication is fundamental in human-human communication, and although slightly beyond the scope of emergent communication, developing artificial agents that can both interpret and produce non-verbal cues remains an open challenge for enhancing human-machine communication.

\begin{figure}[h]
    \centering
    \includegraphics[width=\linewidth]{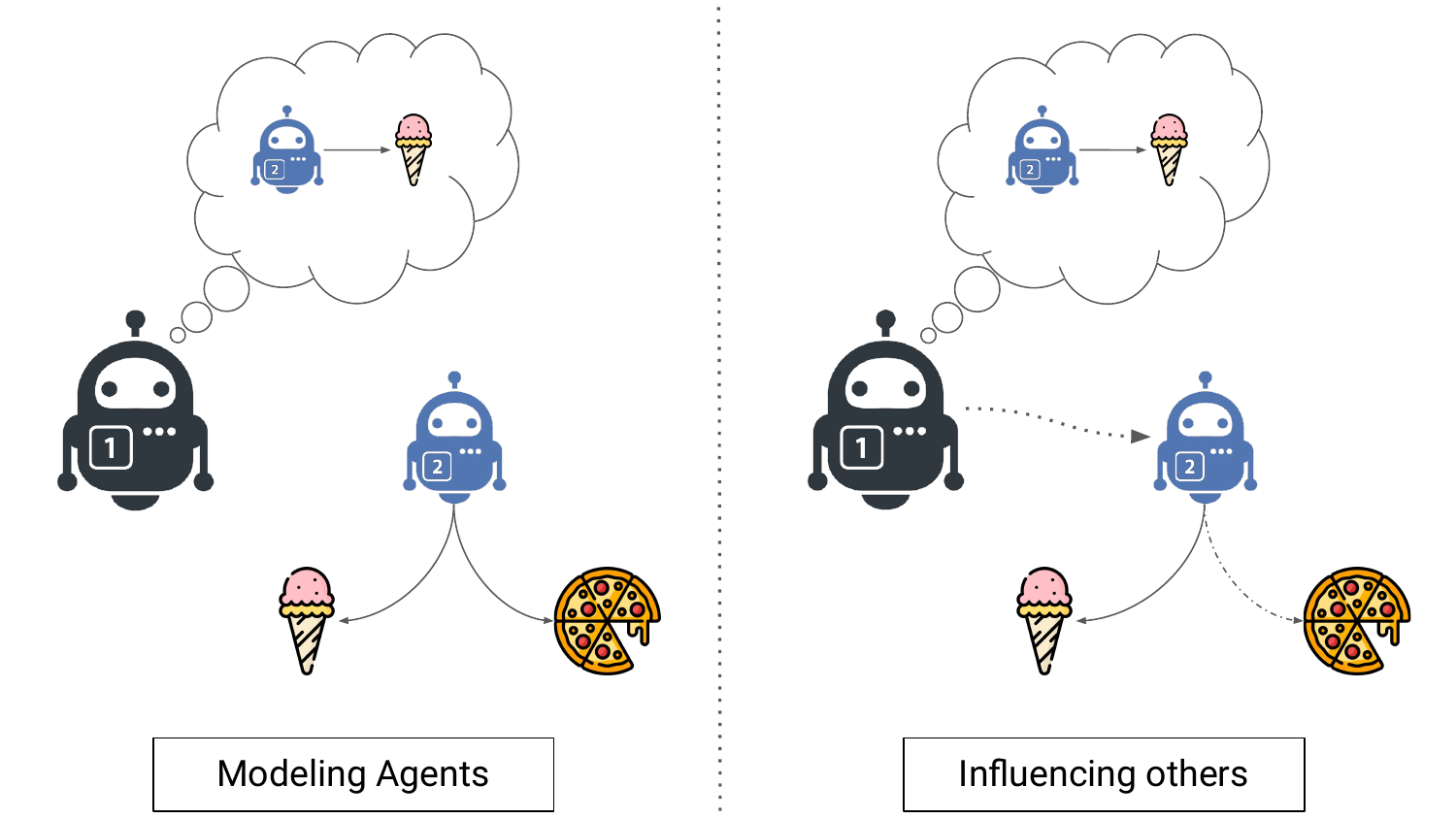}
    \caption{Illustration of Theory of Mind in artificial agents: Agent 2 must choose between pizza and gelato. In the \textit{modeling agents} approach, Agent 1 predicts Agent 2's choice based on their preferences or past behavior. In the \textit{influencing others} approach, Agent 1 takes action to influence Agent 2 to select a specific option.}
    \label{fig:top_type} 
\end{figure}

\subsection{Theory of Mind}
\label{ssec:prop:tom}

The concept of Theory of Mind\footnote{Also known as \textit{opponent's modeling} in the field of RL.} (\textit{ToM}) is a crucial aspect of human behavior that has been modeled in the \emecom field. It refers to our ability to form beliefs about how others might react to certain stimuli and update them with new observations, as shown in various studies~\citep{Gopnik1992, Premack1978}.
More recently, \citet{abs-1802-07740} applied \textit{ToM} to let artificial agents build a model of other agents' observations and behavior.

In \emecom, there are two main approaches to augment agents with \textit{ToM}: (i) \textit{Agent's modeling}, where artificial agents actively model others' behavior to some extent; and (ii) \textit{Influencing others}, where agents manipulate other agents' behavior based on their objectives, an extension of \textit{agent's modeling}, see Figure \ref{fig:top_type}.

\subsubsection{Modeling Agents}
\label{sssec:prop:tom:model}
The concept of modeling other agents is well-established in the field of multi-agent reinforcement learning. However, machines require specific formulations to approximate the same behavior as humans do.

One way to approach this is to leverage the similarities between agents' belief systems. This approach called the \textit{obverter technique}, has been shown to be effective for the emergence of compositional languages~\citep{abs-1804-02341,abs-1809-00549}. Interestingly, the obverter technique bears a striking resemblance to the Rational Speech Act (RSA) \citep{frank2012predicting,goodman2016pragmatic}, a prominent linguistic framework that models how speakers and listeners use reasoning to communicate effectively. This parallel between the obverter technique and RSA further validates the interdisciplinary nature of \emecom, as it demonstrates the potential for cross-pollination between computer science and linguistics.

Alternatively, mental models can also be based on other agents' actions and perceptions without assuming similar belief systems.
For instance, \citet{Raileanu2018} augments agents' policy with predictions of other agents' behavior and demonstrates that agents can learn better policies using their estimates of other players' goals in cooperative and competitive situations. However, this work does not consider environments where communication is present.

Several studies focusing on communication in artificial agents model their mental states and adjust communication protocols accordingly. For example, \citet{Lowe2017} describe how agents adapt to each other when trained in conjunction, and this finding led to the study of agents who can reason about other agents and adjust the communication protocol accordingly~\citep{Andreas2016,HawkinsKSG20,abs-2107-05697}. 

\citet{Rodriguez2019} let agents model the conceptual understanding of others by switching partners with different proprieties\footnote{For example, a color-blinded agent will not benefit from a color-oriented description of an image.}.
\citet{Grover2018} split the representation learning into two parts: a \textit{generative} embedding simulates an agent's policy, while a \textit{discriminative} one distinguishes one agent from another. These works demonstrate how agent modeling allows communication to quickly adapt and specialize to the task at hand, but this specialization can lead to languages that are difficult to interpret by humans (see Section \ref{sssec:method:natural:ld}). 

The identity of the recipient in a multi-agent environment can be just as important as the message being conveyed. For instance, \citet{Das2019} introduce \textit{TarMac}, a targeted multi-agent communication architecture that enables agents to choose their communication target using soft attention. This method assigns high attention weights when both the sender and receiver predict similar signature and query vectors. The authors evaluate their approach in four environments, including cooperative and competitive settings, and show improved performance and faster convergence across all scenarios. This finding opens new possibilities for increasing multi-agent system performance without requiring signal sharing between each agent.

\subsubsection{Influencing Others}
\label{sssec:prop:tom:influence}

\citet{Foerster2017} takes the next step by introducing the Learning with Opponent Learning Awareness (\textit{LOLA}) framework, which models the opponent's policy and attempts to actively influence it.
As a result, higher-order \textit{LOLA} emerges, in which agents are aware that opponents are trying to influence them, resulting in computationally expensive third-order derivatives.

A number of agents' modeling efforts focus on leveraging the \textit{ToM} for steering the behavior of other agents.
For example, \citet{abs-2107-05697} presents a referential game where a speaker interacts with a population of agents with different linguistic abilities\footnote{Each agent is trained in one of nine different languages: German, Lithuanian, Chinese, Italian, French, Portuguese, Spanish, Japanese and Greek} and uses model-agnostic meta-learning \citep{FinnAL17} to improve the prediction accuracy of listener's actions. 
Similarly, \citet{HawkinsKSG20} formulates the problem as a continual learning framework and tests the \textit{ToM} model in real-time interactions with humans, finding a significant increase in the probability of a correct response with successive repetitions.

In addition, \citet{abs-1810-08647} investigates the influence of intrinsic social agents equipped with a \textit{ToM} framework to stir the decision of other agents in two sequential social dilemmas.
They report above state-of-the-art performance when their agent is equipped with a \textit{ToM} model and influences the behavior of other agents, leading to effective emergent communication protocols. Furthermore, \citet{Xie2020} builds an RL environment where the agent employs an encoder-decoder architecture to model the action of a human being in a mixed human-robot setting and uses it to approximate the human policy to maximize the total discounted reward.

\subsubsection{Open challenges}

While the majority of the literature in the \emecom field focuses on Multi-Agent Systems (MAS) with artificial agents only, there are a few notable papers that explore mixed human-robot teams~\citep{JaquesLHGOSLF19,HawkinsKSG20,Xie2020}. To enhance human-machine communication, it is crucial to involve humans in the loop and investigate mixed teams' dynamics further \citep{brandizzi2022emergent}. The presence of a human in the system can provide valuable insights into how artificial agents can better adapt to human behavior, preferences, and expectations. The abovementioned papers serve as pioneering examples in this direction, and future research should build upon these foundations by expanding the investigation into various domains and settings. This may include, for instance, exploring different communication modalities, incorporating diverse human characteristics, and adapting to changing human-agent team compositions.

Finally, as artificial agents become more capable of understanding and influencing human behavior, ethical and privacy concerns will arise. It is crucial for the research community to consider these aspects when developing new methodologies and frameworks. For example, ensuring that artificial agents do not exploit vulnerabilities in human decision-making or manipulate human users for unintended purposes is vital to maintaining trust and safety.

\section{Dichotomy of Emergent Communication}
\label{sec:methods}
In Section~\ref{sec:proprieties}, we outlined the common proprieties of \emecom literature, which underscored the similarities between works and how they relate to human interaction. In contrast, in this section, we aim to define two distinct categories of works that investigate language from opposite perspectives.

The first category is Machine-centered \emecom (\symeme) where Artificial Emergent Languages (AELs) without pre-defined (linguistic) structures are considered. The goal of \symeme is to identify the environmental, architectural, and structural factors required for natural language properties to emerge. We define this approach as bottom-up, meaning that it begins with an emergent artificial language and gradually develops it into a human-like natural language over time. 

The second category, Human-centered \emecom (\nateme), emphasizes the use of natural language. In this approach, agents are provided with knowledge of a human natural language (HNL), typically English, which they then learn to apply in dynamic environments.
We define \nateme to be a top-down approach, where agents begin with the necessary knowledge to speak a language and learn how to use it for cooperative behavior in a multi-agent system.

Both methodologies are essential for understanding the dynamics that govern the emergence of language in artificial environments. By examining the strengths and weaknesses of each category, we can gain a more comprehensive understanding of what conditions and factors are necessary for human language to emerge. In fact, many papers cover both methodologies simultaneously, demonstrating their complementary nature.

\subsection{Machine-centered \emecom}
\label{ssec:methods:symbolic}
This section is concerned with exploring the properties of emergent languages, rather than their interpretability by non-expert humans. The literature discussed in this section is focused on AELs without any direct mapping to HNL. This step is essential in understanding the differences in structure and learning between human and machine languages.

Although there may be some similarities with Section~\ref{ssec:natural}, we will highlight the differences between the two sub-fields to provide a clear distinction between them.

\subsubsection{Characteristics}
\label{ssec:methods:symbolic:char}
In Machine-centered \emecom , languages are composed of symbols and numerical vectors, without any direct correspondence to HNL. As a consequence, there is no requirement for a direct mapping between a symbol and a meaning. Therefore, this approach provides greater flexibility in selecting symbols and facilitates the examination of structural and learning differences between human and machine languages.

Regarding communication channels, similarity to human communication is not a prerequisite. Utterances are introduced as a discrete set of symbols that the agents must map to other modalities, such as visual input. However, some research has investigated the use of continuous communication channels, offering an interesting approach to examining what a non-bottleneck form of interaction might look like.

\paragraph{Continuous versus Discrete}

\citet{Foerster2016} compare and analyze the effectiveness of continuous and discrete communication channels in the emergence of communication in artificial agents. The authors define two communication methods, Differentiable Inter-Agent Learning (DIAL) and Reinforced Inter-Agent Learning (RIAL), to study the learning dynamics of agents. While RIAL is similar to other related works, such as parameter sharing~\citep{SachanN18}, DIAL takes advantage of the artificial setting by allowing gradients to flow through the communication channel. The authors show that DIAL is capable of achieving faster convergence than RIAL, demonstrating that gradient provides a more robust and richer source of information.
Similar results are also reported in~\citep{abs-2302-08913,Sukhbaatar2016,Kong2017}.

\paragraph{Non-verbal communication}

In addition to verbal communication, humans also make use of non-verbal communication strategies, such as gestures and signs, to convey meaning. While most Machine-centered \emecom research focuses on symbolic signaling, some studies have explored the role of non-verbal communication in the emergent language.

For instance, \citet{abs-2010-15896} investigated emergent non-verbal communication in embodied agents within high-dimensional simulated environments. They designed a referential game in which agents produced a sequence\footnote{As a multi-step process.} of limb motion in a simulated 3D world. By providing explicit latent features, such as an energy-based structure, the agents were able to generalize to novel patterns.

Similarly, \citet{Mordatch2018}  explore the emergence of language in the context of agents embodied in a physic simulator, and notices how non-verbal communication, such as pushing, pointing, and guiding, arises as a by-product. 

In contrast, \citet{MihaiH21} focused on sketching as a form of non-verbal communication. By leveraging the differentiability of the drawing procedure, they developed a referential game and demonstrated how agents could communicate effectively. With the appropriate inductive bias, the drawings became interpretable by humans, although the authors were unsure if this was due to the visual pretrained network bias or if the agents captured some fundamental generalization of visual perception.

In a similar vein, \citet{abs-2111-14210} created a referential game where agents used sketches as a medium of communication. Unlike \citet{MihaiH21}, \citet{abs-2111-14210} used a framework more akin to Task and Talk, where the sender continuously improved the sketch until the receiver is ready for prediction. The authors reported successful communication and developed a set of evaluation metrics inspired by cognitive science. They showed how mutual adaptation and sequential decision-making could encourage symbolicity, defined as the consistent separability of drawings in high-level visual embeddings, which facilitated easy categorization of drawings by new communication participants.

Although this research is not directly related to \symeme, it can provide new insights into communication in general, given that a significant portion of human communication (70\% to 93\%) is non-verbal~\citep{Mehrabian1971, Mehrabian2017}. Furthermore, nonverbal communication is essential to human-robot interaction~\citep{VasconezGC19,BacimRSSB12}, although such interactions are often programmed manually.

\subsubsection{Hunt for Generalization}
\label{sssec:methods:symbolic:hunt}

The field of \symeme, as well as \emecom more broadly, strives to achieve a set of desirable features when emergent language is developed. These features include learning meaningful token representations~\citep{tucker2021emergent} and achieving non-trivial compositionality~\citep{steinert2020toward}, which ultimately enables the language to generalize to new concepts and ideas without having encountered them before. In this field, generalization is often studied in conjunction with compositionality, the latter is the idea that the meaning of a complex expression is determined by the meanings of its constituent and the rules governing their combination~\citep{frege1892begriff}. However, it is important to note that achieving these features, especially compositionality, can be challenging and require careful interpretation of the results.

\paragraph{Study of compositionality}

Compositionality is a highly desirable property in both human and artificial languages \citep{baroni2020linguistic}, as it allows for the generalization of concepts and ideas~\citep{pelletier1994principle,janssen1997compositionality}. However, defining compositionality can be challenging~\citep{abs-2010-15058,andreas2019measuring}, and its necessity for generalization has been debated in the literature~\citep{abs-2004-03420}. In their study, \citet{Chaabouni2020} investigate input reconstruction in a simplified signaling game. Their findings suggest that although compositionality is not a strict requirement for generalization, its presence considerably enhances the learning speed and accuracy of newly introduced agents.

Other works, such as~\citet{abs-1804-02341} and~\citet{Kottur2017}, attribute the emergence of compositionality to environmental constraints rather than specific model architecture, highlighting the importance of constructing appropriate settings for language development. For example, \citet{abs-1910-06079} employs curriculum learning, gradually increasing the difficulty of referential games, and reports an emergence of compositionality using topographic similarity and zero-shot generalization accuracy. Similarly, the introduction of iterative learning, see Section~\ref{ssec:prop:interaction}, in the language development process, has been shown to lead to increasingly compositional languages with each generation~\citep{abs-1904-09067,abs-2002-01365,abs-1912-06208,Chaabouni2019}.
The complex and dynamic chaotic environments where languages can emerge offer a rich foundation for language development~\citep{larsen1997chaos}. Consequently, a significant portion of the research efforts is focused on creating settings that closely resemble real-world conditions.

However, it is important to note that while compositionality is a desirable property, defining it can be challenging, and care should be taken when interpreting results that rely on it. Furthermore, recent studies highlight the importance of inductive biases on both the training framework and the data for the development of compositional communication \citep{abs-2010-15896,MihaiH21}. \citet{kucinski2021catalytic} theoretically and experimentally demonstrate that inductive biases on the training framework and the data are necessary for the development of compositional communication and that a noisy communication channel\footnote{A noisy channel is a communication channel that may introduce errors in the transmission of messages by independently swapping symbols  with other, uniformly sampled, symbols.} can promote compositionality in signaling games. 

These studies suggest that inductive biases should be carefully considered when evaluating and designing models for language development. This opens up new and interesting research paths for investigating the relationship between inductive biases and language learning in both humans and machines. By understanding the biases that influence language development, researchers can design more effective models and evaluation metrics for artificial language learning, as well as gain insights into the fundamental mechanisms underlying natural language evolution.

\subsubsection{Evaluating performance}
\label{sssec:methods:symbolic:eval}
The ability to recognize properties such as compositionality, verbal agreement, and deception can be challenging when dealing with symbolic languages. Therefore, several metrics have been proposed to evaluate the effectiveness of such languages. The importance of this analysis lies in the fact that artificial learning differs significantly from natural learning.

\paragraph{Cheating behaviors}
As a matter of fact, the evaluation problem has been acknowledged by \citet{Lowe2019}, who highlighted that most research in this area is focused on enhancing task performance rather than examining the semantics of the language. The authors examined how the neural network's capacity affects its ability to learn compositional languages. 
In a related work by the same authors, \citet{Resnick2019}, they investigated the relationship between the size of the language space, denoted by $\rvert L \rvert$, and the necessary number of bits required to solve a task, which is given by $=\log(\rvert L \rvert)$. They found that for large enough neural networks, agents were able to memorize the environment, which allowed them to solve the task without actually using the language, effectively bypassing the intended communication requirement.

Similar cheating behavior has been reported by \citet{Bouchacourt2018} in a referential game, where the agents achieve perfect results by communicating low-level details of the image rather than conceptual properties. The authors validate this result by providing the agents with noise images and observing that the performance is not adversely affected by such inputs.

\paragraph{Evaluation metrics}

\begin{figure}
    \centering
    \includegraphics[width=\linewidth]{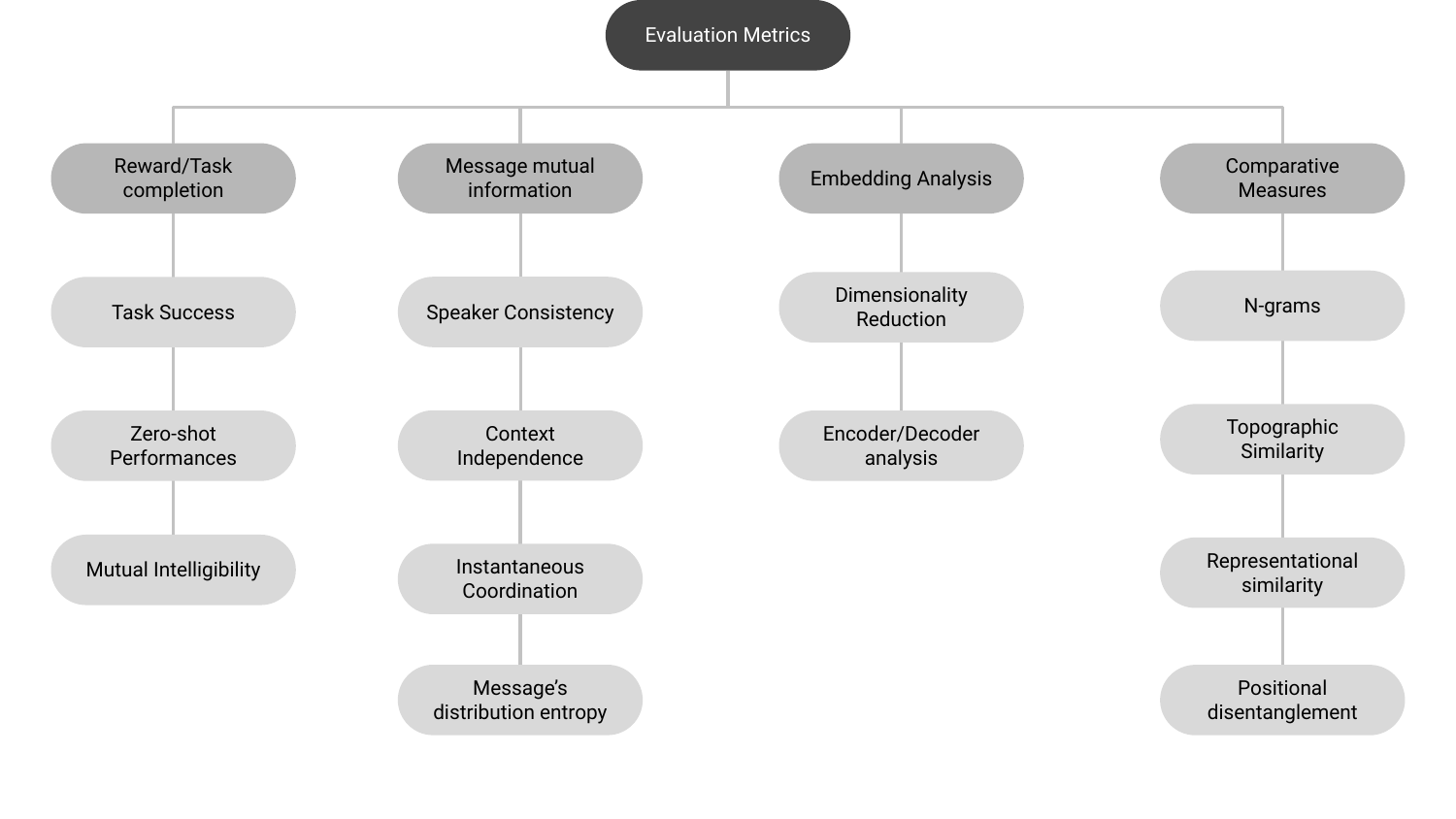}
    \caption{Hierarchical view of evaluation metrics used in Emergent Communication literature, divided into four types: (1) Reward/task completion, (2) message mutual information, (3) embedding analysis, and (4) similarity measures. Each type is further divided into specific metrics used to assess different aspects of emergent communication.}
    \label{fig:eval_metrics}
\end{figure}

\begin{figure}
    \centering
    \includegraphics[scale=0.3]{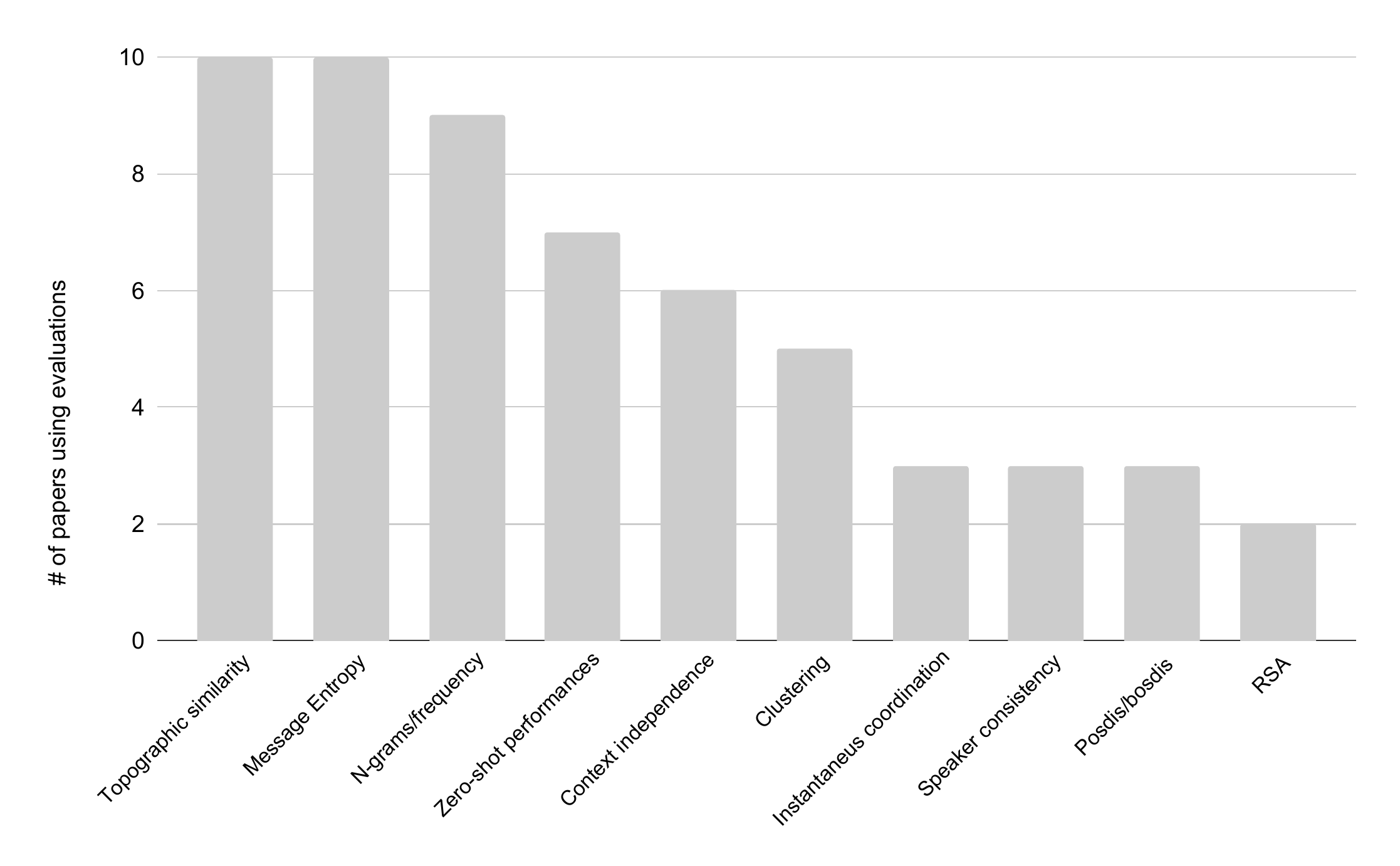}
    \caption{Bar plot showing the frequency of use of different evaluation metrics in emergent communication research, with the height of each bar representing the number of papers the metric has been used in.}
    \label{fig:eval_chart}
\end{figure}

While concepts such as compositionality and generalization have clear definitions in linguistic contexts, there is a lack of formal measurement implementations in experimental settings. As a result, researchers have developed various metrics to evaluate emerging languages. In this section, we present 14 metrics divided into four categories. Figure~\ref{fig:eval_metrics} shows a hierarchical view of the metrics, with four main types of evaluation metrics and their respective subcategories. 
Furthermore, we provide a detailed analysis of the most commonly used evaluation metrics in the surveyed literature. It is noteworthy that we found a strong preference for five specific metrics among the studies we reviewed, as illustrated in Figure~\ref{fig:eval_chart}.
These metrics provide a useful framework for evaluating the effectiveness of emergent communication models. However, it is important to note that there is no single best metric, and the choice of metric(s) should depend on the research question and specific context. Further research can be done to develop new metrics or refine existing ones to better capture the complex nature of emergent communication.

\subparagraph{Reward and task completion}

In Machine-centered \emecom (and \emecom in general), the focus is on reinforcement learning in game-like environments (as discussed in Section~\ref{ssec:prop:env}). As a result, metrics for evaluating agents' learning behavior, such as performance and game score, are essential. The most intuitive evaluation metric is \textbf{task success}, which measures the final agent's performance. All of the examined papers so far have used this metric to demonstrate the advantages of their methods.

To address the problem of catastrophic forgetting that RL agents may experience, \citet{Graesser2019} introduce a metric called \textbf{mutual intelligibility}, which estimates each agent's ability to play against itself\footnote{This metric is particularly effective in this study because the agents act as both senders and receivers, which allows for a meaningful evaluation of their mutual intelligibility.}. According to the authors: \textit{`if a shared communication protocol has emerged, the agent would not have any trouble playing a game with itself during test time'}.

However, reward and task completion metrics do not account for novel, unseen stimuli. Therefore, \textbf{zero-shot performance} is mentioned as a measure of generalization in~\citep{abs-1804-02341, Mordatch2018, Lazaridou2018, abs-1904-09067, Bouchacourt2019}. For this metric, the authors either remove specific samples from the input space or generate unseen distributions during training and evaluation. Once the model is ready for testing, these samples are reintroduced, and the performance on these previously unseen instances is reported to assess the model's generalization capabilities.

Despite their usefulness in emphasizing the adaptation capabilities of models in a game environment, these metrics do not provide insights into the characteristics of the emerged language or how it is influencing agents' behavior in the game environment~\citep{Lowe2019a}.

\subparagraph{Message mutual information}

In order to address the issue of evaluating the language content and its influence on agent behavior, some works analyze the relationship between the message content, speaker, listener, and context. One such metric is \textbf{speaker consistency}~\citep{abs-1810-08647}, which measures the alignment between an agent's message and its future action, delivering a normalized score. This measure shows \textit{how consistently a speaker agent emits a particular symbol when it takes a particular action and vice versa}. While this method is reported in ~\citep{Liang2020,Chaabouni2019,abs-1912-05676} as a reliable metric, but it fails to capture the listener behavior. 

To account for this, the same authors~\citep{abs-1810-08647} introduced \textbf{instantaneous coordination}, a similar measure of mutual information between the speaker's message and the listener's next action.
It is only natural that the same authors that used speaker consistency above also reported this metric~\citep{Liang2020,abs-1912-05676, Bouchacourt2019, Chaabouni2019}. 

Lastly, \textbf{context independence}~\citep{abs-1809-00549} measures the alignment between an agent's message and the task concept, such as the number of objects or colors in a categorical feature. 
While this formulation provides an interesting point of view for the correlation between objects and concepts, and it is used frequently in the literature~\citep{abs-1804-02341,Mordatch2018,abs-1910-06079,abs-1904-09067,Chaabouni2020,abs-1911-01892}, it requires the dataset to be feature-based.

Similarly, the \textbf{message distribution's entropy} is reported in~\citep{abs-1804-02341,Graesser2019,Liang2020,Lazaridou2018,Chaabouni2019,abs-2001-03361,Bouchacourt2019,abs-1905-12561,abs-1911-06992,KharitonovCBB20} as a measure of the correlation between the speaker's input and the message used to describe it. When the entropy is low, the speaker is consistently using the same message to describe that input, thus showing some kind of communication protocol.

\begin{figure}[tb]
    \centering
    \includegraphics[scale=0.15]{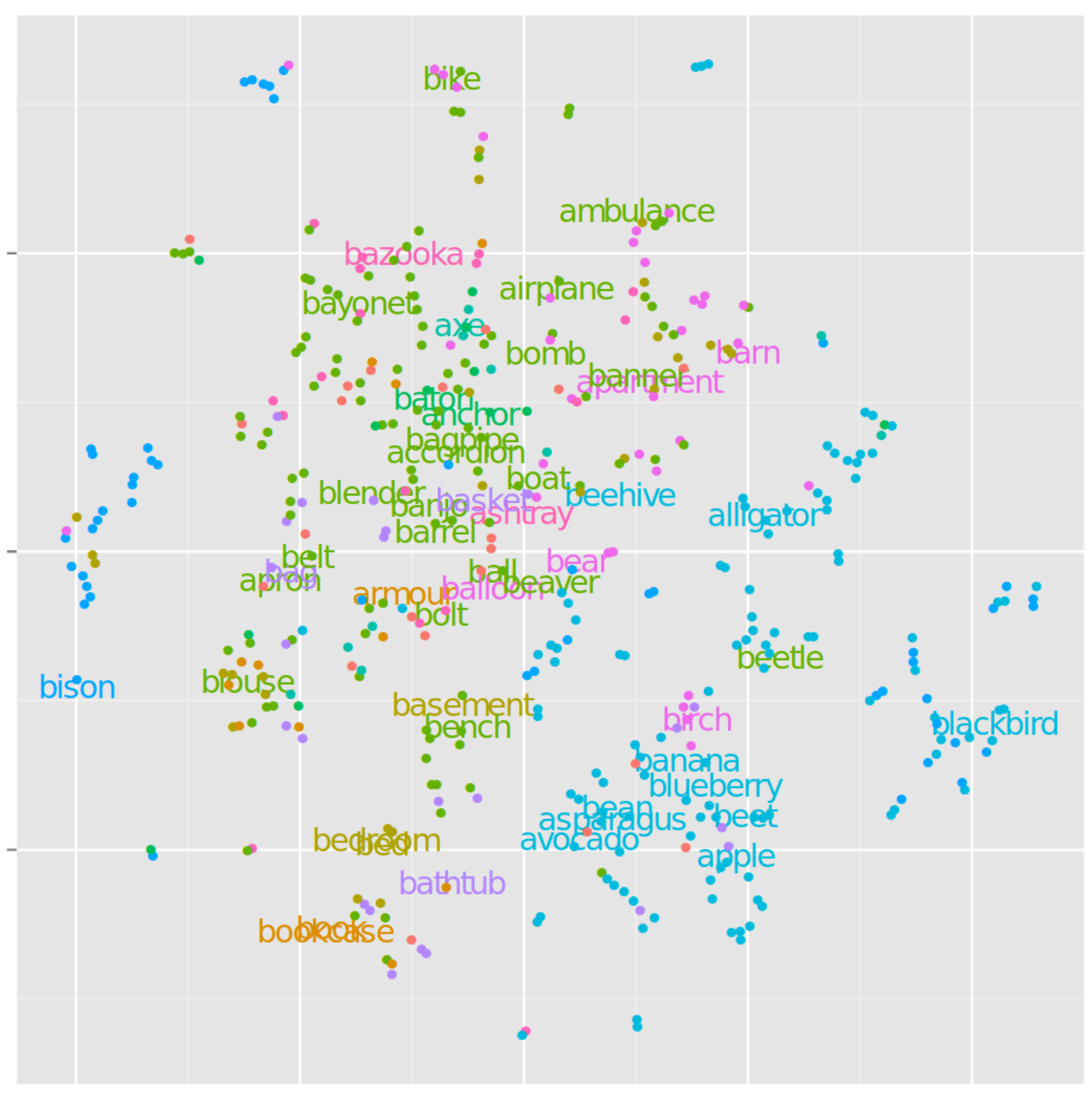}
    \caption{t-SNE plots of object fc vectors color-coded by majority symbols assigned to them by informed sender.\\ Image and Caption taken from~\citet{Lazaridou2016}.}
    \label{fig:tsne-example}
\end{figure}

\subparagraph{Embedding Analysis}
Symbolic languages are typically represented as discrete or one-hot vectors. While this is necessary for computational modeling, it also enables the use of statistical analysis and clustering techniques developed in machine learning research. 
\textbf{Dimensionality reduction} techniques such as Principal Component Analysis (PCA)~\citep{pearson1901liii,hotelling1933analysis}, and t-SNE~\citep{van2008visualizing} are often used in \symeme~\citep{Cao2018,Lazaridou2016,Sukhbaatar2016,abs-2012-10776} to identify meaningful clusters of data with respect to symbolic messages\footnote{It should be noted, however, that t-SNE can be tricky to interpret and should be used with care~\citep{wattenberg2016use}.}. For example, Figure~\ref{fig:tsne-example} shows a t-SNE projection of object vectors color-coded by majority symbols, revealing a cluster of fruits in blue on the bottom right and demonstrating how the sender relates symbols and features.

While other clustering techniques can be used for the same purpose, \citet{Cao2018} employs the \textbf{encoder/decoder} architecture of an LSTM to analyze the correlation between messages and agents' decisions. In their study, the authors trained an LSTM on generated messages and used it to predict agents' actions, assuming that the two must be correlated for the LSTM to function effectively. With this metric, they were able to identify the intention of an agent inside the message that was being generated.

\subparagraph{Comparative Measures}
While artificial settings allow for direct measurement of beliefs and intentions, studying HNL requires evaluating their syntactical, grammatical, and semantic components. 
Frequency analysis is a common statistical method that considers the frequency of symbols or sequences of symbols. \textbf{N-grams} are particularly useful for this purpose, as they count the frequency of patches of symbols and can be used to derive meaningful distributions, e.g. word lengths.  For example, Zipf's law~\citep{zipf2013psycho} is a mathematical distribution often found in human natural languages, which states that the $r$th most frequent word has a frequency of $\frac{1}{r^{\alpha}}$, with the most common word occurring twice as often as the second most frequent word, three times as often as the subsequent word, and so on:

 \begin{equation}
     f(r) \propto \frac{1}{r^{\alpha}}
     \label{eq:zipf}
 \end{equation}

Through this simple frequency analysis, many works~\citep{abs-1804-02341,Graesser2019,Liang2020,Lazaridou2018,Chaabouni2019,abs-2001-03361,Bouchacourt2019,abs-1905-12561} try to verify under which constraints an artificial language is similar to Zipf's law or other significant distributions.

On the same line, \textbf{topographic similarity} has been introduced in \citep{brighton2006understanding} to study the correlation between the distance of all the possible pairs of meaning and the corresponding pairs of signals. 
It is the most referenced evaluation metric in the literature~\citep{Li2019,Lazaridou2018,abs-1910-05291,abs-1910-06079,abs-2002-01365,Chaabouni2020,abs-2001-03361,ChaabouniSATTDM22,Bouchacourt2019} positively correlating with compositionality and indicating the close relationship between \emecom and computational linguistic analysis of human languages.

Topographic similarity is not the only measure taken from an area outside of artificial intelligence. Indeed, researchers such as~\citet{Bouchacourt2018} and~\citet{abs-1912-06208}, have borrowed \textbf{representational similarity analysis} (RSA)~\citep{kriegeskorte2008representational} from computational neuroscience to compare the similarity structure of input in the speaker and listener space. RSA is typically used to compare the similarity between evoked fMRI responses in selected brain regions. However, researchers in \emecom have found it useful to measure similarity between different kinds of input representation or to build a test set on which to analyze the generalization capabilities of emerging languages~\citep{abs-1912-06208}.

Finally, \textbf{positional disentanglement} (\textit{posdis}) and \textbf{bag-of-symbols disentanglement} (\textit{bosdis}) are metrics introduced by~\citet{Chaabouni2020} to evaluate the compositional structure of emerging languages. \textit{Posdis} measures whether symbols in specific positions tend to univocally refer to the values of a specific attribute, capturing the intuition that each position of the message should only be informative about a single attribute. In contrast, \textit{bosdis} captures the intuition of a permutation-invariant language, where only symbol counts are informative, and symbols univocally refer to distinct input elements independently of where they occur. 
These metrics offer supplementary perspectives on the structure and compositionality of artificial languages, as demonstrated in the literature by \citep{abs-2010-15058,kucinski2021catalytic}, who utilize them to examine various aspects of emerging communication systems.

\subsubsection{Open challenges}
\label{sssec:methods:symbolic:op}

Despite the challenges and limitations, \symeme has emerged as a fascinating and rapidly evolving sub-field of research with the potential to shed light on the fundamental mechanisms of human communication and language evolution. 
As we have seen, researchers have proposed a variety of metrics to evaluate the performance of artificial communication systems and to analyze the properties of the emergent languages they produce. However, there is still much work to be done in identifying the most relevant aspects of human natural language to emulate in artificial systems, and in establishing the corresponding evaluation metrics. To achieve this, interdisciplinary approaches should be adopted, drawing on insights from fields such as linguistics, cognitive science, and anthropology.

One possible future direction for research is to explore how \symeme can be integrated with more traditional approaches to NLP, such as rule-based or statistical methods. For example, researchers could investigate how \symeme could be used as a pretraining regime for large language models, allowing them to learn faster and with less data, similar to what has been done by \citep{Lowe2020,YaoYZNTG22,abs-2304-01662}. 

Although not relevant to human-machine interaction, another promising area of research is to explore how \symeme could be used to develop new communication protocols for multi-agent systems. By allowing agents to communicate in an emergent language, it may be possible to develop more efficient and effective communication strategies than those currently used in multi-agent systems. This could have important implications for a wide range of applications, from robotics and automation to online gaming and social media.

Finally, another interesting path of research is to explore how \symeme could be used to study the evolution of language itself. By creating artificial communication systems that mirror the basic principles of human language evolution, researchers may be able to gain new insights into the origins and development of language in humans, as well as the factors that contribute to the diversity of human languages around the world. This research direction can significantly benefit the fields of human language evolution and development, where artificial settings are frequently employed.

In conclusion, the study of \symeme represents an exciting and rapidly evolving area of research that holds great promise for advancing our understanding of language acquisition and communication. By continuing to develop new evaluation metrics, explore new applications, and integrate \symeme with other fields of research, we may be able to unlock new insights into the nature of language and the ways in which it evolves over time.

\subsection{Human-centered \emecom}
\label{ssec:natural}
In this section, we shift our focus to a new framework based on HNL, something that we define as Human-centered \emecom (\nateme). While some references have already been cited in the previous section on Machine-centered Emergent Communication, in this section we provide a fresh perspective on these works by examining their contributions to the \nateme sub-field.

To incorporate human natural language into the \emecom pipeline, researchers utilize datasets with human captions, such as COCO~\citep{LinMBHPRDZ14} or the Abstract Scenes Dataset~\citep{ZitnickP13}, often in conjunction with pretrained language models. 
These datasets equip artificial agents with prior knowledge about human language during training. Notably, some works also investigate the possibility of training these models within the pipeline itself.

Through the exploration of Human-centered \emecom, we aim to understand the ways in which the incorporation of natural language differs from that of symbolic language, and how it poses unique challenges to the field. Our examination of this emerging sub-field includes a review of its key works and an analysis of the various techniques and models used to overcome the obstacles presented by the complexity and variability of human language.

\paragraph{Human-centered \emecom and Image Captioning}
Before introducing \nateme, we should point out its differences with the field of Image Captioning (IC). Both utilize datasets with human captions and/or pretrained language models and aim to develop agents capable of perceiving multi-modal settings, such as vision and language, and reasoning about them using natural language. 
However, there are subtle differences in their respective methodologies. Human-centered \emecom research is developed in game settings and thus employs reinforcement learning, whereas IC  predominantly uses supervised learning. 
While \nateme explicitly models the interaction among multiple agents with a shared goal, IC is focused on architectures capable of mimicking the human ability to use language in a visual context, which may not align with human understanding \citep{DessiGFBB22}. As both fields aim to refine artificial languages to better resemble human-like ones, they should be regarded as complementary components of the broader challenge of achieving this goal.

\subsubsection{Characteristics}
\label{sssec:natural:char}
The first instance of \nateme can be traced back to the work of \citet{Andreas2016}. The authors developed a reference game where a speaker generated pragmatic\footnote{Defined as informative, fluent, concise captions encoding the understanding of listener behavior.} captions. Utilizing a reasoning speaker (see Theory of Mind in Section~\ref{ssec:prop:env}) that employed multi-modal representation, the authors evaluated their approach using two metrics: accuracy, measured by the success rate of the game, and fluency, measured by showing isolated sentences to human evaluators and asking them to rate their language quality. The authors introduced a trade-off parameter $\lambda$, which allowed for weighting the joint probability of a sentence uttered by the speaker and correctly interpreted by the listener. They found that small $\lambda$ values led to highly specific utterances with low fluency while increasing $\lambda$ caused the captions to become more generic.

While Andreas et al. focused on grounding the problem around a natural language captioned dataset, \citet{Lazaridou2016} concentrated on porting a \symeme pipeline to \nateme using a model pretrained on an image classification dataset. They developed a pipeline that allowed the sender to switch between two tasks equiprobably: a referential game and a supervised language captioning task. This approach aimed to ground the sender in HNL while simultaneously teaching it to communicate using that grounding. As a result, the relationship between images and captions made the mapping between pairs of images and supervised categories humanly interpretable. For a follow-up experiment, they used one dataset for supervised image captioning and another for referential games (ReferItGame~\citep{KazemzadehOMB14}) and asked human evaluators to determine which image a sender caption referred to. They reported an accuracy rate of $68\%$ for the latter task, concluding that supervised learning can provide a foundation for communication with humans that is generalizable beyond the distribution of image caption datasets.

While both~\citet{Andreas2016} and~\citet{Lazaridou2016} aimed to create artificial agents capable of reasoning in a multimodal environment, the methods used to achieve this goal differed, as previously noted. \citet{Lazaridou2016} started from a \symeme setup and expanded it to include HNL, an approach that we identify as \nateme. In this regard, we report related works that follow a similar approach~\citep{Das2017,Havrylov2017,abs-2003-12694,abs-2010-02975,Lazaridou2020,Lowe2020}. Conversely, \citet{Andreas2016} approached the task following the classical IC pipeline, a methodology often used in the literature~\citep{HawkinsKSG20,abs-2107-05697,abs-2110-05422}. These seemingly dual designs can blend to form numerous alternatives:~\citet{Lee2017} developed a referential game where two agents were pretrained on different languages and must evolve a \textit{common} interpretation to solve the task; \citet{Cogswell2020} used pretrained language models but then extended the game to a Task and Talk setting, which is more similar to~\citet{Andreas2016}; and \citet{Li2020} used \symeme as a pretraining framework for machine translation.

\subsubsection{Balancing Supervised and Reinforcement Learning}
\label{sssec:method:natural:balance}
As previously noted, \symeme involves the use of both supervised and reinforcement learning, which introduces distinct challenges, particularly regarding the balance between the two.

To address this balance, \citet{Lazaridou2020}  split functional learning, where agents are focused on maximizing a task-specific reward, and structural learning, where the aim is to keep the language correct and fluid. They propose various training techniques, including reward finetuning, multi-task learning, and reward-learned rerankers. 

In \citet{Evtimova2017}, a multi-modal\footnote{The sender accessed the visual portion of the game while the receiver accessed the textual portion.}, multi-step\footnote{The conversation between the agents is not limited to a single pair of question-answer.} referential game was set up to simulate more realistic settings, with the sender accessing the visual portion of the game and the receiver accessing the textual portion. The authors showed that a robust and efficient communication protocol emerges, and their work demonstrated a positive correlation between the length of conversation and the receiver's prediction confidence. They observed that sender entropy increases as the receiver asks more specific questions and investigated how agents' communication protocols become highly specialized with limited bandwidths.

\citet{Lowe2020} explore emergent communication with respect to supervised learning and self-play. They define two test cases, one inspired by the Lewis singling game and the other from \citet{Lee2017}, and they investigate what kind of scheduled learning achieves the best performance and generalization.
Their study concludes that the initial supervision phase helps overcome the discovery problem while starting the learning process with self-play leads to inconsistent language drift. 
Furthermore, population-based learning was found to outperform the previous method in mitigating both language drift and accuracy.

Similarly, \citet{Lowe2019} employed pre-trained agents, each trained on different communication protocols, as a foundation for training a meta-learning agent. This meta-learning agent was then further trained through collaboration with other agents as they learned to communicate. The resulting meta-learner could be introduced to new populations of agents with different languages and adapt more efficiently. In practice, the authors developed a dataset with a diverse range of communication paradigms, using it to train an agent based on a standard structure for general communication.

These studies suggest that \nateme puts more emphasis on balancing the competing learning paradigms of supervised language tasks and reinforcement-based referential games than on the environmental constraints of emerging compositional languages, which were the main focus of \symeme research. This shift in focus may indicate that \symeme researchers could encounter similar difficulties in their efforts to create artificial agents that can generalize to new concepts through learned communication protocols. Therefore, the lessons learned from \nateme research could potentially benefit \symeme research in the future.

\subsubsection{Language drift}
\label{sssec:method:natural:ld}
The phenomenon of human language drift (LD) has been the subject of study since the early 20th century. In his work, \citet{sapir1921language} expresses his fascination with the apparent paradox of dialect variation and analyzes how drifts are formed as a historical product, drawing similarities with the iterative learning process discussed in Section~\ref{ssec:prop:env}. However, Sapir also highlights an important aspect of LD, namely the \textit{cumulative shifting in some special direction}, and emphasizes that these shifts are \textit{not ultimately random, of course, only relatively so}. \citet{lakoff1972language} similarly argues that LD is not accidental but an inherent part of human linguistic ability. 

\paragraph{Language Drift in machines}
The phenomenon of language drift in machines shares a fundamental aspect with human language: the co-evolution and adaptation to conventional agreements between speakers, leading to an interest in computational linguistics for studying such phenomena in artificial settings~\citep{hamilton2016cultural}. However, language drifts are also seen as a misalignment between emergent communication and human language, leading to recent studies exploring how to reduce language drift by finding learning constraints. These drifts, identified as behaviors that degrade a learned language's syntactic and semantic performance, arise when a supervised language task is coupled with a reinforcement learning one. Some of the recent works exploring this topic include~\citep{Lee2019,abs-2003-12694,abs-2010-02975,Cogswell2020,Lazaridou2020,Li2020,Lowe2020,abs-2110-05422}.

\paragraph{Drift Detection}
To accurately evaluate and mitigate language drift in artificial settings, clear evaluation metrics are necessary, and the majority of research in \nateme has relied on metrics such as BLEU~\citep{Lee2019,abs-2010-02975,Li2020}, Negative Log-Likelihood\footnote{NLL measures structural drift between the generated language and the pretrained model.}~\citep{abs-2010-02975,abs-2003-12694}, uncertainty~\citep{abs-2110-05422,Cogswell2020}, cosine similarity~\citep{abs-2003-12694}, and machine translation~\citep{Lee2017,Li2020}. 

\begin{figure}
    \centering
    \includegraphics[scale=0.25]{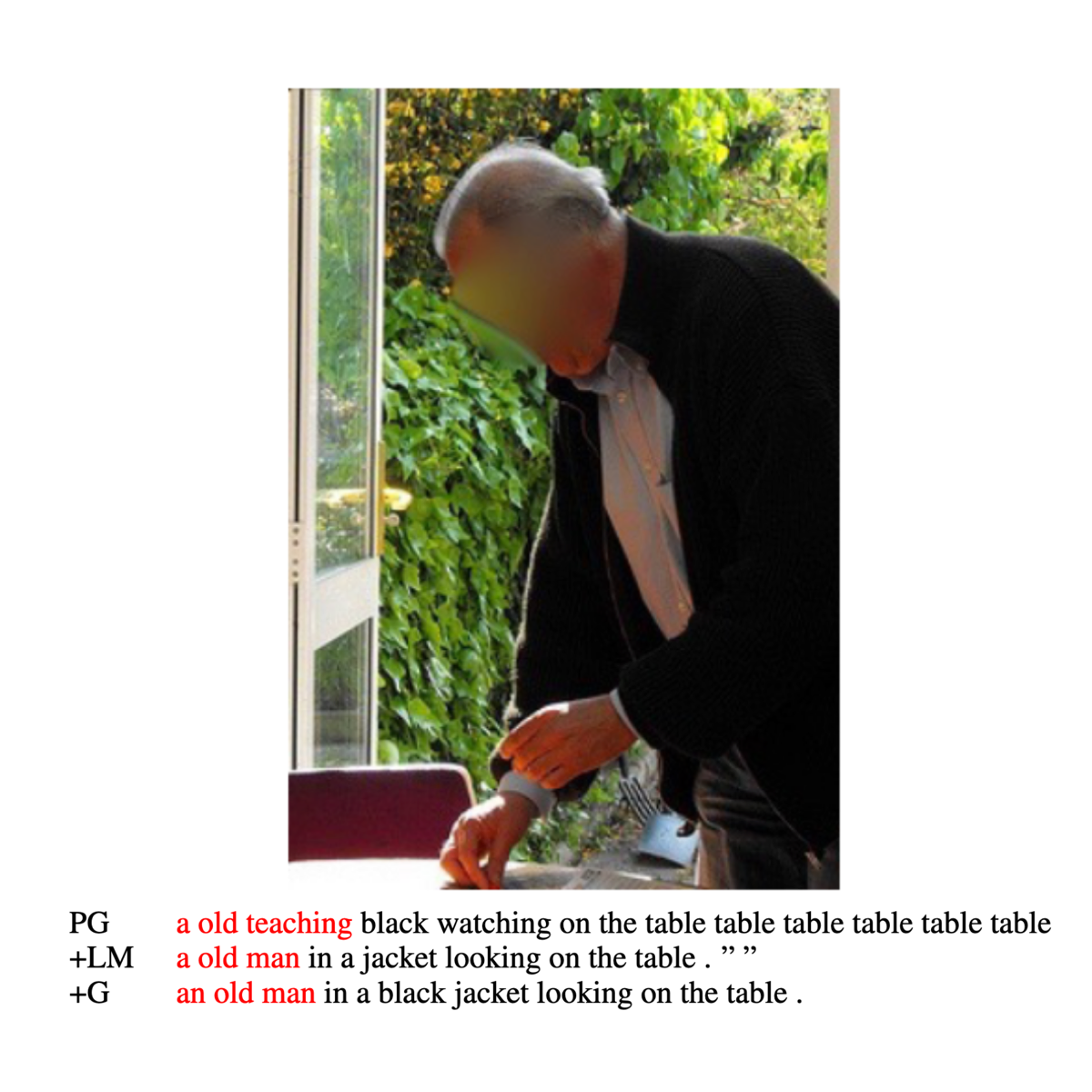}
    \caption{Communication example from Multi30k~\citep{elliott2016multi30k} dataset with different models.\\
    Image taken from~\citet{Lee2019}}
    \label{fig:ld_lee}
\end{figure}

Furthermore, \citet{Lee2019} introduce a way to estimate syntactic and semantic drifts and propose constraints to mitigate them. On one hand, syntactic constraints (LM) are expressed at the level of the pretrained language model with an auxiliary loss to measure the "Englishness" of the message. On the other hand, semantic constraints (G) are implemented on a visual ground and capture how much of the message is based on the original semantic content~\citep{KielaCJN17}. Figure~\ref{fig:ld_lee} illustrates how the communication becomes more human-interpretable and grounded in visual context when both constraints are applied (LM+G).

Additionally, \citet{Lazaridou2020} introduces an automatic method for detecting the canonical language drifts, structural and semantic, as well as a third type of drift, pragmatic drift, arising from the divergence between the human interpretation of a message and the interpretation assumed by the speaker agent due to the co-adaptation of the speaker and listener agents.

\paragraph{Iterative and population learning}

As discussed in Section~\ref{ssec:prop:env}, Iterative learning is a framework that promotes the emergence of desirable properties, such as transmissibility, efficiency, and ease of teaching, by passing knowledge down to generations of agents. In \nateme, the use of iterative learning frameworks is aimed at achieving these properties in artificial languages and mitigating the negative effects of language drift.
For instance, ~\citet{GuptaLFKP19} identify multiple algorithms\footnote{\emecom with supervised fine-tuning, supervised learning with self-play, random updates~\citep{Lazaridou2016}, scheduled updates, scheduled updates with speaker freezing~\citep{lewis2017deal}, scheduled updates with random speaker freezing.} dealing with communication in multi-agent environments and supervised dataset and introduce the term supervised self-play (S2P). The S2P loss encourages agents to maximize task completion while staying close to the initial distribution of languages by playing with past versions of themselves. Various combinations are tested, and S2P is shown to act as a regularizer between the two learning tasks. 

In contrast, \citet{abs-2003-12694} propose what they call Seeded Iterated Learning (SIL), a setting more similar to human experience, where a teacher-student architecture is coupled with an imitation learning to play a simple referential game. According to their findings, BLEU scores increased when SIL was applied compared to S2P, although no human evaluation was conducted. 

Based on the previous two works, \citet{abs-2010-02975} propose a mix called Seeded Iterated Learning (SSIL), where agents in a SIL pipeline are trained using the S2P loss on a translation game. In comparison with other baselines, the authors demonstrate that their architecture is robust to distributional shifts using BLEU and NLL.

\paragraph{Machine Translation}
Incorporating language tasks, such as machine translation, into reinforcement learning environments offers a promising solution to mitigate language drifts in \nateme. For instance, \citet{Lee2017} employs two agents, grounded in different languages, to play a visual referential game, resulting in a translation module based on emergent communication. Although this approach relies on supervised learning for their labeling task, it shows a significant increase in learning speed in an experiment with a multilingual community of agents. To address the need for unsupervised learning in machine translation, \citet{Li2020} proposes a three-step process consisting of training two agents on an unlabeled referential game, fine-tuning the resulting model on a linguistic dataset, and regularizing the model parameters. The authors report significant increases in BLEU scores on the machine translation task using this architecture. While challenges remain regarding unsupervised learning and data availability, these successes demonstrate the potential for machine translation to contribute to the development of more robust and adaptive artificial languages.

\subsubsection{Open Challenges}
\label{sssec:method:natural:op}

Natural Emergent Communication aims to develop artificial agents capable of using human natural language (HNL) in a way that goes beyond simple prediction and can effectively communicate and learn new concepts. To achieve this, researchers balance supervised language tasks and reinforcement-based referential games, creating a unique challenge. Recent studies propose various training techniques, including reward finetuning and multi-task learning. The phenomenon of language drift is a common issue in \nateme research, leading to the exploration of various ways to reduce drift, such as Seeded Iterated Learning and supervised self-play. 
While \nateme research has made significant progress in creating artificial agents capable of using natural language, there are still many open challenges that need to be addressed. 

\subparagraph{Long-term understanding} Most \nateme studies have focused on simple referential games, but real-world communication is much more complex. To effectively communicate in HNL, agents need to understand context, infer intentions, and reason about long-term goals. Developing techniques that enable agents to learn these skills will be a crucial step toward achieving HNL communication.

\subparagraph{Ethics} As \nateme research moves towards more complex communication and decision-making, there is a growing concern about the ethical implications of these technologies. Developing ethical guidelines and incorporating ethical considerations into the design of \nateme agents will be necessary to ensure that they are used in a responsible and beneficial manner.

\subparagraph{Human interaction} Ultimately, one goal of \nateme research is to develop agents that can effectively communicate with humans. To achieve this goal, it will be necessary to explore how humans interact with artificial agents and how to design agents that can effectively communicate with humans.

In conclusion, \nateme research has made significant progress in creating artificial agents capable of using natural language. However, there are still many challenges that need to be addressed to achieve robust and adaptive communication protocols that can handle a wide range of situations. 

\section{Conclusion}
\label{sec:conclusion}

The present review provides an analysis of the state of the emergent communication (\emecom) literature. Our aim is to establish a link between specific characteristics of this field and human interactions, by drawing parallels with various fields including linguistics, cognitive science, computer science, and sociology, as shown in Figure \ref{fig:prop}.

To achieve this objective, we begin by examining the common properties that are prevalent in the literature, as outlined in Section~\ref{sec:proprieties}. Our analysis identifies four key components that frequently arise in real-world interactions and we investigate their parallels within \emecom.

In Section~\ref{ssec:prop:env}, we delve into the role of environment design. Specifically, we examine the distinction between communication as the primary objective (\textit{Communication-focused}) versus communication as a tool to achieve other tasks (\textit{Communication-assisted}), as discussed in Section~\ref{sssec:prop:env:comm}. Additionally, we explore the influence of input representation on \emecom, as outlined in Section~\ref{sssec:prop:env:input}. 

For newcomers to the field, the question of how to train artificial agents in the \emecom pipeline can be overwhelming and hinder understanding. To address this issue, in Section~\ref{ssec:prop:learning}, we identify the most common learning paradigms employed in \emecom research. Specifically, we focus on two popular methods: reinforcement learning (Section~\ref{sssec:prop:learning:rl}) and supervised learning (Section~\ref{sssec:prop:learning:sl}), while drawing comparisons to human learning capabilities.

The design of the environment in emergent communication research is influenced not only by task-oriented goals but also by the type of interaction that occurs between agents. To further explore this aspect, Section~\ref{ssec:prop:interaction} focuses on identifying two types of interactions that can take place. The first type, grounded in a spatial component, is discussed in Section~\ref{ssec:prop:interaction:space}. This section introduces the concept of both inner and outer interactions, which can be cooperative or competitive and occur between agents belonging to the same or different teams. While these interactions are rooted in a spatial context, we also emphasize the importance of the temporal aspect by introducing the concept of iterative learning, discussed in Section~\ref{ssec:prop:interaction:time}.

Researchers in the field of \emecom have drawn inspiration from various disciplines, including cognitive science. The presence of multiple agents in the system has led to the adoption of ideas from cognitive science, such as the Theory of Mind (\textit{ToM}). Section~\ref{ssec:prop:tom} examines how \textit{ToM} enables artificial agents to model other intelligent entities as distinct individuals separate from their environments, as discussed in Section~\ref{sssec:prop:tom:model}. Furthermore, this modeling naturally extends to the concept of \textit{influencing other} agents, which is explored in Section~\ref{sssec:prop:tom:influence}.

In our analysis of \emecom, we distinguish between two primary sub-fields that vary in their approach, as outlined in Section~\ref{sec:methods}. The first, Machine-centered \emecom (\symeme), presented in Section~\ref{ssec:methods:symbolic}, predominantly employs symbolic languages represented as numerical vectors, with an emphasis on discovering the appropriate constraints to observe common properties of natural languages, such as generalization and compositionality. Our discussion in Section~\ref{sssec:methods:symbolic:hunt} revolves around the quest for generalization and the various techniques used to detect it within the field, as explored in Section~\ref{sssec:methods:symbolic:eval}.

Conversely, Human-centered \emecom (\nateme), as presented in Section~\ref{ssec:natural}, encompasses works that utilize natural language in their settings, with a focus on balancing task-oriented learning and language supervision. In Section~\ref{sssec:natural:char}, we delineate the characteristics of this sub-field, and in Section~\ref{sssec:method:natural:balance}, we outline its objectives. Lastly, in Section~\ref{sssec:method:natural:ld}, we discuss the issue of language drift, which is often regarded as the most significant challenge in this sub-field.

\subsection{Implications}

This review has made two significant contributions to the field of emergent communication. First, we have provided an extensive review of the relevant literature, distinguishing the commonalities and differences among various approaches. The list of references presented in Table~\ref{tab:list} serves as a valuable resource for researchers interested in this dynamic field.

Second, we have emphasized the robust connection between emergent communication and human-machine interaction. Although the majority of the analyzed literature has concentrated on multi-agent systems comprising solely artificial agents, we propose that incorporating a human-in-the-loop approach, particularly in mixed human-robot teams, offers great potential for future research. This approach facilitates a more realistic approximation of human communication in real-world settings, allowing researchers to more accurately model and analyze the intricate interplay of language, cognition, and social interaction.

In conclusion, our examination of the emergent communication literature demonstrates that this field holds great potential for enhancing our understanding of human communication and for developing more robust and trustworthy artificial communication systems. By underscoring the connection between emergent communication and human-machine interaction, we aim to inspire future research that considers the rich complexity of human communication and interaction.

\section*{Acknowledgments}

I want to express my sincere gratitude to Marco Baroni, Roberto Dess{ì}, Lukas Galke, Kristina Kobrock, and Mitja Nikolaus for providing valuable feedback and insightful suggestions, which have greatly improved the quality and clarity of this work.
Moreover, I would like to thank Roberto Cipollone, Francesco Frattolillo and Luca Iocchi for reviewing parts of this work. 

This material is based upon work supported by the Air Force Office of Scientific Research under award number FA8655-23-1-7257.

\bibliographystyle{unsrtnat}
\bibliography{citations}  

\begin{thebibliography}{177}
\providecommand{\natexlab}[1]{#1}
\providecommand{\url}[1]{\texttt{#1}}
\expandafter\ifx\csname urlstyle\endcsname\relax
  \providecommand{\doi}[1]{doi: #1}\else
  \providecommand{\doi}{doi: \begingroup \urlstyle{rm}\Url}\fi

\bibitem[Silver et~al.(2018)Silver, Hubert, Schrittwieser, Antonoglou, Lai,
  Guez, Lanctot, Sifre, Kumaran, Graepel, Lillicrap, Simonyan, and
  Hassabis]{silver2017mastering}
David Silver, Thomas Hubert, Julian Schrittwieser, Ioannis Antonoglou, Matthew
  Lai, Arthur Guez, Marc Lanctot, Laurent Sifre, Dharshan Kumaran, Thore
  Graepel, Timothy Lillicrap, Karen Simonyan, and Demis Hassabis.
\newblock A general reinforcement learning algorithm that masters chess, shogi,
  and go through self-play.
\newblock \emph{Science}, 362\penalty0 (6419):\penalty0 1140--1144, 2018.
\newblock \doi{10.1126/science.aar6404}.
\newblock URL \url{https://www.science.org/doi/abs/10.1126/science.aar6404}.

\bibitem[Silver et~al.(2016)Silver, Huang, Maddison, Guez, Sifre, Van
  Den~Driessche, Schrittwieser, Antonoglou, Panneershelvam, Lanctot,
  et~al.]{silver2016mastering}
David Silver, Aja Huang, Chris~J Maddison, Arthur Guez, Laurent Sifre, George
  Van Den~Driessche, Julian Schrittwieser, Ioannis Antonoglou, Veda
  Panneershelvam, Marc Lanctot, et~al.
\newblock Mastering the game of go with deep neural networks and tree search.
\newblock \emph{nature}, 529\penalty0 (7587):\penalty0 484--489, 2016.

\bibitem[Vinyals et~al.(2019)Vinyals, Babuschkin, Czarnecki, Mathieu, Dudzik,
  Chung, Choi, Powell, Ewalds, Georgiev, et~al.]{vinyals2019grandmaster}
Oriol Vinyals, Igor Babuschkin, Wojciech~M Czarnecki, Micha{\"e}l Mathieu,
  Andrew Dudzik, Junyoung Chung, David~H Choi, Richard Powell, Timo Ewalds,
  Petko Georgiev, et~al.
\newblock Grandmaster level in starcraft ii using multi-agent reinforcement
  learning.
\newblock \emph{Nature}, 575\penalty0 (7782):\penalty0 350--354, 2019.

\bibitem[Xu(2019)]{3328485}
Wei Xu.
\newblock Toward human-centered ai: A perspective from human-computer
  interaction.
\newblock \emph{Interactions}, 26\penalty0 (4):\penalty0 42–46, 09 2019.
\newblock ISSN 1072-5520.
\newblock \doi{10.1145/3328485}.
\newblock URL \url{https://doi.org/10.1145/3328485}.

\bibitem[Riedl(2019)]{hbe2}
Mark~O. Riedl.
\newblock Human-centered artificial intelligence and machine learning.
\newblock \emph{Human Behavior and Emerging Technologies}, 1\penalty0
  (1):\penalty0 33--36, 2019.
\newblock \doi{https://doi.org/10.1002/hbe2.117}.
\newblock URL \url{https://onlinelibrary.wiley.com/doi/abs/10.1002/hbe2.117}.

\bibitem[Shneiderman(2021)]{shneiderman2022human}
Ben Shneiderman.
\newblock Human-centered {AI:} {A} new synthesis.
\newblock In Carmelo Ardito, Rosa Lanzilotti, Alessio Malizia, Helen Petrie,
  Antonio Piccinno, Giuseppe Desolda, and Kori Inkpen, editors,
  \emph{Human-Computer Interaction - {INTERACT} 2021 - 18th {IFIP} {TC} 13
  International Conference, Bari, Italy, August 30 - September 3, 2021,
  Proceedings, Part {I}}, volume 12932 of \emph{Lecture Notes in Computer
  Science}, pages 3--8. Springer, 2021.
\newblock \doi{10.1007/978-3-030-85623-6\_1}.
\newblock URL \url{https://doi.org/10.1007/978-3-030-85623-6\_1}.

\bibitem[Mikolov et~al.(2016)Mikolov, Joulin, and Baroni]{Mikolov2016}
Tomas Mikolov, Armand Joulin, and Marco Baroni.
\newblock A roadmap towards machine intelligence.
\newblock In \emph{International Conference on Intelligent Text Processing and
  Computational Linguistics}, pages 29--61. Springer, 2016.

\bibitem[Vaswani et~al.(2017)Vaswani, Shazeer, Parmar, Uszkoreit, Jones, Gomez,
  Kaiser, and Polosukhin]{vaswani2017attention}
Ashish Vaswani, Noam Shazeer, Niki Parmar, Jakob Uszkoreit, Llion Jones,
  Aidan~N Gomez, {\L}ukasz Kaiser, and Illia Polosukhin.
\newblock Attention is all you need.
\newblock \emph{Advances in neural information processing systems}, 30, 2017.

\bibitem[Brown et~al.(2020)Brown, Mann, Ryder, Subbiah, Kaplan, Dhariwal,
  Neelakantan, Shyam, Sastry, Askell, Agarwal, Herbert{-}Voss, Krueger,
  Henighan, Child, Ramesh, Ziegler, Wu, Winter, Hesse, Chen, Sigler, Litwin,
  Gray, Chess, Clark, Berner, McCandlish, Radford, Sutskever, and
  Amodei]{Brown2020}
Tom~B. Brown, Benjamin Mann, Nick Ryder, Melanie Subbiah, Jared Kaplan,
  Prafulla Dhariwal, Arvind Neelakantan, Pranav Shyam, Girish Sastry, Amanda
  Askell, Sandhini Agarwal, Ariel Herbert{-}Voss, Gretchen Krueger, Tom
  Henighan, Rewon Child, Aditya Ramesh, Daniel~M. Ziegler, Jeffrey Wu, Clemens
  Winter, Christopher Hesse, Mark Chen, Eric Sigler, Mateusz Litwin, Scott
  Gray, Benjamin Chess, Jack Clark, Christopher Berner, Sam McCandlish, Alec
  Radford, Ilya Sutskever, and Dario Amodei.
\newblock Language models are few-shot learners.
\newblock In Hugo Larochelle, Marc'Aurelio Ranzato, Raia Hadsell,
  Maria{-}Florina Balcan, and Hsuan{-}Tien Lin, editors, \emph{Advances in
  Neural Information Processing Systems 33: Annual Conference on Neural
  Information Processing Systems 2020, NeurIPS 2020, December 6-12, 2020,
  virtual}, 2020.
\newblock URL
  \url{https://proceedings.neurips.cc/paper/2020/hash/1457c0d6bfcb4967418bfb8ac142f64a-Abstract.html}.

\bibitem[Touvron et~al.(2023)Touvron, Lavril, Izacard, Martinet, Lachaux,
  Lacroix, Rozi{\`{e}}re, Goyal, Hambro, Azhar, Rodriguez, Joulin, Grave, and
  Lample]{abs-2302-13971}
Hugo Touvron, Thibaut Lavril, Gautier Izacard, Xavier Martinet, Marie{-}Anne
  Lachaux, Timoth{\'{e}}e Lacroix, Baptiste Rozi{\`{e}}re, Naman Goyal, Eric
  Hambro, Faisal Azhar, Aur{\'{e}}lien Rodriguez, Armand Joulin, Edouard Grave,
  and Guillaume Lample.
\newblock Llama: Open and efficient foundation language models.
\newblock \emph{CoRR}, abs/2302.13971, 2023.
\newblock \doi{10.48550/arXiv.2302.13971}.
\newblock URL \url{https://doi.org/10.48550/arXiv.2302.13971}.

\bibitem[Thoppilan et~al.(2022)Thoppilan, Freitas, Hall, Shazeer, Kulshreshtha,
  Cheng, Jin, Bos, Baker, Du, Li, Lee, Zheng, Ghafouri, Menegali, Huang,
  Krikun, Lepikhin, Qin, Chen, Xu, Chen, Roberts, Bosma, Zhou, Chang, Krivokon,
  Rusch, Pickett, Meier{-}Hellstern, Morris, Doshi, Santos, Duke, Soraker,
  Zevenbergen, Prabhakaran, Diaz, Hutchinson, Olson, Molina, Hoffman{-}John,
  Lee, Aroyo, Rajakumar, Butryna, Lamm, Kuzmina, Fenton, Cohen, Bernstein,
  Kurzweil, Aguera{-}Arcas, Cui, Croak, Chi, and Le]{thoppilan2022lamda}
Romal Thoppilan, Daniel~De Freitas, Jamie Hall, Noam Shazeer, Apoorv
  Kulshreshtha, Heng{-}Tze Cheng, Alicia Jin, Taylor Bos, Leslie Baker, Yu~Du,
  YaGuang Li, Hongrae Lee, Huaixiu~Steven Zheng, Amin Ghafouri, Marcelo
  Menegali, Yanping Huang, Maxim Krikun, Dmitry Lepikhin, James Qin, Dehao
  Chen, Yuanzhong Xu, Zhifeng Chen, Adam Roberts, Maarten Bosma, Yanqi Zhou,
  Chung{-}Ching Chang, Igor Krivokon, Will Rusch, Marc Pickett, Kathleen~S.
  Meier{-}Hellstern, Meredith~Ringel Morris, Tulsee Doshi, Renelito~Delos
  Santos, Toju Duke, Johnny Soraker, Ben Zevenbergen, Vinodkumar Prabhakaran,
  Mark Diaz, Ben Hutchinson, Kristen Olson, Alejandra Molina, Erin
  Hoffman{-}John, Josh Lee, Lora Aroyo, Ravi Rajakumar, Alena Butryna, Matthew
  Lamm, Viktoriya Kuzmina, Joe Fenton, Aaron Cohen, Rachel Bernstein, Ray
  Kurzweil, Blaise Aguera{-}Arcas, Claire Cui, Marian Croak, Ed~H. Chi, and
  Quoc Le.
\newblock Lamda: Language models for dialog applications.
\newblock \emph{CoRR}, abs/2201.08239, 2022.
\newblock URL \url{https://arxiv.org/abs/2201.08239}.

\bibitem[Linzen(2020)]{Linzen2020}
Tal Linzen.
\newblock How can we accelerate progress towards human-like linguistic
  generalization?
\newblock In Dan Jurafsky, Joyce Chai, Natalie Schluter, and Joel~R. Tetreault,
  editors, \emph{Proceedings of the 58th Annual Meeting of the Association for
  Computational Linguistics, {ACL} 2020, Online, July 5-10, 2020}, pages
  5210--5217. Association for Computational Linguistics, 2020.
\newblock \doi{10.18653/v1/2020.acl-main.465}.
\newblock URL \url{https://doi.org/10.18653/v1/2020.acl-main.465}.

\bibitem[Wagner et~al.(2003)Wagner, Reggia, Uriagereka, and
  Wilkinson]{Wagner2003}
Kyle Wagner, James~A. Reggia, Juan Uriagereka, and Gerald Wilkinson.
\newblock Progress in the simulation of emergent communication and language.
\newblock \emph{Adaptive Behavior}, 11\penalty0 (1):\penalty0 37--69, 2003.
\newblock \doi{10.1177/10597123030111003}.
\newblock URL \url{https://doi.org/10.1177/10597123030111003}.

\bibitem[King(2009)]{king2009emergent}
Cynthia~L King.
\newblock Emergent communication strategies.
\newblock \emph{International journal of strategic communication}, 4\penalty0
  (1):\penalty0 19--38, 2009.

\bibitem[Boldt and Mortensen(2022)]{boldt2022recommendations}
Brendon Boldt and David Mortensen.
\newblock Recommendations for systematic research on emergent language.
\newblock \emph{CoRR}, abs/2206.11302, 2022.
\newblock \doi{10.48550/arXiv.2206.11302}.
\newblock URL \url{https://doi.org/10.48550/arXiv.2206.11302}.

\bibitem[Jaderberg et~al.(2019)Jaderberg, Czarnecki, Dunning, Marris, Lever,
  Castaneda, Beattie, Rabinowitz, Morcos, Ruderman, et~al.]{jaderberg2019human}
Max Jaderberg, Wojciech~M Czarnecki, Iain Dunning, Luke Marris, Guy Lever,
  Antonio~Garcia Castaneda, Charles Beattie, Neil~C Rabinowitz, Ari~S Morcos,
  Avraham Ruderman, et~al.
\newblock Human-level performance in 3d multiplayer games with population-based
  reinforcement learning.
\newblock \emph{Science}, 364\penalty0 (6443):\penalty0 859--865, 2019.

\bibitem[Liu et~al.(2021)Liu, Lamb, Kawaguchi, ALIAS PARTH~GOYAL, Sun, Mozer,
  and Bengio]{liu2021discrete}
Dianbo Liu, Alex~M Lamb, Kenji Kawaguchi, Anirudh~Goyal ALIAS PARTH~GOYAL, Chen
  Sun, Michael~C Mozer, and Yoshua Bengio.
\newblock Discrete-valued neural communication.
\newblock \emph{Advances in Neural Information Processing Systems},
  34:\penalty0 2109--2121, 2021.

\bibitem[Lake and Baroni(2018)]{lake2018generalization}
Brenden Lake and Marco Baroni.
\newblock Generalization without systematicity: On the compositional skills of
  sequence-to-sequence recurrent networks.
\newblock In \emph{International conference on machine learning}, pages
  2873--2882. PMLR, 2018.

\bibitem[Jiang et~al.(2019)Jiang, Gu, Murphy, and Finn]{jiang2019language}
Yiding Jiang, Shixiang~Shane Gu, Kevin~P Murphy, and Chelsea Finn.
\newblock Language as an abstraction for hierarchical deep reinforcement
  learning.
\newblock \emph{Advances in Neural Information Processing Systems}, 32, 2019.

\bibitem[Selten and Warglien(2007)]{selten2007emergence}
Reinhard Selten and Massimo Warglien.
\newblock The emergence of simple languages in an experimental coordination
  game.
\newblock \emph{Proceedings of the National Academy of Sciences}, 104\penalty0
  (18):\penalty0 7361--7366, 2007.

\bibitem[Winters et~al.(2015)Winters, Kirby, and Smith]{winters2015languages}
James Winters, Simon Kirby, and Kenny Smith.
\newblock Languages adapt to their contextual niche.
\newblock \emph{Language and Cognition}, 7\penalty0 (3):\penalty0 415--449,
  2015.

\bibitem[Raviv et~al.(2019)Raviv, Meyer, and Lev-Ari]{raviv2019larger}
Limor Raviv, Antje Meyer, and Shiri Lev-Ari.
\newblock Larger communities create more systematic languages.
\newblock \emph{Proceedings of the Royal Society B}, 286\penalty0
  (1907):\penalty0 20191262, 2019.

\bibitem[Raczaszek-Leonardi et~al.(2018)Raczaszek-Leonardi, Nomikou, Rohlfing,
  and Deacon]{Rączaszeklang}
Joanna Raczaszek-Leonardi, Iris Nomikou, Katharina~J. Rohlfing, and Terrence~W.
  Deacon.
\newblock Language development from an ecological perspective: Ecologically
  valid ways to abstract symbols.
\newblock \emph{Ecological Psychology}, 30\penalty0 (1):\penalty0 39--73, 2018.
\newblock \doi{10.1080/10407413.2017.1410387}.
\newblock URL \url{https://doi.org/10.1080/10407413.2017.1410387}.

\bibitem[Graesser et~al.(2019)Graesser, Cho, and Kiela]{Graesser2019}
Laura Graesser, Kyunghyun Cho, and Douwe Kiela.
\newblock Emergent linguistic phenomena in multi-agent communication games.
\newblock \emph{CoRR}, abs/1901.08706, 2019.
\newblock URL \url{http://arxiv.org/abs/1901.08706}.

\bibitem[Li and Bowling(2019)]{Li2019}
Fushan Li and Michael Bowling.
\newblock Ease-of-teaching and language structure from emergent communication.
\newblock In \emph{Advances in Neural Information Processing Systems}, pages
  15851--15861, 2019.

\bibitem[Eccles et~al.(2019)Eccles, Bachrach, Lever, Lazaridou, and
  Graepel]{abs-1912-05676}
Tom Eccles, Yoram Bachrach, Guy Lever, Angeliki Lazaridou, and Thore Graepel.
\newblock Biases for emergent communication in multi-agent reinforcement
  learning.
\newblock \emph{CoRR}, abs/1912.05676, 2019.
\newblock URL \url{http://arxiv.org/abs/1912.05676}.

\bibitem[Yang et~al.(2018)Yang, Yu, Bai, Wen, Zhang, and Wang]{YangYBWZW18}
Yaodong Yang, Lantao Yu, Yiwei Bai, Ying Wen, Weinan Zhang, and Jun Wang.
\newblock A study of {AI} population dynamics with million-agent reinforcement
  learning.
\newblock In Elisabeth Andr{\'{e}}, Sven Koenig, Mehdi Dastani, and Gita
  Sukthankar, editors, \emph{Proceedings of the 17th International Conference
  on Autonomous Agents and MultiAgent Systems, {AAMAS} 2018, Stockholm, Sweden,
  July 10-15, 2018}, pages 2133--2135. International Foundation for Autonomous
  Agents and Multiagent Systems Richland, SC, {USA} / {ACM}, 2018.
\newblock URL \url{http://dl.acm.org/citation.cfm?id=3238096}.

\bibitem[Spinka et~al.(2001)Spinka, Newberry, and Bekoff]{Spinka2001}
Marek Spinka, Ruth~C Newberry, and Marc Bekoff.
\newblock Mammalian play: training for the unexpected.
\newblock \emph{The Quarterly review of biology}, 76\penalty0 (2):\penalty0
  141--168, 2001.

\bibitem[Tahmores(2011)]{Tahmores2011}
Aghajani~Hashtchin Tahmores.
\newblock Role of play in social skills and intelligence of children.
\newblock \emph{Procedia-Social and Behavioral Sciences}, 30:\penalty0
  2272--2279, 2011.

\bibitem[Kirriemuir and McFarlane(2004)]{Kirriemuir2004}
John Kirriemuir and Angela McFarlane.
\newblock Literature review in games and learning.
\newblock 2004.

\bibitem[Lazaridou and Baroni(2020)]{abs-2006-02419}
Angeliki Lazaridou and Marco Baroni.
\newblock Emergent multi-agent communication in the deep learning era.
\newblock \emph{CoRR}, abs/2006.02419, 2020.
\newblock URL \url{https://arxiv.org/abs/2006.02419}.

\bibitem[Lazaridou et~al.(2016)Lazaridou, Pham, and Baroni]{LazaridouPB16a}
Angeliki Lazaridou, Nghia~The Pham, and Marco Baroni.
\newblock Towards multi-agent communication-based language learning.
\newblock \emph{CoRR}, abs/1605.07133, 2016.
\newblock URL \url{http://arxiv.org/abs/1605.07133}.

\bibitem[Eitan et~al.(2020)Eitan, Smolyansky, Harpaz, and
  Perets]{connected_papers}
Alex~Tarnavsky Eitan, Eddie Smolyansky, Itay~Knaan Harpaz, and Sahar Perets.
\newblock Find and explore academic papers, 2020.
\newblock URL \url{https://www.connectedpapers.com/}.

\bibitem[Allee et~al.(1949)Allee, Park, Emerson, Park, Schmidt,
  et~al.]{Allee1949}
Warder~Clyde Allee, Orlando Park, Alfred~Edwards Emerson, Thomas Park,
  Karl~Patterson Schmidt, et~al.
\newblock \emph{Principles of animal ecology}.
\newblock Saunders Company Philadelphia, Pennsylvania, USA, 1949.

\bibitem[Lewis(1969)]{Lewis2008}
David~Kellogg Lewis.
\newblock \emph{Convention: A Philosophical Study}.
\newblock Cambridge, MA, USA: Wiley-Blackwell, 1969.

\bibitem[Das et~al.(2017)Das, Kottur, Moura, Lee, and Batra]{Das2017}
Abhishek Das, Satwik Kottur, Jos{\'e}~MF Moura, Stefan Lee, and Dhruv Batra.
\newblock Learning cooperative visual dialog agents with deep reinforcement
  learning.
\newblock In \emph{Proceedings of the IEEE international conference on computer
  vision}, pages 2951--2960, 2017.

\bibitem[Yuan et~al.(2020)Yuan, Fu, Shen, Xu, Shen, and Zhu]{abs-2001-07752}
Luyao Yuan, Zipeng Fu, Jingyue Shen, Lu~Xu, Junhong Shen, and Song{-}Chun Zhu.
\newblock Emergence of pragmatics from referential game between theory of mind
  agents.
\newblock \emph{CoRR}, abs/2001.07752, 2020.
\newblock URL \url{https://arxiv.org/abs/2001.07752}.

\bibitem[Rodriguez et~al.(2019)Rodriguez, Alaniz, and Akata]{Rodriguez2019}
Rodolfo~Corona Rodriguez, Stephan Alaniz, and Zeynep Akata.
\newblock Modeling conceptual understanding in image reference games.
\newblock In \emph{Advances in Neural Information Processing Systems}, pages
  13155--13165, 2019.

\bibitem[Dagan et~al.(2020)Dagan, Hupkes, and Bruni]{abs-2001-03361}
Gautier Dagan, Dieuwke Hupkes, and Elia Bruni.
\newblock Co-evolution of language and agents in referential games.
\newblock \emph{CoRR}, abs/2001.03361, 2020.
\newblock URL \url{https://arxiv.org/abs/2001.03361}.

\bibitem[Chaabouni et~al.(2020)Chaabouni, Kharitonov, Bouchacourt, Dupoux, and
  Baroni]{Chaabouni2020}
Rahma Chaabouni, Eugene Kharitonov, Diane Bouchacourt, Emmanuel Dupoux, and
  Marco Baroni.
\newblock Compositionality and generalization in emergent languages.
\newblock In Dan Jurafsky, Joyce Chai, Natalie Schluter, and Joel~R. Tetreault,
  editors, \emph{Proceedings of the 58th Annual Meeting of the Association for
  Computational Linguistics, {ACL} 2020, Online, July 5-10, 2020}, pages
  4427--4442. Association for Computational Linguistics, 2020.
\newblock \doi{10.18653/v1/2020.acl-main.407}.
\newblock URL \url{https://doi.org/10.18653/v1/2020.acl-main.407}.

\bibitem[Havrylov and Titov(2017)]{Havrylov2017}
Serhii Havrylov and Ivan Titov.
\newblock Emergence of language with multi-agent games: Learning to communicate
  with sequences of symbols.
\newblock In \emph{Advances in neural information processing systems}, pages
  2149--2159, 2017.

\bibitem[Lazaridou et~al.(2017)Lazaridou, Peysakhovich, and
  Baroni]{Lazaridou2016}
Angeliki Lazaridou, Alexander Peysakhovich, and Marco Baroni.
\newblock Multi-agent cooperation and the emergence of (natural) language.
\newblock In \emph{5th International Conference on Learning Representations,
  {ICLR} 2017, Toulon, France, April 24-26, 2017, Conference Track
  Proceedings}. OpenReview.net, 2017.
\newblock URL \url{https://openreview.net/forum?id=Hk8N3Sclg}.

\bibitem[Wang et~al.(2021)Wang, White, Mu, and Goodman]{abs-2110-05422}
Rose~E. Wang, Julia White, Jesse Mu, and Noah~D. Goodman.
\newblock Calibrate your listeners! robust communication-based training for
  pragmatic speakers.
\newblock \emph{CoRR}, abs/2110.05422, 2021.
\newblock URL \url{https://arxiv.org/abs/2110.05422}.

\bibitem[Kottur et~al.(2017)Kottur, Moura, Lee, and Batra]{Kottur2017}
Satwik Kottur, Jos{\'{e}} M.~F. Moura, Stefan Lee, and Dhruv Batra.
\newblock Natural language does not emerge 'naturally' in multi-agent dialog.
\newblock \emph{CoRR}, abs/1706.08502, 2017.
\newblock URL \url{http://arxiv.org/abs/1706.08502}.

\bibitem[Sally(1995)]{Sally1995}
David Sally.
\newblock Conversation and cooperation in social dilemmas: A meta-analysis of
  experiments from 1958 to 1992.
\newblock \emph{Rationality and society}, 7\penalty0 (1):\penalty0 58--92,
  1995.

\bibitem[Liang et~al.(2020)Liang, Chen, Salakhutdinov, Morency, and
  Kottur]{Liang2020}
Paul~Pu Liang, Jeffrey Chen, Ruslan Salakhutdinov, Louis{-}Philippe Morency,
  and Satwik Kottur.
\newblock On emergent communication in competitive multi-agent teams.
\newblock \emph{CoRR}, abs/2003.01848, 2020.
\newblock URL \url{https://arxiv.org/abs/2003.01848}.

\bibitem[Cogswell et~al.(2019)Cogswell, Lu, Lee, Parikh, and
  Batra]{abs-1904-09067}
Michael Cogswell, Jiasen Lu, Stefan Lee, Devi Parikh, and Dhruv Batra.
\newblock Emergence of compositional language with deep generational
  transmission.
\newblock \emph{CoRR}, abs/1904.09067, 2019.
\newblock URL \url{http://arxiv.org/abs/1904.09067}.

\bibitem[Grover et~al.(2018)Grover, Al{-}Shedivat, Gupta, Burda, and
  Edwards]{Grover2018}
Aditya Grover, Maruan Al{-}Shedivat, Jayesh~K. Gupta, Yura Burda, and Harrison
  Edwards.
\newblock Learning policy representations in multiagent systems.
\newblock \emph{CoRR}, abs/1806.06464, 2018.
\newblock URL \url{http://arxiv.org/abs/1806.06464}.

\bibitem[Mordatch and Abbeel(2018)]{Mordatch2018}
Igor Mordatch and Pieter Abbeel.
\newblock Emergence of grounded compositional language in multi-agent
  populations.
\newblock In \emph{Thirty-Second AAAI Conference on Artificial Intelligence},
  2018.

\bibitem[Das et~al.(2019)Das, Gervet, Romoff, Batra, Parikh, Rabbat, and
  Pineau]{Das2019}
Abhishek Das, Th{\'e}ophile Gervet, Joshua Romoff, Dhruv Batra, Devi Parikh,
  Mike Rabbat, and Joelle Pineau.
\newblock Tarmac: Targeted multi-agent communication.
\newblock In \emph{International Conference on Machine Learning}, pages
  1538--1546. PMLR, 2019.

\bibitem[Lowe et~al.(2017)Lowe, Wu, Tamar, Harb, Abbeel, and
  Mordatch]{Lowe2017}
Ryan Lowe, Yi~Wu, Aviv Tamar, Jean Harb, Pieter Abbeel, and Igor Mordatch.
\newblock Multi-agent actor-critic for mixed cooperative-competitive
  environments.
\newblock In Isabelle Guyon, Ulrike von Luxburg, Samy Bengio, Hanna~M. Wallach,
  Rob Fergus, S.~V.~N. Vishwanathan, and Roman Garnett, editors, \emph{Advances
  in Neural Information Processing Systems 30: Annual Conference on Neural
  Information Processing Systems 2017, December 4-9, 2017, Long Beach, CA,
  {USA}}, pages 6379--6390, 2017.
\newblock URL
  \url{https://proceedings.neurips.cc/paper/2017/hash/68a9750337a418a86fe06c1991a1d64c-Abstract.html}.

\bibitem[Chaabouni et~al.(2019{\natexlab{a}})Chaabouni, Kharitonov, Lazaric,
  Dupoux, and Baroni]{Chaabouni2019}
Rahma Chaabouni, Eugene Kharitonov, Alessandro Lazaric, Emmanuel Dupoux, and
  Marco Baroni.
\newblock Word-order biases in deep-agent emergent communication.
\newblock In Anna Korhonen, David~R. Traum, and Llu{\'{\i}}s M{\`{a}}rquez,
  editors, \emph{Proceedings of the 57th Conference of the Association for
  Computational Linguistics, {ACL} 2019, Florence, Italy, July 28- August 2,
  2019, Volume 1: Long Papers}, pages 5166--5175. Association for Computational
  Linguistics, 2019{\natexlab{a}}.
\newblock \doi{10.18653/v1/p19-1509}.
\newblock URL \url{https://doi.org/10.18653/v1/p19-1509}.

\bibitem[Zhu et~al.(2021)Zhu, Neubig, and Bisk]{abs-2107-05697}
Hao Zhu, Graham Neubig, and Yonatan Bisk.
\newblock Few-shot language coordination by modeling theory of mind.
\newblock \emph{CoRR}, abs/2107.05697, 2021.
\newblock URL \url{https://arxiv.org/abs/2107.05697}.

\bibitem[Bachrach et~al.(2020)Bachrach, Everett, Hughes, Lazaridou, Leibo,
  Lanctot, Johanson, Czarnecki, and Graepel]{Bachrach2020}
Yoram Bachrach, Richard Everett, Edward Hughes, Angeliki Lazaridou, Joel~Z
  Leibo, Marc Lanctot, Michael Johanson, Wojciech~M Czarnecki, and Thore
  Graepel.
\newblock Negotiating team formation using deep reinforcement learning.
\newblock \emph{Artificial Intelligence}, 288:\penalty0 103356, 2020.

\bibitem[Cao et~al.(2018)Cao, Lazaridou, Lanctot, Leibo, Tuyls, and
  Clark]{Cao2018}
Kris Cao, Angeliki Lazaridou, Marc Lanctot, Joel~Z. Leibo, Karl Tuyls, and
  Stephen Clark.
\newblock Emergent communication through negotiation.
\newblock \emph{CoRR}, abs/1804.03980, 2018.
\newblock URL \url{http://arxiv.org/abs/1804.03980}.

\bibitem[Chen et~al.(2020)Chen, Xu, Liu, Li, and Zhao]{abs-2005-05441}
Baiming Chen, Mengdi Xu, Zuxin Liu, Liang Li, and Ding Zhao.
\newblock Delay-aware multi-agent reinforcement learning.
\newblock \emph{CoRR}, abs/2005.05441, 2020.
\newblock URL \url{https://arxiv.org/abs/2005.05441}.

\bibitem[Brandizzi et~al.(2021)Brandizzi, Grossi, and Iocchi]{Brandizzi2021}
Nicolo' Brandizzi, Davide Grossi, and Luca Iocchi.
\newblock Rlupus: Cooperation through emergent communication in the werewolf
  social deduction game.
\newblock \emph{Intelligenza Artificiale}, 15:\penalty0 55--70, 2021.
\newblock ISSN 2211-0097.
\newblock \doi{10.3233/IA-210081}.
\newblock URL \url{https://doi.org/10.3233/IA-210081}.
\newblock 2.

\bibitem[{Nakamura} et~al.(2016){Nakamura}, {Inaba}, {Takahashi}, {Toriumi},
  {Osawa}, {Katagami}, and {Shinoda}]{Nakamura2016}
N.~{Nakamura}, M.~{Inaba}, K.~{Takahashi}, F.~{Toriumi}, H.~{Osawa},
  D.~{Katagami}, and K.~{Shinoda}.
\newblock Constructing a human-like agent for the werewolf game using a
  psychological model based multiple perspectives.
\newblock In \emph{2016 IEEE Symposium Series on Computational Intelligence
  (SSCI)}, pages 1--8, 2016.

\bibitem[Jaques et~al.(2018)Jaques, Lazaridou, Hughes, G{\"{u}}l{\c{c}}ehre,
  Ortega, Strouse, Leibo, and de~Freitas]{abs-1810-08647}
Natasha Jaques, Angeliki Lazaridou, Edward Hughes, {\c{C}}aglar
  G{\"{u}}l{\c{c}}ehre, Pedro~A. Ortega, DJ~Strouse, Joel~Z. Leibo, and Nando
  de~Freitas.
\newblock Intrinsic social motivation via causal influence in multi-agent {RL}.
\newblock \emph{CoRR}, abs/1810.08647, 2018.
\newblock URL \url{http://arxiv.org/abs/1810.08647}.

\bibitem[Lazaridou et~al.(2018)Lazaridou, Hermann, Tuyls, and
  Clark]{Lazaridou2018}
Angeliki Lazaridou, Karl~Moritz Hermann, Karl Tuyls, and Stephen Clark.
\newblock Emergence of linguistic communication from referential games with
  symbolic and pixel input.
\newblock In \emph{6th International Conference on Learning Representations,
  {ICLR} 2018, Vancouver, BC, Canada, April 30 - May 3, 2018, Conference Track
  Proceedings}. OpenReview.net, 2018.
\newblock URL \url{https://openreview.net/forum?id=HJGv1Z-AW}.

\bibitem[Guo et~al.(2019)Guo, Ren, Havrylov, Frank, Titov, and
  Smith]{abs-1910-05291}
Shangmin Guo, Yi~Ren, Serhii Havrylov, Stella Frank, Ivan Titov, and Kenny
  Smith.
\newblock The emergence of compositional languages for numeric concepts through
  iterated learning in neural agents.
\newblock \emph{CoRR}, abs/1910.05291, 2019.
\newblock URL \url{http://arxiv.org/abs/1910.05291}.

\bibitem[Denamgana{ï} and Walker(2020)]{abs-2012-10776}
Kevin Denamgana{ï} and James~Alfred Walker.
\newblock On (emergent) systematic generalisation and compositionality in
  visual referential games with straight-through gumbel-softmax estimator.
\newblock \emph{CoRR}, abs/2012.10776, 2020.
\newblock URL \url{https://arxiv.org/abs/2012.10776}.

\bibitem[Higgins et~al.(2017)Higgins, Matthey, Pal, Burgess, Glorot, Botvinick,
  Mohamed, and Lerchner]{HigginsMPBGBML17}
Irina Higgins, Lo{\"{\i}}c Matthey, Arka Pal, Christopher~P. Burgess, Xavier
  Glorot, Matthew~M. Botvinick, Shakir Mohamed, and Alexander Lerchner.
\newblock beta-vae: Learning basic visual concepts with a constrained
  variational framework.
\newblock In \emph{5th International Conference on Learning Representations,
  {ICLR} 2017, Toulon, France, April 24-26, 2017, Conference Track
  Proceedings}. OpenReview.net, 2017.
\newblock URL \url{https://openreview.net/forum?id=Sy2fzU9gl}.

\bibitem[Anderson(1991)]{anderson1991adaptive}
John~R Anderson.
\newblock The adaptive nature of human categorization.
\newblock \emph{Psychological review}, 98\penalty0 (3):\penalty0 409, 1991.

\bibitem[Wierzbicka(1984)]{wierzbicka1984apples}
Anna Wierzbicka.
\newblock Apples are not a “kind of fruit”: The semantics of human
  categorization.
\newblock \emph{American Ethnologist}, 11\penalty0 (2):\penalty0 313--328,
  1984.

\bibitem[Merriman(1991)]{markman1989categorization}
William~E. Merriman.
\newblock Categorization and naming in children: Problems of induction. ellen
  markman. cambridge, ma: Mit press, 1989. pp. 250.
\newblock \emph{Applied Psycholinguistics}, 12\penalty0 (3):\penalty0
  385–392, 1991.
\newblock \doi{10.1017/S0142716400009310}.

\bibitem[Zaslavsky et~al.(2018)Zaslavsky, Kemp, Regier, and
  Tishby]{zaslavsky2018efficient}
Noga Zaslavsky, Charles Kemp, Terry Regier, and Naftali Tishby.
\newblock Efficient compression in color naming and its evolution.
\newblock \emph{Proceedings of the National Academy of Sciences}, 115\penalty0
  (31):\penalty0 7937--7942, 2018.

\bibitem[Tishby et~al.(2000)Tishby, Pereira, and Bialek]{physics-0004057}
Naftali Tishby, Fernando C.~N. Pereira, and William Bialek.
\newblock The information bottleneck method.
\newblock \emph{CoRR}, physics/0004057, 2000.
\newblock URL \url{http://arxiv.org/abs/physics/0004057}.

\bibitem[Mao et~al.(2019)Mao, Zhang, Xiao, Gong, and Ni]{abs-1912-05304}
Hangyu Mao, Zhengchao Zhang, Zhen Xiao, Zhibo Gong, and Yan Ni.
\newblock Learning agent communication under limited bandwidth by message
  pruning.
\newblock \emph{CoRR}, abs/1912.05304, 2019.
\newblock URL \url{http://arxiv.org/abs/1912.05304}.

\bibitem[Wang et~al.(2019)Wang, He, Yu, Qiu, An, and
  Rabinovich]{abs-1911-06992}
Rundong Wang, Xu~He, Runsheng Yu, Wei Qiu, Bo~An, and Zinovi Rabinovich.
\newblock Learning efficient multi-agent communication: An information
  bottleneck approach.
\newblock \emph{CoRR}, abs/1911.06992, 2019.
\newblock URL \url{http://arxiv.org/abs/1911.06992}.

\bibitem[Kharitonov et~al.(2020)Kharitonov, Chaabouni, Bouchacourt, and
  Baroni]{KharitonovCBB20}
Eugene Kharitonov, Rahma Chaabouni, Diane Bouchacourt, and Marco Baroni.
\newblock Entropy minimization in emergent languages.
\newblock In \emph{Proceedings of the 37th International Conference on Machine
  Learning, {ICML} 2020, 13-18 July 2020, Virtual Event}, volume 119 of
  \emph{Proceedings of Machine Learning Research}, pages 5220--5230. {PMLR},
  2020.
\newblock URL \url{http://proceedings.mlr.press/v119/kharitonov20a.html}.

\bibitem[Kirby et~al.(2015)Kirby, Tamariz, Cornish, and
  Smith]{kirby2015compression}
Simon Kirby, Monica Tamariz, Hannah Cornish, and Kenny Smith.
\newblock Compression and communication in the cultural evolution of linguistic
  structure.
\newblock \emph{Cognition}, 141:\penalty0 87--102, 2015.

\bibitem[Resnick et~al.(2020)Resnick, Gupta, Foerster, Dai, and
  Cho]{Resnick2019}
Cinjon Resnick, Abhinav Gupta, Jakob~N. Foerster, Andrew~M. Dai, and Kyunghyun
  Cho.
\newblock Capacity, bandwidth, and compositionality in emergent language
  learning.
\newblock In Amal El~Fallah Seghrouchni, Gita Sukthankar, Bo~An, and Neil
  Yorke{-}Smith, editors, \emph{Proceedings of the 19th International
  Conference on Autonomous Agents and Multiagent Systems, {AAMAS} '20,
  Auckland, New Zealand, May 9-13, 2020}, pages 1125--1133. International
  Foundation for Autonomous Agents and Multiagent Systems, 2020.
\newblock \doi{10.5555/3398761.3398892}.
\newblock URL \url{https://dl.acm.org/doi/10.5555/3398761.3398892}.

\bibitem[Tucker et~al.(2022)Tucker, Levy, Shah, and
  Zaslavsky]{NEURIPS2022_8bb5f663}
Mycal Tucker, Roger Levy, Julie~A Shah, and Noga Zaslavsky.
\newblock Trading off utility, informativeness, and complexity in emergent
  communication.
\newblock In S.~Koyejo, S.~Mohamed, A.~Agarwal, D.~Belgrave, K.~Cho, and A.~Oh,
  editors, \emph{Advances in Neural Information Processing Systems}, volume~35,
  pages 22214--22228. Curran Associates, Inc., 2022.
\newblock URL
  \url{https://proceedings.neurips.cc/paper_files/paper/2022/file/8bb5f66371c7e4cbf6c223162c62c0f4-Paper-Conference.pdf}.

\bibitem[Salakhutdinov and Hinton(2009)]{SalakhutdinovH09}
Ruslan Salakhutdinov and Geoffrey~E. Hinton.
\newblock Semantic hashing.
\newblock \emph{International Journal of Approximate Reasoning}, 50\penalty0
  (7):\penalty0 969--978, 2009.
\newblock \doi{10.1016/j.ijar.2008.11.006}.
\newblock URL \url{https://doi.org/10.1016/j.ijar.2008.11.006}.

\bibitem[Kaiser and Bengio(2018)]{abs-1801-09797}
Lukasz Kaiser and Samy Bengio.
\newblock Discrete autoencoders for sequence models.
\newblock \emph{CoRR}, abs/1801.09797, 2018.
\newblock URL \url{http://arxiv.org/abs/1801.09797}.

\bibitem[Jang et~al.(2017)Jang, Gu, and Poole]{Jang2016}
Eric Jang, Shixiang Gu, and Ben Poole.
\newblock Categorical reparameterization with gumbel-softmax.
\newblock In \emph{5th International Conference on Learning Representations,
  {ICLR} 2017, Toulon, France, April 24-26, 2017, Conference Track
  Proceedings}. OpenReview.net, 2017.
\newblock URL \url{https://openreview.net/forum?id=rkE3y85ee}.

\bibitem[Maddison et~al.(2017)Maddison, Mnih, and Teh]{Maddison2016}
Chris~J. Maddison, Andriy Mnih, and Yee~Whye Teh.
\newblock The concrete distribution: {A} continuous relaxation of discrete
  random variables.
\newblock In \emph{5th International Conference on Learning Representations,
  {ICLR} 2017, Toulon, France, April 24-26, 2017, Conference Track
  Proceedings}. OpenReview.net, 2017.
\newblock URL \url{https://openreview.net/forum?id=S1jE5L5gl}.

\bibitem[Williams(1992)]{williams1992simple}
Ronald~J Williams.
\newblock Simple statistical gradient-following algorithms for connectionist
  reinforcement learning.
\newblock \emph{Machine learning}, 8\penalty0 (3-4):\penalty0 229--256, 1992.

\bibitem[Mnih and Gregor(2014)]{MnihG14}
Andriy Mnih and Karol Gregor.
\newblock Neural variational inference and learning in belief networks.
\newblock In \emph{Proceedings of the 31th International Conference on Machine
  Learning, {ICML} 2014, Beijing, China, 21-26 June 2014}, volume~32 of
  \emph{{JMLR} Workshop and Conference Proceedings}, pages 1791--1799.
  JMLR.org, 2014.
\newblock URL \url{http://proceedings.mlr.press/v32/mnih14.html}.

\bibitem[Gu et~al.(2016)Gu, Levine, Sutskever, and Mnih]{GuLSM15}
Shixiang Gu, Sergey Levine, Ilya Sutskever, and Andriy Mnih.
\newblock Muprop: Unbiased backpropagation for stochastic neural networks.
\newblock In Yoshua Bengio and Yann LeCun, editors, \emph{4th International
  Conference on Learning Representations, {ICLR} 2016, San Juan, Puerto Rico,
  May 2-4, 2016, Conference Track Proceedings}, 2016.
\newblock URL \url{http://arxiv.org/abs/1511.05176}.

\bibitem[Bengio et~al.(2013)Bengio, L{\'{e}}onard, and Courville]{BengioLC13}
Yoshua Bengio, Nicholas L{\'{e}}onard, and Aaron~C. Courville.
\newblock Estimating or propagating gradients through stochastic neurons for
  conditional computation.
\newblock \emph{CoRR}, abs/1308.3432, 2013.
\newblock URL \url{http://arxiv.org/abs/1308.3432}.

\bibitem[Omidshafiei et~al.(2019)Omidshafiei, Kim, Liu, Tesauro, Riemer, Amato,
  Campbell, and How]{Omidshafiei2019}
Shayegan Omidshafiei, Dong-Ki Kim, Miao Liu, Gerald Tesauro, Matthew Riemer,
  Christopher Amato, Murray Campbell, and Jonathan~P How.
\newblock Learning to teach in cooperative multiagent reinforcement learning.
\newblock In \emph{Proceedings of the AAAI Conference on Artificial
  Intelligence}, volume~33, pages 6128--6136, 2019.

\bibitem[Rita et~al.(2022)Rita, Strub, Grill, Pietquin, and
  Dupoux]{abs-2204-12982}
Mathieu Rita, Florian Strub, Jean{-}Bastien Grill, Olivier Pietquin, and
  Emmanuel Dupoux.
\newblock On the role of population heterogeneity in emergent communication.
\newblock \emph{CoRR}, abs/2204.12982, 2022.
\newblock \doi{10.48550/arXiv.2204.12982}.
\newblock URL \url{https://doi.org/10.48550/arXiv.2204.12982}.

\bibitem[Buzs{\'a}ki and Tingley(2018)]{buzsaki2018space}
Gy{\"o}rgy Buzs{\'a}ki and David Tingley.
\newblock Space and time: the hippocampus as a sequence generator.
\newblock \emph{Trends in cognitive sciences}, 22\penalty0 (10):\penalty0
  853--869, 2018.

\bibitem[Gentner et~al.(2001)Gentner, Holyoak, and
  Kokinov]{gentner2001analogical}
Dedre Gentner, Keith~J Holyoak, and Boicho~N Kokinov.
\newblock \emph{The analogical mind: Perspectives from cognitive science}.
\newblock MIT press, 2001.

\bibitem[Rumjaun and Narod(2020)]{rumjaun2020social}
Anwar Rumjaun and Fawzia Narod.
\newblock Social learning theory—albert bandura.
\newblock \emph{Science education in theory and practice: An introductory guide
  to learning theory}, pages 85--99, 2020.

\bibitem[Cogswell et~al.(2020)Cogswell, Lu, Jain, Lee, Parikh, and
  Batra]{Cogswell2020}
Michael Cogswell, Jiasen Lu, Rishabh Jain, Stefan Lee, Devi Parikh, and Dhruv
  Batra.
\newblock Dialog without dialog data: Learning visual dialog agents from {VQA}
  data.
\newblock In Hugo Larochelle, Marc'Aurelio Ranzato, Raia Hadsell,
  Maria{-}Florina Balcan, and Hsuan{-}Tien Lin, editors, \emph{Advances in
  Neural Information Processing Systems 33: Annual Conference on Neural
  Information Processing Systems 2020, NeurIPS 2020, December 6-12, 2020,
  virtual}, 2020.
\newblock URL
  \url{https://proceedings.neurips.cc/paper/2020/hash/e7023ba77a45f7e84c5ee8a28dd63585-Abstract.html}.

\bibitem[Li et~al.(2020)Li, Ponti, Vulic, and Korhonen]{Li2020}
Yaoyiran Li, Edoardo~Maria Ponti, Ivan Vulic, and Anna Korhonen.
\newblock Emergent communication pretraining for few-shot machine translation.
\newblock In Donia Scott, N{\'{u}}ria Bel, and Chengqing Zong, editors,
  \emph{Proceedings of the 28th International Conference on Computational
  Linguistics, {COLING} 2020, Barcelona, Spain (Online), December 8-13, 2020},
  pages 4716--4731. International Committee on Computational Linguistics, 2020.
\newblock \doi{10.18653/v1/2020.coling-main.416}.
\newblock URL \url{https://doi.org/10.18653/v1/2020.coling-main.416}.

\bibitem[Lu et~al.(2020{\natexlab{a}})Lu, Singhal, Strub, Pietquin, and
  Courville]{abs-2003-12694}
Yuchen Lu, Soumye Singhal, Florian Strub, Olivier Pietquin, and Aaron~C.
  Courville.
\newblock Countering language drift with seeded iterated learning.
\newblock \emph{CoRR}, abs/2003.12694, 2020{\natexlab{a}}.
\newblock URL \url{https://arxiv.org/abs/2003.12694}.

\bibitem[Hawkins et~al.(2020)Hawkins, Kwon, Sadigh, and Goodman]{HawkinsKSG20}
Robert X.~D. Hawkins, Minae Kwon, Dorsa Sadigh, and Noah~D. Goodman.
\newblock Continual adaptation for efficient machine communication.
\newblock In Raquel Fern{\'{a}}ndez and Tal Linzen, editors, \emph{Proceedings
  of the 24th Conference on Computational Natural Language Learning, CoNLL
  2020, Online, November 19-20, 2020}, pages 408--419. Association for
  Computational Linguistics, 2020.
\newblock \doi{10.18653/v1/2020.conll-1.33}.
\newblock URL \url{https://doi.org/10.18653/v1/2020.conll-1.33}.

\bibitem[Lu et~al.(2020{\natexlab{b}})Lu, Singhal, Strub, Pietquin, and
  Courville]{abs-2010-02975}
Yuchen Lu, Soumye Singhal, Florian Strub, Olivier Pietquin, and Aaron~C.
  Courville.
\newblock Supervised seeded iterated learning for interactive language
  learning.
\newblock \emph{CoRR}, abs/2010.02975, 2020{\natexlab{b}}.
\newblock URL \url{https://arxiv.org/abs/2010.02975}.

\bibitem[Lazaridou et~al.(2020)Lazaridou, Potapenko, and
  Tieleman]{Lazaridou2020}
Angeliki Lazaridou, Anna Potapenko, and Olivier Tieleman.
\newblock Multi-agent communication meets natural language: Synergies between
  functional and structural language learning.
\newblock In Dan Jurafsky, Joyce Chai, Natalie Schluter, and Joel~R. Tetreault,
  editors, \emph{Proceedings of the 58th Annual Meeting of the Association for
  Computational Linguistics, {ACL} 2020, Online, July 5-10, 2020}, pages
  7663--7674. Association for Computational Linguistics, 2020.
\newblock \doi{10.18653/v1/2020.acl-main.685}.
\newblock URL \url{https://doi.org/10.18653/v1/2020.acl-main.685}.

\bibitem[Raileanu et~al.(2018)Raileanu, Denton, Szlam, and
  Fergus]{Raileanu2018}
Roberta Raileanu, Emily Denton, Arthur Szlam, and Rob Fergus.
\newblock Modeling others using oneself in multi-agent reinforcement learning.
\newblock \emph{CoRR}, abs/1802.09640, 2018.
\newblock URL \url{http://arxiv.org/abs/1802.09640}.

\bibitem[Jaques et~al.(2019)Jaques, Lazaridou, Hughes, G{\"{u}}l{\c{c}}ehre,
  Ortega, Strouse, Leibo, and de~Freitas]{JaquesLHGOSLF19}
Natasha Jaques, Angeliki Lazaridou, Edward Hughes, {\c{C}}aglar
  G{\"{u}}l{\c{c}}ehre, Pedro~A. Ortega, DJ~Strouse, Joel~Z. Leibo, and Nando
  de~Freitas.
\newblock Social influence as intrinsic motivation for multi-agent deep
  reinforcement learning.
\newblock In Kamalika Chaudhuri and Ruslan Salakhutdinov, editors,
  \emph{Proceedings of the 36th International Conference on Machine Learning,
  {ICML} 2019, 9-15 June 2019, Long Beach, California, {USA}}, volume~97 of
  \emph{Proceedings of Machine Learning Research}, pages 3040--3049. {PMLR},
  2019.
\newblock URL \url{http://proceedings.mlr.press/v97/jaques19a.html}.

\bibitem[Choi et~al.(2018)Choi, Lazaridou, and de~Freitas]{abs-1804-02341}
Edward Choi, Angeliki Lazaridou, and Nando de~Freitas.
\newblock Compositional obverter communication learning from raw visual input.
\newblock \emph{CoRR}, abs/1804.02341, 2018.
\newblock URL \url{http://arxiv.org/abs/1804.02341}.

\bibitem[Bogin et~al.(2018)Bogin, Geva, and Berant]{abs-1809-00549}
Ben Bogin, Mor Geva, and Jonathan Berant.
\newblock Emergence of communication in an interactive world with consistent
  speakers.
\newblock \emph{CoRR}, abs/1809.00549, 2018.
\newblock URL \url{http://arxiv.org/abs/1809.00549}.

\bibitem[Dess{ì} et~al.(2021)Dess{ì}, Kharitonov, and Baroni]{abs-2106-04258}
Roberto Dess{ì}, Eugene Kharitonov, and Marco Baroni.
\newblock Interpretable agent communication from scratch(with a generic visual
  processor emerging on the side).
\newblock \emph{CoRR}, abs/2106.04258, 2021.
\newblock URL \url{https://arxiv.org/abs/2106.04258}.

\bibitem[Smith(2010)]{Smith2010}
Eric~Alden Smith.
\newblock Communication and collective action: language and the evolution of
  human cooperation.
\newblock \emph{Evolution and human behavior}, 31\penalty0 (4):\penalty0
  231--245, 2010.

\bibitem[Nowak and Krakauer(1999)]{Nowak1999}
Martin~A Nowak and David~C Krakauer.
\newblock The evolution of language.
\newblock \emph{Proceedings of the National Academy of Sciences}, 96\penalty0
  (14):\penalty0 8028--8033, 1999.

\bibitem[Evtimova et~al.(2018)Evtimova, Drozdov, Kiela, and Cho]{Evtimova2017}
Katrina Evtimova, Andrew Drozdov, Douwe Kiela, and Kyunghyun Cho.
\newblock Emergent communication in a multi-modal, multi-step referential game.
\newblock In \emph{6th International Conference on Learning Representations,
  {ICLR} 2018, Vancouver, BC, Canada, April 30 - May 3, 2018, Conference Track
  Proceedings}. OpenReview.net, 2018.
\newblock URL \url{https://openreview.net/forum?id=rJGZq6g0-}.

\bibitem[Briscoe(2002)]{Briscoe2002}
Ted Briscoe.
\newblock \emph{Linguistic Evolution through Language Acquisition}.
\newblock Cambridge University Press, 2002.
\newblock \doi{10.1017/CBO9780511486524}.

\bibitem[Kirby et~al.(2014)Kirby, Griffiths, and Smith]{Kirby2014}
Simon Kirby, Tom Griffiths, and Kenny Smith.
\newblock Iterated learning and the evolution of language.
\newblock \emph{Current opinion in neurobiology}, 28:\penalty0 108--114, 2014.

\bibitem[Tieleman et~al.(2019)Tieleman, Lazaridou, Mourad, Blundell, and
  Precup]{abs-1912-06208}
Olivier Tieleman, Angeliki Lazaridou, Shibl Mourad, Charles Blundell, and Doina
  Precup.
\newblock Shaping representations through communication: community size effect
  in artificial learning systems.
\newblock \emph{CoRR}, abs/1912.06208, 2019.
\newblock URL \url{http://arxiv.org/abs/1912.06208}.

\bibitem[Lowe et~al.(2019{\natexlab{a}})Lowe, Gupta, Foerster, Kiela, and
  Pineau]{Lowe2019}
Ryan Lowe, Abhinav Gupta, Jakob Foerster, Douwe Kiela, and Joelle Pineau.
\newblock Learning to learn to communicate, 2019{\natexlab{a}}.

\bibitem[Fitzgerald(2019)]{Fitzgerald2019}
Nicole Fitzgerald.
\newblock To populate is to regulate.
\newblock \emph{CoRR}, abs/1911.04362, 2019.
\newblock URL \url{http://arxiv.org/abs/1911.04362}.

\bibitem[Kirby(2001)]{kirby2001spontaneous}
Simon Kirby.
\newblock Spontaneous evolution of linguistic structure-an iterated learning
  model of the emergence of regularity and irregularity.
\newblock \emph{IEEE Transactions on Evolutionary Computation}, 5\penalty0
  (2):\penalty0 102--110, 2001.

\bibitem[Scott-Phillips and Kirby(2010)]{ScottPhillips2010}
Thomas~C Scott-Phillips and Simon Kirby.
\newblock Language evolution in the laboratory.
\newblock \emph{Trends in cognitive sciences}, 14\penalty0 (9):\penalty0
  411--417, 2010.

\bibitem[Ren et~al.(2020)Ren, Guo, Labeau, Cohen, and Kirby]{abs-2002-01365}
Yi~Ren, Shangmin Guo, Matthieu Labeau, Shay~B. Cohen, and Simon Kirby.
\newblock Compositional languages emerge in a neural iterated learning model.
\newblock In \emph{8th International Conference on Learning Representations,
  {ICLR} 2020, Addis Ababa, Ethiopia, April 26-30, 2020}. OpenReview.net, 2020.
\newblock URL \url{https://openreview.net/forum?id=HkePNpVKPB}.

\bibitem[Zhou et~al.(2022)Zhou, Vani, Larochelle, and Courville]{ZhouVLC22}
Hattie Zhou, Ankit Vani, Hugo Larochelle, and Aaron~C. Courville.
\newblock Fortuitous forgetting in connectionist networks.
\newblock In \emph{The Tenth International Conference on Learning
  Representations, {ICLR} 2022, Virtual Event, April 25-29, 2022}.
  OpenReview.net, 2022.
\newblock URL \url{https://openreview.net/forum?id=ei3SY1\_zYsE}.

\bibitem[Frankle and Carbin(2018)]{abs-1803-03635}
Jonathan Frankle and Michael Carbin.
\newblock The lottery ticket hypothesis: Training pruned neural networks.
\newblock \emph{CoRR}, abs/1803.03635, 2018.
\newblock URL \url{http://arxiv.org/abs/1803.03635}.

\bibitem[Barrett and Zollman(2009)]{barrett2009role}
Jeffrey Barrett and Kevin~JS Zollman.
\newblock The role of forgetting in the evolution and learning of language.
\newblock \emph{Journal of Experimental \& Theoretical Artificial
  Intelligence}, 21\penalty0 (4):\penalty0 293--309, 2009.

\bibitem[da~Silva et~al.(2020)da~Silva, Warnell, Costa, and Stone]{SilvaWCS20}
Felipe~Leno da~Silva, Garrett Warnell, Anna Helena~Reali Costa, and Peter
  Stone.
\newblock Agents teaching agents: a survey on inter-agent transfer learning.
\newblock \emph{Autonomous Agents and Multi-Agent Systems}, 34\penalty0
  (1):\penalty0 9, 2020.
\newblock \doi{10.1007/s10458-019-09430-0}.
\newblock URL \url{https://doi.org/10.1007/s10458-019-09430-0}.

\bibitem[Tesauro(1994)]{Tesauro94}
Gerald Tesauro.
\newblock Td-gammon, a self-teaching backgammon program, achieves master-level
  play.
\newblock \emph{Neural computation}, 6\penalty0 (2):\penalty0 215--219, 1994.
\newblock \doi{10.1162/neco.1994.6.2.215}.
\newblock URL \url{https://doi.org/10.1162/neco.1994.6.2.215}.

\bibitem[Lowe et~al.(2020)Lowe, Gupta, Foerster, Kiela, and Pineau]{Lowe2020}
Ryan Lowe, Abhinav Gupta, Jakob~N. Foerster, Douwe Kiela, and Joelle Pineau.
\newblock On the interaction between supervision and self-play in emergent
  communication.
\newblock In \emph{8th International Conference on Learning Representations,
  {ICLR} 2020, Addis Ababa, Ethiopia, April 26-30, 2020}. OpenReview.net, 2020.
\newblock URL \url{https://openreview.net/forum?id=rJxGLlBtwH}.

\bibitem[Gupta et~al.(2019)Gupta, Lowe, Foerster, Kiela, and
  Pineau]{GuptaLFKP19}
Abhinav Gupta, Ryan Lowe, Jakob~N. Foerster, Douwe Kiela, and Joelle Pineau.
\newblock Seeded self-play for language learning.
\newblock In Aditya Mogadala, Dietrich Klakow, Sandro Pezzelle, and
  Marie{-}Francine Moens, editors, \emph{Proceedings of the Beyond Vision and
  LANguage: inTEgrating Real-world kNowledge, LANTERN@EMNLP-IJCNLP 2019, Hong
  Kong, China, November 3, 2019}, pages 62--66. Association for Computational
  Linguistics, 2019.
\newblock \doi{10.18653/v1/D19-6409}.
\newblock URL \url{https://doi.org/10.18653/v1/D19-6409}.

\bibitem[Gopnik and Wellman(1992)]{Gopnik1992}
Alison Gopnik and Henry~M. Wellman.
\newblock Why the child's theory of mind really is a theory.
\newblock \emph{Mind \& Language}, 7\penalty0 (1-2):\penalty0 145--171, 1992.
\newblock \doi{https://doi.org/10.1111/j.1468-0017.1992.tb00202.x}.
\newblock URL
  \url{https://onlinelibrary.wiley.com/doi/abs/10.1111/j.1468-0017.1992.tb00202.x}.

\bibitem[Premack and Woodruff(1978)]{Premack1978}
David Premack and Guy Woodruff.
\newblock Does the chimpanzee have a theory of mind?
\newblock \emph{Behavioral and brain sciences}, 1\penalty0 (4):\penalty0
  515--526, 1978.

\bibitem[Rabinowitz et~al.(2018)Rabinowitz, Perbet, Song, Zhang, Eslami, and
  Botvinick]{abs-1802-07740}
Neil~C. Rabinowitz, Frank Perbet, H.~Francis Song, Chiyuan Zhang, S.~M.~Ali
  Eslami, and Matthew~M. Botvinick.
\newblock Machine theory of mind.
\newblock \emph{CoRR}, abs/1802.07740, 2018.
\newblock URL \url{http://arxiv.org/abs/1802.07740}.

\bibitem[Frank and Goodman(2012)]{frank2012predicting}
Michael~C Frank and Noah~D Goodman.
\newblock Predicting pragmatic reasoning in language games.
\newblock \emph{Science}, 336\penalty0 (6084):\penalty0 998--998, 2012.

\bibitem[Goodman and Frank(2016)]{goodman2016pragmatic}
Noah~D Goodman and Michael~C Frank.
\newblock Pragmatic language interpretation as probabilistic inference.
\newblock \emph{Trends in cognitive sciences}, 20\penalty0 (11):\penalty0
  818--829, 2016.

\bibitem[Andreas and Klein(2016)]{Andreas2016}
Jacob Andreas and Dan Klein.
\newblock Reasoning about pragmatics with neural listeners and speakers.
\newblock In Jian Su, Xavier Carreras, and Kevin Duh, editors,
  \emph{Proceedings of the 2016 Conference on Empirical Methods in Natural
  Language Processing, {EMNLP} 2016, Austin, Texas, USA, November 1-4, 2016},
  pages 1173--1182. The Association for Computational Linguistics, 2016.
\newblock \doi{10.18653/v1/d16-1125}.
\newblock URL \url{https://doi.org/10.18653/v1/d16-1125}.

\bibitem[Foerster et~al.(2018)Foerster, Chen, Al{-}Shedivat, Whiteson, Abbeel,
  and Mordatch]{Foerster2017}
Jakob~N. Foerster, Richard~Y. Chen, Maruan Al{-}Shedivat, Shimon Whiteson,
  Pieter Abbeel, and Igor Mordatch.
\newblock Learning with opponent-learning awareness.
\newblock In Elisabeth Andr{\'{e}}, Sven Koenig, Mehdi Dastani, and Gita
  Sukthankar, editors, \emph{Proceedings of the 17th International Conference
  on Autonomous Agents and MultiAgent Systems, {AAMAS} 2018, Stockholm, Sweden,
  July 10-15, 2018}, pages 122--130. International Foundation for Autonomous
  Agents and Multiagent Systems Richland, SC, {USA} / {ACM}, 2018.
\newblock URL \url{http://dl.acm.org/citation.cfm?id=3237408}.

\bibitem[Finn et~al.(2017)Finn, Abbeel, and Levine]{FinnAL17}
Chelsea Finn, Pieter Abbeel, and Sergey Levine.
\newblock Model-agnostic meta-learning for fast adaptation of deep networks.
\newblock In Doina Precup and Yee~Whye Teh, editors, \emph{Proceedings of the
  34th International Conference on Machine Learning, {ICML} 2017, Sydney, NSW,
  Australia, 6-11 August 2017}, volume~70 of \emph{Proceedings of Machine
  Learning Research}, pages 1126--1135. {PMLR}, 2017.
\newblock URL \url{http://proceedings.mlr.press/v70/finn17a.html}.

\bibitem[Xie et~al.(2020)Xie, Losey, Tolsma, Finn, and Sadigh]{Xie2020}
Annie Xie, Dylan~P. Losey, Ryan Tolsma, Chelsea Finn, and Dorsa Sadigh.
\newblock Learning latent representations to influence multi-agent interaction.
\newblock In Jens Kober, Fabio Ramos, and Claire~J. Tomlin, editors, \emph{4th
  Conference on Robot Learning, CoRL 2020, 16-18 November 2020, Virtual Event /
  Cambridge, MA, {USA}}, volume 155 of \emph{Proceedings of Machine Learning
  Research}, pages 575--588. {PMLR}, 2020.
\newblock URL \url{https://proceedings.mlr.press/v155/xie21a.html}.

\bibitem[Brandizzi and Iocchi(2022)]{brandizzi2022emergent}
Nicolo' Brandizzi and Luca Iocchi.
\newblock Emergent communication in human-machine games.
\newblock In \emph{Emergent Communication Workshop at ICLR 2022}, 2022.
\newblock URL \url{https://openreview.net/forum?id=rqLgeQWCXZ9}.

\bibitem[Foerster et~al.(2016)Foerster, Assael, De~Freitas, and
  Whiteson]{Foerster2016}
Jakob Foerster, Ioannis~Alexandros Assael, Nando De~Freitas, and Shimon
  Whiteson.
\newblock Learning to communicate with deep multi-agent reinforcement learning.
\newblock In \emph{Advances in neural information processing systems}, pages
  2137--2145, 2016.

\bibitem[Sachan and Neubig(2018)]{SachanN18}
Devendra~Singh Sachan and Graham Neubig.
\newblock Parameter sharing methods for multilingual self-attentional
  translation models.
\newblock In Ondrej Bojar, Rajen Chatterjee, Christian Federmann, Mark Fishel,
  Yvette Graham, Barry Haddow, Matthias Huck, Antonio Jimeno{-}Yepes, Philipp
  Koehn, Christof Monz, Matteo Negri, Aur{\'{e}}lie N{\'{e}}v{\'{e}}ol,
  Mariana~L. Neves, Matt Post, Lucia Specia, Marco Turchi, and Karin Verspoor,
  editors, \emph{Proceedings of the Third Conference on Machine Translation:
  Research Papers, {WMT} 2018, Belgium, Brussels, October 31 - November 1,
  2018}, pages 261--271. Association for Computational Linguistics, 2018.
\newblock \doi{10.18653/v1/w18-6327}.
\newblock URL \url{https://doi.org/10.18653/v1/w18-6327}.

\bibitem[Mahaut et~al.(2023)Mahaut, Franzon, Dess{ì}, and
  Baroni]{abs-2302-08913}
Mateo Mahaut, Francesca Franzon, Roberto Dess{ì}, and Marco Baroni.
\newblock Referential communication in heterogeneous communities of pre-trained
  visual deep networks.
\newblock \emph{CoRR}, abs/2302.08913, 2023.
\newblock \doi{10.48550/arXiv.2302.08913}.
\newblock URL \url{https://doi.org/10.48550/arXiv.2302.08913}.

\bibitem[Sukhbaatar et~al.(2016)Sukhbaatar, Fergus, et~al.]{Sukhbaatar2016}
Sainbayar Sukhbaatar, Rob Fergus, et~al.
\newblock Learning multiagent communication with backpropagation.
\newblock \emph{Advances in neural information processing systems},
  29:\penalty0 2244--2252, 2016.

\bibitem[Kong et~al.(2017)Kong, Xin, Wang, and Hua]{Kong2017}
Xiangyu Kong, Bo~Xin, Yizhou Wang, and Gang Hua.
\newblock Collaborative deep reinforcement learning for joint object search.
\newblock In \emph{Proceedings of the IEEE Conference on Computer Vision and
  Pattern Recognition}, pages 1695--1704, 2017.

\bibitem[Bullard et~al.(2020)Bullard, Meier, Kiela, Pineau, and
  Foerster]{abs-2010-15896}
Kalesha Bullard, Franziska Meier, Douwe Kiela, Joelle Pineau, and Jakob~N.
  Foerster.
\newblock Exploring zero-shot emergent communication in embodied multi-agent
  populations.
\newblock \emph{CoRR}, abs/2010.15896, 2020.
\newblock URL \url{https://arxiv.org/abs/2010.15896}.

\bibitem[Mihai and Hare(2021)]{MihaiH21}
Daniela Mihai and Jonathon~S. Hare.
\newblock Learning to draw: Emergent communication through sketching.
\newblock In Marc'Aurelio Ranzato, Alina Beygelzimer, Yann~N. Dauphin, Percy
  Liang, and Jennifer~Wortman Vaughan, editors, \emph{Advances in Neural
  Information Processing Systems 34: Annual Conference on Neural Information
  Processing Systems 2021, NeurIPS 2021, December 6-14, 2021, virtual}, pages
  7153--7166, 2021.
\newblock URL
  \url{https://proceedings.neurips.cc/paper/2021/hash/39d0a8908fbe6c18039ea8227f827023-Abstract.html}.

\bibitem[Qiu et~al.(2021)Qiu, Xie, Fan, Gao, Zhu, and Zhu]{abs-2111-14210}
Shuwen Qiu, Sirui Xie, Lifeng Fan, Tao Gao, Song{-}Chun Zhu, and Yixin Zhu.
\newblock Emergent graphical conventions in a visual communication game.
\newblock \emph{CoRR}, abs/2111.14210, 2021.
\newblock URL \url{https://arxiv.org/abs/2111.14210}.

\bibitem[Mehrabian et~al.(1971)]{Mehrabian1971}
Albert Mehrabian et~al.
\newblock \emph{Silent messages}, volume~8.
\newblock Wadsworth Belmont, CA, 1971.

\bibitem[Mehrabian(2017)]{Mehrabian2017}
Albert Mehrabian.
\newblock \emph{Nonverbal communication}.
\newblock Routledge, 2017.

\bibitem[Vasconez et~al.(2019)Vasconez, Guevara, and Chee{ì}n]{VasconezGC19}
Juan~Pablo Vasconez, Leonardo Guevara, and Fernando Alfredo~Auat Chee{ì}n.
\newblock Social robot navigation based on {HRI} non-verbal communication: a
  case study on avocado harvesting.
\newblock In Chih{-}Cheng Hung and George~A. Papadopoulos, editors,
  \emph{Proceedings of the 34th {ACM/SIGAPP} Symposium on Applied Computing,
  {SAC} 2019, Limassol, Cyprus, April 8-12, 2019}, pages 957--960. {ACM}, 2019.
\newblock \doi{10.1145/3297280.3297569}.
\newblock URL \url{https://doi.org/10.1145/3297280.3297569}.

\bibitem[Bacim et~al.(2012)Bacim, Ragan, Stinson, Scerbo, and
  Bowman]{BacimRSSB12}
Felipe Bacim, Eric~D. Ragan, Cheryl Stinson, Siroberto Scerbo, and Doug~A.
  Bowman.
\newblock Collaborative navigation in virtual search and rescue.
\newblock In Mark Billinghurst, Joseph J.~LaViola Jr., and Anatole
  L{\'{e}}cuyer, editors, \emph{{IEEE} Symposium on 3D User Interfaces 2012,
  3DUI 2012, Costa Mesa, CA, USA, March 4-5, 2012}, pages 187--188. {IEEE}
  Computer Society, 2012.
\newblock \doi{10.1109/3DUI.2012.6184224}.
\newblock URL \url{https://doi.org/10.1109/3DUI.2012.6184224}.

\bibitem[Tucker et~al.(2021)Tucker, Li, Agrawal, Hughes, Sycara, Lewis, and
  Shah]{tucker2021emergent}
Mycal Tucker, Huao Li, Siddharth Agrawal, Dana Hughes, Katia Sycara, Michael
  Lewis, and Julie~A Shah.
\newblock Emergent discrete communication in semantic spaces.
\newblock \emph{Advances in Neural Information Processing Systems},
  34:\penalty0 10574--10586, 2021.

\bibitem[Steinert-Threlkeld(2020)]{steinert2020toward}
Shane Steinert-Threlkeld.
\newblock Toward the emergence of nontrivial compositionality.
\newblock \emph{Philosophy of Science}, 87\penalty0 (5):\penalty0 897--909,
  2020.

\bibitem[Frege(1892)]{frege1892begriff}
Gottlob Frege.
\newblock {\"U}ber begriff und gegenstand.
\newblock \emph{Vierteljahrsschrift f{\"u}r wissenschaftliche Philosophie},
  16\penalty0 (2), 1892.

\bibitem[Baroni(2020)]{baroni2020linguistic}
Marco Baroni.
\newblock Linguistic generalization and compositionality in modern artificial
  neural networks.
\newblock \emph{Philosophical Transactions of the Royal Society B},
  375\penalty0 (1791):\penalty0 20190307, 2020.

\bibitem[Pelletier(1994)]{pelletier1994principle}
Francis~Jeffry Pelletier.
\newblock The principle of semantic compositionality.
\newblock \emph{Topoi}, 13\penalty0 (1):\penalty0 11--24, 1994.

\bibitem[Janssen and Partee(1997)]{janssen1997compositionality}
Theo~MV Janssen and Barbara~H Partee.
\newblock Compositionality.
\newblock In \emph{Handbook of logic and language}, pages 417--473. Elsevier,
  1997.

\bibitem[Korbak et~al.(2020)Korbak, Zubek, and
  Raczaszek{-}Leonardi]{abs-2010-15058}
Tomasz Korbak, Julian Zubek, and Joanna Raczaszek{-}Leonardi.
\newblock Measuring non-trivial compositionality in emergent communication.
\newblock \emph{CoRR}, abs/2010.15058, 2020.
\newblock URL \url{https://arxiv.org/abs/2010.15058}.

\bibitem[Andreas(2019)]{andreas2019measuring}
Jacob Andreas.
\newblock Measuring compositionality in representation learning.
\newblock In \emph{7th International Conference on Learning Representations,
  {ICLR} 2019, New Orleans, LA, USA, May 6-9, 2019}. OpenReview.net, 2019.
\newblock URL \url{https://openreview.net/forum?id=HJz05o0qK7}.

\bibitem[Kharitonov and Baroni(2020)]{abs-2004-03420}
Eugene Kharitonov and Marco Baroni.
\newblock Emergent language generalization and acquisition speed are not tied
  to compositionality.
\newblock \emph{CoRR}, abs/2004.03420, 2020.
\newblock URL \url{https://arxiv.org/abs/2004.03420}.

\bibitem[Korbak et~al.(2019)Korbak, Zubek, Kucinski, Milos, and
  Raczaszek{-}Leonardi]{abs-1910-06079}
Tomasz Korbak, Julian Zubek, Lukasz Kucinski, Piotr Milos, and Joanna
  Raczaszek{-}Leonardi.
\newblock Developmentally motivated emergence of compositional communication
  via template transfer.
\newblock \emph{CoRR}, abs/1910.06079, 2019.
\newblock URL \url{http://arxiv.org/abs/1910.06079}.

\bibitem[Larsen-Freeman(1997)]{larsen1997chaos}
Diane Larsen-Freeman.
\newblock Chaos/complexity science and second language acquisition.
\newblock \emph{Applied linguistics}, 18\penalty0 (2):\penalty0 141--165, 1997.

\bibitem[Kuci{\'n}ski et~al.(2021)Kuci{\'n}ski, Korbak, Ko{\l}odziej, and
  Mi{\l}o{\'s}]{kucinski2021catalytic}
{\L}ukasz Kuci{\'n}ski, Tomasz Korbak, Pawe{\l} Ko{\l}odziej, and Piotr
  Mi{\l}o{\'s}.
\newblock Catalytic role of noise and necessity of inductive biases in the
  emergence of compositional communication.
\newblock \emph{Advances in Neural Information Processing Systems},
  34:\penalty0 23075--23088, 2021.

\bibitem[Bouchacourt and Baroni(2018)]{Bouchacourt2018}
Diane Bouchacourt and Marco Baroni.
\newblock How agents see things: On visual representations in an emergent
  language game.
\newblock In Ellen Riloff, David Chiang, Julia Hockenmaier, and Jun'ichi
  Tsujii, editors, \emph{Proceedings of the 2018 Conference on Empirical
  Methods in Natural Language Processing, Brussels, Belgium, October 31 -
  November 4, 2018}, pages 981--985. Association for Computational Linguistics,
  2018.
\newblock \doi{10.18653/v1/d18-1119}.
\newblock URL \url{https://doi.org/10.18653/v1/d18-1119}.

\bibitem[Bouchacourt and Baroni(2019)]{Bouchacourt2019}
Diane Bouchacourt and Marco Baroni.
\newblock Miss tools and mr fruit: Emergent communication in agents learning
  about object affordances.
\newblock In Anna Korhonen, David~R. Traum, and Llu{\'{\i}}s M{\`{a}}rquez,
  editors, \emph{Proceedings of the 57th Conference of the Association for
  Computational Linguistics, {ACL} 2019, Florence, Italy, July 28- August 2,
  2019, Volume 1: Long Papers}, pages 3909--3918. Association for Computational
  Linguistics, 2019.
\newblock \doi{10.18653/v1/p19-1380}.
\newblock URL \url{https://doi.org/10.18653/v1/p19-1380}.

\bibitem[Lowe et~al.(2019{\natexlab{b}})Lowe, Foerster, Boureau, Pineau, and
  Dauphin]{Lowe2019a}
Ryan Lowe, Jakob~N. Foerster, Y{-}Lan Boureau, Joelle Pineau, and Yann~N.
  Dauphin.
\newblock On the pitfalls of measuring emergent communication.
\newblock In Edith Elkind, Manuela Veloso, Noa Agmon, and Matthew~E. Taylor,
  editors, \emph{Proceedings of the 18th International Conference on Autonomous
  Agents and MultiAgent Systems, {AAMAS} '19, Montreal, QC, Canada, May 13-17,
  2019}, pages 693--701. International Foundation for Autonomous Agents and
  Multiagent Systems, 2019{\natexlab{b}}.
\newblock URL \url{http://dl.acm.org/citation.cfm?id=3331757}.

\bibitem[Dess{ì} et~al.(2019)Dess{ì}, Bouchacourt, Crepaldi, and
  Baroni]{abs-1911-01892}
Roberto Dess{ì}, Diane Bouchacourt, Davide Crepaldi, and Marco Baroni.
\newblock Focus on what's informative and ignore what's not: Communication
  strategies in a referential game.
\newblock \emph{CoRR}, abs/1911.01892, 2019.
\newblock URL \url{http://arxiv.org/abs/1911.01892}.

\bibitem[Chaabouni et~al.(2019{\natexlab{b}})Chaabouni, Kharitonov, Dupoux, and
  Baroni]{abs-1905-12561}
Rahma Chaabouni, Eugene Kharitonov, Emmanuel Dupoux, and Marco Baroni.
\newblock Anti-efficient encoding in emergent communication.
\newblock \emph{CoRR}, abs/1905.12561, 2019{\natexlab{b}}.
\newblock URL \url{http://arxiv.org/abs/1905.12561}.

\bibitem[Pearson(1901)]{pearson1901liii}
Karl Pearson.
\newblock Liii. on lines and planes of closest fit to systems of points in
  space.
\newblock \emph{The London, Edinburgh, and Dublin philosophical magazine and
  journal of science}, 2\penalty0 (11):\penalty0 559--572, 1901.

\bibitem[Hotelling(1933)]{hotelling1933analysis}
Harold Hotelling.
\newblock Analysis of a complex of statistical variables into principal
  components.
\newblock \emph{Journal of educational psychology}, 24\penalty0 (6):\penalty0
  417, 1933.

\bibitem[Van~der Maaten and Hinton(2008)]{van2008visualizing}
Laurens Van~der Maaten and Geoffrey Hinton.
\newblock Visualizing data using t-sne.
\newblock \emph{Journal of machine learning research}, 9\penalty0 (11), 2008.

\bibitem[Wattenberg et~al.(2016)Wattenberg, Vi{\'e}gas, and
  Johnson]{wattenberg2016use}
Martin Wattenberg, Fernanda Vi{\'e}gas, and Ian Johnson.
\newblock How to use t-sne effectively.
\newblock \emph{Distill}, 1\penalty0 (10):\penalty0 e2, 2016.

\bibitem[Zipf(2013)]{zipf2013psycho}
George~Kingsley Zipf.
\newblock \emph{The psycho-biology of language: An introduction to dynamic
  philology}.
\newblock Routledge, 2013.

\bibitem[Brighton and Kirby(2006)]{brighton2006understanding}
Henry Brighton and Simon Kirby.
\newblock Understanding linguistic evolution by visualizing the emergence of
  topographic mappings.
\newblock \emph{Artificial life}, 12\penalty0 (2):\penalty0 229--242, 2006.

\bibitem[Chaabouni et~al.(2022)Chaabouni, Strub, Altch{\'{e}}, Tarassov,
  Tallec, Davoodi, Mathewson, Tieleman, Lazaridou, and Piot]{ChaabouniSATTDM22}
Rahma Chaabouni, Florian Strub, Florent Altch{\'{e}}, Eugene Tarassov, Corentin
  Tallec, Elnaz Davoodi, Kory~Wallace Mathewson, Olivier Tieleman, Angeliki
  Lazaridou, and Bilal Piot.
\newblock Emergent communication at scale.
\newblock In \emph{The Tenth International Conference on Learning
  Representations, {ICLR} 2022, Virtual Event, April 25-29, 2022}.
  OpenReview.net, 2022.
\newblock URL \url{https://openreview.net/forum?id=AUGBfDIV9rL}.

\bibitem[Kriegeskorte et~al.(2008)Kriegeskorte, Mur, and
  Bandettini]{kriegeskorte2008representational}
Nikolaus Kriegeskorte, Marieke Mur, and Peter~A Bandettini.
\newblock Representational similarity analysis-connecting the branches of
  systems neuroscience.
\newblock \emph{Frontiers in systems neuroscience}, page~4, 2008.

\bibitem[Yao et~al.(2022)Yao, Yu, Zhang, Narasimhan, Tenenbaum, and
  Gan]{YaoYZNTG22}
Shunyu Yao, Mo~Yu, Yang Zhang, Karthik~R. Narasimhan, Joshua~B. Tenenbaum, and
  Chuang Gan.
\newblock Linking emergent and natural languages via corpus transfer.
\newblock In \emph{The Tenth International Conference on Learning
  Representations, {ICLR} 2022, Virtual Event, April 25-29, 2022}.
  OpenReview.net, 2022.
\newblock URL \url{https://openreview.net/forum?id=49A1Y6tRhaq}.

\bibitem[Dess{ì} et~al.(2023)Dess{ì}, Bevilacqua, Gualdoni, Rakotonirina,
  Franzon, and Baroni]{abs-2304-01662}
Roberto Dess{ì}, Michele Bevilacqua, Eleonora Gualdoni, Nathana{\"{e}}l~Carraz
  Rakotonirina, Francesca Franzon, and Marco Baroni.
\newblock Cross-domain image captioning with discriminative finetuning.
\newblock \emph{CoRR}, abs/2304.01662, 2023.
\newblock \doi{10.48550/arXiv.2304.01662}.
\newblock URL \url{https://doi.org/10.48550/arXiv.2304.01662}.

\bibitem[Lin et~al.(2014)Lin, Maire, Belongie, Bourdev, Girshick, Hays, Perona,
  Ramanan, Doll{\'{a}}r, and Zitnick]{LinMBHPRDZ14}
Tsung{-}Yi Lin, Michael Maire, Serge~J. Belongie, Lubomir~D. Bourdev, Ross~B.
  Girshick, James Hays, Pietro Perona, Deva Ramanan, Piotr Doll{\'{a}}r, and
  C.~Lawrence Zitnick.
\newblock Microsoft {COCO:} common objects in context.
\newblock \emph{CoRR}, abs/1405.0312, 2014.
\newblock URL \url{http://arxiv.org/abs/1405.0312}.

\bibitem[Zitnick and Parikh(2013)]{ZitnickP13}
C.~Lawrence Zitnick and Devi Parikh.
\newblock Bringing semantics into focus using visual abstraction.
\newblock In \emph{2013 {IEEE} Conference on Computer Vision and Pattern
  Recognition, Portland, OR, USA, June 23-28, 2013}, pages 3009--3016. {IEEE}
  Computer Society, 2013.
\newblock \doi{10.1109/CVPR.2013.387}.
\newblock URL \url{https://doi.org/10.1109/CVPR.2013.387}.

\bibitem[Dess{ì} et~al.(2022)Dess{ì}, Gualdoni, Franzon, Boleda, and
  Baroni]{DessiGFBB22}
Roberto Dess{ì}, Eleonora Gualdoni, Francesca Franzon, Gemma Boleda, and Marco
  Baroni.
\newblock Communication breakdown: On the low mutual intelligibility between
  human and neural captioning.
\newblock In Yoav Goldberg, Zornitsa Kozareva, and Yue Zhang, editors,
  \emph{Proceedings of the 2022 Conference on Empirical Methods in Natural
  Language Processing, {EMNLP} 2022, Abu Dhabi, United Arab Emirates, December
  7-11, 2022}, pages 7998--8007. Association for Computational Linguistics,
  2022.
\newblock URL \url{https://aclanthology.org/2022.emnlp-main.546}.

\bibitem[Kazemzadeh et~al.(2014)Kazemzadeh, Ordonez, Matten, and
  Berg]{KazemzadehOMB14}
Sahar Kazemzadeh, Vicente Ordonez, Mark Matten, and Tamara~L. Berg.
\newblock Referitgame: Referring to objects in photographs of natural scenes.
\newblock In Alessandro Moschitti, Bo~Pang, and Walter Daelemans, editors,
  \emph{Proceedings of the 2014 Conference on Empirical Methods in Natural
  Language Processing, {EMNLP} 2014, October 25-29, 2014, Doha, Qatar, {A}
  meeting of SIGDAT, a Special Interest Group of the {ACL}}, pages 787--798.
  {ACL}, 2014.
\newblock \doi{10.3115/v1/d14-1086}.
\newblock URL \url{https://doi.org/10.3115/v1/d14-1086}.

\bibitem[Lee et~al.(2018)Lee, Cho, Weston, and Kiela]{Lee2017}
Jason Lee, Kyunghyun Cho, Jason Weston, and Douwe Kiela.
\newblock Emergent translation in multi-agent communication.
\newblock In \emph{6th International Conference on Learning Representations,
  {ICLR} 2018, Vancouver, BC, Canada, April 30 - May 3, 2018, Conference Track
  Proceedings}. OpenReview.net, 2018.
\newblock URL \url{https://openreview.net/forum?id=H1vEXaxA-}.

\bibitem[Sapir(2014)]{sapir1921language}
Edward Sapir.
\newblock \emph{Language as a historical product: Drift}, page 157–182.
\newblock Cambridge Library Collection - Linguistics. Cambridge University
  Press, 2014.
\newblock \doi{10.1017/CBO9781139629430.008}.

\bibitem[Lakoff(1972)]{lakoff1972language}
Robin Lakoff.
\newblock Language in context.
\newblock \emph{Language}, pages 907--927, 1972.

\bibitem[Hamilton et~al.(2016)Hamilton, Leskovec, and
  Jurafsky]{hamilton2016cultural}
William~L Hamilton, Jure Leskovec, and Dan Jurafsky.
\newblock Cultural shift or linguistic drift? comparing two computational
  measures of semantic change.
\newblock In \emph{Proceedings of the Conference on Empirical Methods in
  Natural Language Processing. Conference on Empirical Methods in Natural
  Language Processing}, volume 2016, page 2116. NIH Public Access, 2016.

\bibitem[Lee et~al.(2019)Lee, Cho, and Kiela]{Lee2019}
Jason Lee, Kyunghyun Cho, and Douwe Kiela.
\newblock Countering language drift via visual grounding.
\newblock In Kentaro Inui, Jing Jiang, Vincent Ng, and Xiaojun Wan, editors,
  \emph{Proceedings of the 2019 Conference on Empirical Methods in Natural
  Language Processing and the 9th International Joint Conference on Natural
  Language Processing, {EMNLP-IJCNLP} 2019, Hong Kong, China, November 3-7,
  2019}, pages 4384--4394. Association for Computational Linguistics, 2019.
\newblock \doi{10.18653/v1/D19-1447}.
\newblock URL \url{https://doi.org/10.18653/v1/D19-1447}.

\bibitem[Kiela et~al.(2017)Kiela, Conneau, Jabri, and Nickel]{KielaCJN17}
Douwe Kiela, Alexis Conneau, Allan Jabri, and Maximilian Nickel.
\newblock Learning visually grounded sentence representations.
\newblock \emph{CoRR}, abs/1707.06320, 2017.
\newblock URL \url{http://arxiv.org/abs/1707.06320}.

\bibitem[Elliott et~al.(2016)Elliott, Frank, Sima'an, and
  Specia]{elliott2016multi30k}
Desmond Elliott, Stella Frank, Khalil Sima'an, and Lucia Specia.
\newblock Multi30k: Multilingual english-german image descriptions.
\newblock In \emph{Proceedings of the 5th Workshop on Vision and Language,
  hosted by the 54th Annual Meeting of the Association for Computational
  Linguistics, VL@ACL 2016, August 12, Berlin, Germany}. The Association for
  Computer Linguistics, 2016.
\newblock \doi{10.18653/v1/w16-3210}.
\newblock URL \url{https://doi.org/10.18653/v1/w16-3210}.

\bibitem[Lewis et~al.(2017)Lewis, Yarats, Dauphin, Parikh, and
  Batra]{lewis2017deal}
Mike Lewis, Denis Yarats, Yann Dauphin, Devi Parikh, and Dhruv Batra.
\newblock Deal or no deal? end-to-end learning of negotiation dialogues.
\newblock In \emph{Proceedings of the 2017 Conference on Empirical Methods in
  Natural Language Processing}, pages 2443--2453, Copenhagen, Denmark,
  September 2017. Association for Computational Linguistics.
\newblock \doi{10.18653/v1/D17-1259}.
\newblock URL \url{https://aclanthology.org/D17-1259}.

\end{thebibliography}

\clearpage

\begin{landscape}
\tiny  

\setlength\tabcolsep{1.5pt} 
\begin{longtable}{|c|cc|ccccc|cc|cccc|cc|}
\caption{This table contains all the relevant literature cited in this review sorted by year. 
A first row reports the broad categories mentioned in Section~\ref{sec:proprieties} and Section~\ref{sec:methods}, whereas a second row describes their specific features.
The \textbf{interaction type} category has the following abbreviations as mentioned in Section~\ref{ssec:prop:interaction}: \textit{inner} (In), \textit{outer} (Out), \textit{cooperation} (Coop), \textit{competition} (Comp), \textit{population} (Pop), \textit{iterative} (Iter).
Most columns are marked with either a cross or a white space, indicating whether or not the corresponding feature is present. Three categories, however, contain some abbreviations.
The \textbf{game environment} is divided into two features \textit{communication-focused} (Com-focus) and \textit{communication-assisted} (Com-ast); each of these features has the following values specifying the possible game settings: \textit{referential} (Ref) , \textit{navigation} (Nav) , \textit{negotiation} (Neg), \textit{social game} (SG), \textit{visual question answering} (VQA), \textit{machine translation} (MT), \textit{task and talk} (TnT).
The \textbf{update algorithm} contains the following category abbreviations:  \textit{reinforcement learning} (RL), \textit{Gumbel Softmax} (GS), \textit{supervised learning} (SL). Moreover it specifies alternative methods: \textit{image labeling} (IML), \textit{dialoguing} (DG), \textit{theory of mind} (ToM), \textit{unsupervised} (UnS), \textit{evolutionary learning} (EvL) and \textit{stochastic computational graphs} (SCG).
The \textbf{Theory of Mind} contains the following abbreviations: \textit{action-prediction} (ActPred), \textit{Trustworthiness} (Trust), \textit{imitation learning} (ImL), \textit{obverter technique} (ObvTec), \textit{speaker-selection} (SpeakSlct).}
\label{tab:list}\\
\hline
Citation & \multicolumn{2}{c|}{Theory of mind} & \multicolumn{5}{c|}{Interaction Type} & \multicolumn{2}{c|}{Game Environment} & \multicolumn{4}{c|}{Update Algorithm} & \multicolumn{2}{c|}{Dichotomy} \\ \hline
\endfirsthead
\multicolumn{16}{c}%
{{\bfseries Table \thetable\ continued from previous page}} \\
\hline
Citation & Modeling & Influencing & In. Coop & Out. Coop & Comp & Pop. & Iter. & Com-focus & Com-ast & RL & GS & SL & other & Machine & Human\\ \hline
\endhead
 & Modeling & Influencing & In. Coop & Out. Coop & Comp & Pop. & Iter. & Com-focus & Com-ast & RL & GS & SL & other & Machine & Human \\ \hline
\cite{Kirby2014} &  &  & x &  &  &  & x & x &  &  &  &  &  &  & x \\ \hline
\cite{Nakamura2016} & Trust &  & x & x & x &  &  &  & SG &  &  &  &  & x &  \\ \hline
\cite{Foerster2016} &  &  & x &  &  &  &  & x &  & x &  &  &  & x &  \\ \hline
\cite{Lazaridou2016} &  &  & x &  &  &  &  & Ref &  &  &  & IML &  & x & x \\ \hline
\cite{Sukhbaatar2016} &  &  & x & x & x &  &  &  & x & x &  & x &  & x &  \\ \hline
\cite{Andreas2016} & x & x & x &  &  &  &  & Ref &  &  &  & x &  &  & x \\ \hline
\cite{Kottur2017} &  &  & x &  &  &  &  & TnT &  & x &  &  &  & x &  \\ \hline
\cite{Lowe2017} & ActPred &  & x & x & x &  &  & x & Nav & x & x &  &  & x &  \\ \hline
\cite{Evtimova2017} & x &  & x &  &  &  &  & Ref &  & x &  &  &  & x &  \\ \hline
\cite{Havrylov2017} &  &  &  &  &  &  &  & Ref &  & x & x &  &  & x & x \\ \hline
\cite{Kong2017} &  &  & x &  &  &  &  &  & x & x &  &  &  & x &  \\ \hline
\cite{Das2017} &  &  & x &  &  &  &  & VQA &  & x &  & DG &  &  & x \\ \hline
\cite{Foerster2017} & x & x &  &  &  &  &  &  &  & x &  &  &  &  &  \\ \hline
\cite{Lee2017} &  &  & x &  &  &  &  & Ref &  & x &  & x &  &  & x \\ \hline
\cite{Mordatch2018} &  &  & x &  &  &  &  &  & Nav & x &  &  &  & x &  \\ \hline
\cite{Grover2018} & ImL &  & x & x &  &  &  &  & Nav & x &  &  & UnS & x &  \\ \hline
\cite{abs-1804-02341} & ObvTec &  & x &  &  &  &  & Ref &  & x &  & ObvTec &  & x &  \\ \hline
\cite{Cao2018} & Trust &  & x & x &  & x &  &  & Neg & x &  &  &  & x &  \\ \hline
\cite{abs-1810-08647} & x & x & x &  &  &  &  &  & SG & x &  & ToM &  & x &  \\ \hline
\cite{Bouchacourt2018} &  &  & x &  &  &  &  & Ref &  & x &  &  &  & x &  \\ \hline
\cite{Lazaridou2018} &  &  & x &  &  &  &  & Red &  & x &  &  &  & x &  \\ \hline
\cite{Raileanu2018} & Trust &  & x &  &  &  &  &  &  & x &  & ToM &  &  &  \\ \hline
\cite{Fitzgerald2019} &  &  & x &  &  & x &  & Ref &  & x &  &  &  & x &  \\ \hline
\cite{Graesser2019} &  &  & x &  &  & x & x & Ref &  & x &  & x &  & x &  \\ \hline
\cite{Li2019} &  &  & x &  &  & x & x & Ref &  & x &  &  &  & x &  \\ \hline
\cite{Lowe2019a} &  &  & x &  &  & x & x & TnT &  & x &  &  &  & x &  \\ \hline
\cite{abs-1809-00549} & ObvTec &  & x &  &  &  &  & x & x & x &  & ObvTec &  & x &  \\ \hline
\cite{Das2019} & SpeakSlct & x & x & x & x &  &  &  & Nav & x &  & x &  & x &  \\ \hline
\cite{Rodriguez2019} & x &  & x & x &  & x &  & Ref &  & x &  & ToM &  & x &  \\ \hline
\cite{abs-1910-05291} &  &  & x & x &  & x & x &  & x & x & x &  &  & x &  \\ \hline
\cite{abs-1910-06079} &  &  & x &  &  &  &  & x &  & x & x &  &  & x &  \\ \hline
\cite{abs-1912-05676} &  &  &  &  &  &  &  & x & Nav & x &  &  &  & x &  \\ \hline
\cite{abs-1912-06208} &  &  & x & x &  & x & x & x &  &  &  & x &  & x &  \\ \hline
\cite{Bouchacourt2019} &  &  &  &  &  &  &  &  &  & x &  &  &  & x &  \\ \hline
\cite{Chaabouni2019} &  &  & x &  &  &  & x &  & Nav &  &  & x &  & x &  \\ \hline
\cite{Omidshafiei2019} &  &  & x & x &  &  &  &  & x & x & x &  &  & x &  \\ \hline
\cite{GuptaLFKP19} &  &  & x &  &  & x & x & Ref &  & x & x & x &  & x & x \\ \hline
\cite{abs-1905-12561} &  &  & x &  &  &  &  & Ref &  & x & x &  &  & x & x \\ \hline
\cite{JaquesLHGOSLF19} & x & x & x & x & x &  &  &  & x & x &  & ToM &  &  &  \\ \hline
\cite{Lee2019} &  &  & x &  &  &  &  & MT &  & x &  & DG &  &  & x \\ \hline
\cite{Resnick2019} &  &  &  &  &  &  &  &  &  &  & x &  &  &  &  \\ \hline
\cite{abs-1911-01892} &  &  & x &  &  &  &  & Ref &  & x &  &  &  & x &  \\ \hline
\cite{Liang2020} &  &  & x &  & x &  &  & TnT &  & x &  &  &  & x &  \\ \hline
\cite{abs-2001-07752} & x &  & x &  &  &  &  & Ref &  & x &  &  &  & x &  \\ \hline
\cite{abs-2002-01365} &  &  & x &  &  &  & x & Ref &  & x &  & x &  & x &  \\ \hline
\cite{Chaabouni2020} &  &  & x &  &  &  &  & Ref &  & x & x &  &  & x &  \\ \hline
\cite{abs-2004-03420} &  &  & x &  &  &  &  & Ref &  &  &  & x &  & x &  \\ \hline
\cite{abs-2012-10776} &  &  & x &  &  &  &  & Ref &  & x & x &  &  & x &  \\ \hline
\cite{abs-1912-05304} &  &  & x &  &  &  &  &  & x & x &  &  & UnS & x &  \\ \hline
\cite{abs-2005-05441} &  &  & x & x & x &  &  &  & x & x & x &  &  & x &  \\ \hline
\cite{abs-1911-06992} &  &  & x & x & x &  &  &  & x & x &  &  &  & x &  \\ \hline
\cite{KharitonovCBB20} &  &  & x &  &  &  &  &  & x & x & x &  & SCG & x &  \\ \hline
\cite{Bachrach2020} &  &  & x & x &  &  &  &  &  & x &  & x &  &  &  \\ \hline
\cite{Xie2020} & x & x &  &  &  &  &  &  &  & x &  & x &  &  &  \\ \hline
\cite{HawkinsKSG20} & x & x & x &  &  &  &  & Ref &  &  &  & x &  &  & x \\ \hline
\cite{abs-2003-12694} &  &  & x &  &  &  & x & Ref &  & x & x & x &  &  & x \\ \hline
\cite{abs-2010-02975} &  &  & x & x &  &  & x & x &  & x & x & x &  &  & x \\ \hline
\cite{Cogswell2020} &  &  &  &  &  &  &  & VQA &  & x & x & x &  &  & x \\ \hline
\cite{Lazaridou2020} &  &  & x &  &  &  &  & Ref &  & x &  & x &  &  & x \\ \hline
\cite{Li2020} &  &  & x &  &  &  &  & Ref &  &  & x & x &  &  & x \\ \hline
\cite{Lowe2020} &  &  & x &  &  & x & x & Ref &  &  & x & x &  &  & x \\ \hline
\cite{abs-2010-15896} &  &  & x &  &  & x &  & Ref &  & x &  & x &  &  &  \\ \hline
\cite{Brandizzi2021} & Trust &  & x & x & x &  &  &  & SG & x &  &  &  & x &  \\ \hline
\cite{abs-2001-03361} &  &  & x & x &  & x & x & Ref &  &  & x & x & EvL & x &  \\ \hline
\cite{abs-2106-04258} &  &  & x &  &  &  &  & Ref &  &  & x & x &  & x &  \\ \hline
\cite{abs-2107-05697} & x & x & x &  &  & x &  & Ref & Nav & x & x & x &  &  & x \\ \hline
\cite{abs-2110-05422} &  &  & x &  &  & x &  & Ref &  &  & x & x &  &  & x \\ \hline
\cite{MihaiH21} &  &  & x &  &  &  &  & Ref &  & x &  & x &  &  &  \\ \hline
\cite{abs-2204-12982} &  &  & x & x &  & x &  & Ref &  & x & x &  &  & x &  \\ \hline
\cite{ChaabouniSATTDM22} &  &  & x & x &  & x &  & x &  & x &  & ImL &  & x &  \\ \hline
\cite{abs-2111-14210} &  &  & x &  &  &  &  & Ref &  & x &  &  &  &  &  \\ \hline
\cite{YaoYZNTG22} &  &  & x &  &  & x &  & Ref &  & x & x & x &  &  & x \\ \hline
\cite{abs-2302-08913} &  &  & x &  &  & x &  & Ref &  & x & x &  &  &  &  \\ \hline
\end{longtable}
\end{landscape}

\end{document}